\documentclass[manuscript]{acmart}

\AtBeginDocument{%
  }

\usepackage[capitalize,noabbrev]{cleveref}
\usepackage{acronym}
\usepackage{graphicx}
\usepackage{subcaption}
\usepackage{todonotes}
\usepackage{array,booktabs}
\usepackage{multirow}
\usepackage{adjustbox}

\acrodef{fido2}[FIDO2]{Fast IDentity Online 2}
\acrodef{dp}[DP]{Differential Privacy}
\acrodef{BPR}[BPR]{Bayesian Personalized Ranking}
\acrodef{DRS}[DRS]{Deep Recommender System}
\acrodef{DPSGD}[DPSGD]{Differentially Private Stochastic Gradient Descent}
\acrodef{RSs}[RSs]{Recommender Systems}
\acrodef{RS}[RS]{Recommender System}
\acrodef{MLP}[MLP]{Multilayer Perceptron}
\acrodef{DP}[DP]{Differential Privacy}
\acrodef{SVD}[SVD]{Singular Value Decomposition}
\acrodef{NCF}[NCF]{Neural Collaborative Filtering}
\acrodef{KLD}[KLD]{Kullback-Leibler Divergence}
\acrodef{PL}[PL]{Popularity Lift}
\acrodef{SGD}[SGD]{Stochastic Gradient Descent}
\acrodef{VAE}{Variational Autoencoder}
\acrodef{VAEs}{Variational Autoencoders}
\acrodef{ldp}[LDP]{Local Differential Privacy}
\acrodef{MSE}[MSE]{Mean Squared Error}
\acrodef{BCE}[BCE]{Binary Cross Entropy}
\acrodef{DPF}[DPF]{Deviation from Producer Fairness}

\begin{document}

\title{Privacy--Utility--Bias Trade-offs for Privacy-Preserving Recommender Systems}

\author{Shiva Parsarad}
\affiliation{%
  \institution{University of Basel}
  \city{Basel}
  \country{Switzerland}}  
\email{shiva.parsarad@unibas.ch}
\orcid{0000-0003-0392-8667}

\author{Isabel Wagner}
\affiliation{%
  \institution{University of Basel}
  \city{Basel}
  \country{Switzerland}}
\email{isabel.wagner@unibas.ch}
\orcid{0000-0003-0242-6278}


\begin{abstract}
    \acp{RS} output ranked lists of items, such as movies or restaurants, that users may find interesting, based on the user's past ratings and ratings from other users.
    %To mitigate privacy concerns, privacy-preserving recommender systems have been proposed, typically incorporating a variant of differential privacy.
    %Here, we study how privacy protection affects (1) the performance of the recommender system and (2) biases in the list of recommended items.
    %We select four popular types of recommender systems (neural collaborative filtering, Bayesian personalized ranking, matrix factorization, and Variational Autoencoders), train them with differentially private stochastic gradient descent and local differential privacy, and evaluate performance, privacy, and bias on two datasets (MovieLens and Yelp).
    \acp{RS} increasingly incorporate \ac{dp} to protect user data, raising questions about how privacy mechanisms affect both recommendation accuracy and fairness.
 	We conduct a comprehensive, cross-model evaluation of two DP mechanisms, \ac{DPSGD} and \ac{ldp}, applied to four recommender systems (\ac{NCF}, \ac{BPR}, \ac{SVD}, \ac{VAE}) on the MovieLens-1M and Yelp datasets.
    %Our results align with prior findings, specifically that privacy protection has a negative impact on performance, as well as a bias towards popularity.
 	%Our results reveal substantial heterogeneity in privacy–utility–bias trade-offs across models, datasets, and user groups.
	%Our findings indicate that utility decreases as privacy increases, but the magnitude of this decrease varies widely.
	%For example, \ac{NCF} trained with \ac{DPSGD} maintains high utility with less than 10\% degradation at $\epsilon\approx1$, whereas \ac{SVD} and \ac{BPR} experience 20–40\% drops for niche users.
	%\ac{VAE} utility is the most vulnerable, with sharp performance drops for sparsely represented groups.
	We find that stronger privacy consistently reduces utility, but not uniformly.
	\ac{NCF} under DPSGD shows the smallest accuracy loss (under 10\% at $\epsilon\approx1$), whereas \ac{SVD} and \ac{BPR} experience larger drops, especially for users with niche preferences. 
	\ac{VAE} is the most sensitive to privacy, with sharp declines for sparsely represented groups.

 	The impact on bias metrics is similarly heterogeneous.
	%Some models maintain stable behavior under privacy, while others show increased popularity bias or worsened calibration for specific datasets.
    \ac{DPSGD} generally reduces the gap between recommendations of popular and less popular items, whereas \ac{ldp} preserves existing patterns more closely.
 	These results highlight that no single DP mechanism is uniformly superior; instead, each provides trade-offs under different privacy regimes and data conditions.

\end{abstract}

%	, Stochastic gradient descent}

% rough plan for page numbers:
% abstract + introduction: 1 page
% related work + background: 0.75 pages
% method: 1 page
% results: 1 page
% conclusion: 0.25 pages

\maketitle
\section{Introduction}
%\todo[inline]{Fairness vs Bias: use only one of the two terms -- we settled on "bias" a while ago, but the text uses "fairness" a lot now.}

%\iw{This comment is still relevant: In your next round of editing, focus on rewriting sentences that use passive voice in active voice. We want to avoid passive voice as much as possible.}
%\iw{In addition, check your paragraph structure: each paragraph should have one topic, and the paper should (roughly) make sense if you only read the first sentences of paragraphs}

%%% Recommender systems
\acfp{RS} are mechanisms for filtering information and presenting personalized ranked lists of items to users.
%, a phenomenon that has been widespread due to the evolving behaviors of computer users, the prevalence of personalization trends, and the increasing availability of internet access.  
%Various entities, such as corporations, governments, and criminal organizations, engage in the active collection of personal information that is publicly accessible on the Internet \cite{caliskanislam2014privacy}, that could be problematic for user's privacy, and arise some critical concerns.
\acp{RS} are widely used for a variety of applications, e.g., books, news, or movies, thanks to their ability to filter and rank vast amounts of information available online.

%%% Privacy in recommender systems
However, past user data can reveal sensitive personal information \cite{razavi2020personality, bi2013inferring, peng2015predicting,weinsberg2012blurme, enaizan2020electronic, polatidis2017privacypreserving}.
%The aforementioned attributes of RSs give rise to new privacy concerns due to the delicate nature of users' data and its susceptibility to unauthorized access.
For example, our preferred movie genres can expose our personality traits \cite{greenberg2016song} or be used for advertising purposes.
%\iw{simplify words: instead of ``utilize'', write ``use''}
To address these privacy concerns, privacy-preserving recommender systems have been proposed, often based on a form of \acf{dp}.
\ac{DPSGD}, for example, applies DP during the training process of a machine learning model, thereby guaranteeing privacy for the trained model \cite{abadi2016deep}, whereas \ac{ldp} applies DP to each user's data before training the recommender model \cite{mullner2024impactb}.
%it is a practical implementation of \ac{DP} used specifically for training machine learning models
%We consider this method as a way of enhancing privacy in multiple known recommendation systems.
While these techniques provide provable privacy guarantees, they may degrade recommendation quality, and increasingly, it is recognized that privacy mechanisms can also influence bias and fairness in \acp{RS} \cite{mullner2024impactb, difazio2024enhancing}.
%\iw{need a reference to substantiate this claim of "increasingly recognized"}
%%% Bias in recommender systems
\ac{RSs} suffer from well-known biases, for example, \acp{RS} can influence which information is readily available to users because users rarely visit search results beyond the first page~\cite{hannak2013measuring}, and they can skew recommended item lists towards popular items, thereby reducing the relevance of recommended items to users with niche tastes \cite{elahi2021investigating}.
%This makes it crucial to evaluate bias and fairness in \ac{RS}.
%individuals' online consumption and experiences, making it crucial to evaluate their impact on society and determine if they are influenced by any potential biases. 
%This miscalibration in recommended item lists can be addressed with calibration methods, i.e., postprocessing steps that adjust the recommended item list to better match the user's interests \cite{dasilva2021exploiting}.
%Recommending only popular items will not enhance new item discovery and will ignore users' interests with niche taste \cite{elahi2021investigating}.
%It is reported that \cite{abdollahpouri2020connection} popularity bias leads to miscalibration in recommendation.
%Calibration shows how a recommendation algorithm covers different user's preferences.
%Calibration could influence long term user's satisfaction and ensure that a particular part of the user's profile will not dominate or overtake the whole of the offer.
%Both privacy and bias play crucial roles in enhancing user happiness, promoting a sense of safety, and addressing ethical concerns.

%%% Gap in research: Interplay between performance, privacy, bias
Although the impact of \ac{RS} on privacy and bias is known, research on these issues remains largely separate, i.e., the important question of how privacy mechanisms influence bias in \ac{RSs} remains understudied.
%But, what is the correlation between these two important concepts? whether privacy enhancing approaches increase fairness or privacy comes at the cost of  fairness?
%Recent work proposes a privacy-preserving implementation of the fairness mechanism equity of amortized attention ranking \cite{sun2023privacypreserving}.
%However, this recent work assumes that the underlying recommender model was already trained in a privacy-preserving manner.
%In \cite{sun2023privacypreserving}, a study of item fairness and user privacy is conducted. However, it focuses on item fairness.
%The impact of \ac{DP} on recommendation accuracy and popularity bias is studied in \cite{mullner2024impactb}, but it focuses only on \ac{ldp} instead of applying \ac{DP} during the training process.
%The relationship between \ac{DP} and bias in recommender systems has been previously investigated~\cite{mullner2024impactb}, although that analysis was limited to the effect of \ac{ldp} on popularity bias. 
Most prior work studies privacy and fairness separately or evaluates only specific models and specific kinds of bias~\cite{mullner2024impactb}.
A few works consider fairness across user or item groups~\cite{deldjoo2024fairness}, but the broader question of how different \ac{dp} mechanisms influence utility and bias across different user groups, item types, and model architectures is still underexplored.
This is an important gap because different \ac{RS} models have different sensitivities to noise: deep neural \ac{RSs} may absorb noise through high-capacity representations, matrix-factorization methods rely on low-dimensional latent factors that are more fragile under perturbation, and generative models such as VAEs can collapse under high sparsity.
Understanding these differences is essential for deploying privacy-preserving systems that are both accurate and debiased.
%\iw{update research questions after finalizing results}

To close this gap in knowledge, we study the following research questions:
How do DPSGD and LDP affect ranking utility? How do privacy mechanisms influence bias metrics such as miscalibration, popularity lift, novelty, coverage, and producer fairness? How are niche, diverse, and blockbuster users affected differently? Under what conditions can private RSs maintain acceptable accuracy without increasing bias?
%%% Contributions
%\iw{rewrite contributions section after finalizing results}

In this work, we systematically examine the trade-offs between privacy, utility, and bias across four representative families of RS models including \ac{SVD}, \ac{BPR}, \ac{NCF}, and \ac{VAE}, using two real-world datasets (MovieLens-1M and Yelp) and two DP mechanisms (\ac{DPSGD} and \ac{ldp}). 
Our study spans a wide range of privacy budgets ($\epsilon$), multiple user and item subgroups, and five complementary bias metrics. 
Our main contributions are as follows:
\begin{itemize}
	\item \textbf{Comprehensive evaluation of DP mechanisms across model families}. 
	We implement and analyze \ac{DPSGD} and \ac{ldp} across four representative recommender-system architectures: \ac{SVD}, \ac{BPR}, \ac{NCF}, and \ac{VAE}.
	\item \textbf{Analysis across multiple privacy levels and datasets}. We evaluate the impact of both DP mechanisms over a wide range of privacy budgets ($\epsilon$), using two widely adopted benchmark datasets: MovieLens 1M and Yelp.
	\item \textbf{Utility and bias trade-off evaluation}. We measure ranking quality (NDCG) and assess multiple bias dimensions, including miscalibration, popularity lift, novelty, item coverage, and deviation from producer fairness.
	\item \textbf{Fine-grained examination of group fairness\footnote{Please note that in this paper, we interpret fairness as reduced or controlled bias, and we measure it using bias-oriented metrics.} across user and item subgroups}. We analyze impacts across different user profiles (niche, blockbuster, and diverse) and item categories (popular vs.\ unpopular), revealing heterogeneous fairness impacts across sub-populations.
	\item \textbf{Identification of privacy–fairness “sweet spots"}. We uncover privacy regimes, especially moderate $\epsilon$ that offer strong utility with only mild changes in bias, while also highlighting conditions where no such sweet spots exist (e.g., \ac{VAE} on sparse data).
\end{itemize}

%This research tries to determine whether there exists a specific threshold in our examined recommender systems where we can get both accurate and private outcomes simultaneously.
%Consequently, we evaluate the impact of DPSGD on popularity bias and miscalibration to determine whether DPSGD method may exacerbate bias and unequal effects for various user groups in recommender systems. 

%But in this paper, we measure different metrics on both the group and individual levels. 
%For group-level evaluation, we have defined two groups for items and three for users. 
%Therefore, we evaluated bias for consumers/users and producers/items.
%\iw{rewrite findings section after we have finalized results}
%%% Findings
%We find that the privacy--performance trade-off depends on both the dataset and the recommender approach that is selected.
%For the same privacy cost, we observe worse performance for Yelp compared to 1M.
%In addition, we find privacy settings where the performance loss compared to non-private recommendations is acceptable, at reasonable privacy cost. 
%\ac{DPSGD} generally provides better ranking effectiveness but can reduce coverage.
%\ac{ldp} maintains more stable performance across privacy levels but tends to favor popular items.
%Privacy constraints impact different user groups and item popularity differently, with niche users and unpopular items being more affected.
Our findings show that the impact of \ac{DP} on utility and bias is highly dependent on the choice of DP mechanism, the underlying model architecture, and dataset characteristics.
\ac{NCF} trained with DPSGD maintains high utility (within ~10\% of non-private performance) even under strong privacy ($\epsilon\approx1$), forming a clear sweet spot where utility remains high and most bias metrics remain within their acceptable ranges (e.g., no extreme positive or negative \ac{PL}).
 %\iw{"behave smoothly" - what does this even mean? please use clear language and plainly say what you mean.}
In contrast, \ac{SVD}, and \ac{BPR} show moderate sensitivity to privacy mechanisms, 
%\iw{to what?}
and \ac{VAE} collapses under increased sparsity, offering no meaningful sweet spots.
Bias outcomes vary similarly: miscalibration is mostly stable except for NCF/Yelp; \ac{DPF} ranges widely ($\approx -0.8$ to +1.0), with \ac{DPSGD} reducing head–tail distinctions at low $\epsilon$ and \ac{ldp} preserving baseline biases.

%Specifically, \ac{DPSGD} tends to preserve utility more effectively at moderate privacy levels, whereas \ac{ldp} produces more stable but weaker performance.
%We also observe heterogeneous bias effects across user and item groups: privacy noise reduces popularity amplification but introduces disparities in accuracy, particularly for niche users and long-tail items.
%This aligns with prior work \cite{shams2024evaluating} that found the popularity bias is worst for niche users. 
%These results highlight that no single DP mechanism uniformly outperforms the other, and that utility–bias trade-offs vary significantly across models and privacy regimes.
%In this paper, fairness is interpreted as reduced or controlled bias, and we measure it using bias-oriented metrics.
Overall, our findings show that no single DP mechanism is universally superior.
Instead, the best privacy–utility–bias trade-off depends on the model architecture, dataset sparsity, and user or item subgroup.
These insights help guide the deployment of privacy-preserving recommender systems that balance privacy, accuracy, and fairness in practice.

\section{Background and Related Work}
%\todo[inline]{merge background and related work sections, make sure there is a logical flow for readers that starts with basic+general and progresses to more detailed information, without duplicating content}
\subsection{Recommender systems}
%Recommender systems play a crucial role in improving user experience by customizing content and services.
The three main types of \ac{RSs} involve collaborative filtering, content-based filtering, and hybrid approaches.
%\iw{it is not clear from the following text which of the four RS we studied is of which type} 
%Each type uses different methods to predict and personalized recommendations based on individual user preferences.
%\todo[inline]{start with an introductory sentence on the different types of recommender systems}
The most common technique used for recommendations is collaborative filtering.
In this paper, we investigate several methods for recommendation systems, including ranking-based Bayesian Personalized Ranking (BPR), matrix factorization methods such as Singular Value Decomposition (SVD), hybrid models like Neural Collaborative Filtering (NCF), and Variational Autoencoders (VAEs) \cite{liang2018variational}. 
%We also explored the use of Variational Autoencoders (VAEs) \cite{liang2018variational} for collaborative filtering with implicit feedback.
\ac{BPR} and \ac{SVD} are both collaborative filtering methods based on matrix factorization. \ac{NCF} is a hybrid model that combines neural network architectures with collaborative filtering principles.
\ac{VAEs} are also used for collaborative filtering with implicit feedback.

\ac{BPR} is an optimization criterion based on the maximum posterior estimator which enables us to train models that are optimized for ranking instead of the item prediction task \cite{rendle2009bpr}.
%\todo[inline]{BPR ... is a loss -- is this true? is BPR a "loss" or is it an algorithm?}
%BPR \cite{rendle2009bpr} is a pairwise personalized ranking loss that is derived from the maximum posterior estimator. 
The training data, $D_{s}$ consists of tuples $(u,i,j)$, meaning that user $u$ prefers the positive item $i$ over the negative item $j$. 
To maximize the posterior probability, BPR uses stochastic gradient descent.
BPR can support learning different model classes, including k-nearest-neighbor and matrix factorization, used for recommender systems

\ac{SVD} decomposes a user-item interaction matrix $ R $ into three matrices: $ U $ containing user-specific latent factors, $ \Sigma $ containing singular values that scale the latent factors, and $ V $ containing item-specific latent factors.
By truncating $ \Sigma $ and the corresponding columns of $ U $ and $ V $ to the top $ k $ components, SVD reduces the dimensionality of $ R $, capturing the most significant underlying structures while reducing computational complexity.
The resulting low-rank approximation $ \hat{R} = U_k \Sigma_k V_k^T $ estimates the ratings matrix, predicting unknown ratings and revealing patterns in user-item interactions \cite{mehta2017review}.

\ac{RSs} based on deep learning, such as \ac{NCF} \cite{he2017neural}, can 
%has recently attracted considerable interest due to its ability to 
overcome limitations of traditional models and achieve better performance \cite{zhang2020deep}.
Specifically, a \ac{NCF} model that combines a Multilayer Perceptron and Matrix Factorization (MF) to predict user interests
%In the study conducted by \cite{heNeuralCollaborativeFiltering2017}, the researchers employed a Multilayer Perceptron (MLP) to predict user interests.
%The results 
showed significant improvements in performance compared to conventional approaches \cite{he2017neural}.
% like Matrix Factorization.
%They proposed neural matrix factorization model that combines the MLP and Matrix Factorization (MF) concepts.
In this paper, we use the same architecture for the deep recommender system we investigated.

\ac{VAEs} generalize traditional latent-factor models by introducing non-linear, probabilistic latent variables through neural networks.
For each user $ u $, a latent representation $ \mathbf{z}_u $ is sampled from a prior distribution $\mathbf{z}_u \sim \mathcal{N}(0, I)$. 
The user-item interactions $ \mathbf{x}_u $ are then generated by using a multinomial likelihood $\mathbf{x}_u \sim \text{Multinomial}(N_u, \text{softmax}(f_{\theta}(\mathbf{z}_u)))$.

The model is trained by maximizing the Evidence Lower Bound (ELBO), which balances reconstructing the observed data and regularizing the latent space:
\[
\mathcal{L}(\mathbf{x}_u; \theta, \phi) = \mathbb{E}_{q_{\phi}(\mathbf{z}_u | \mathbf{x}_u)} [\log p_{\theta}(\mathbf{x}_u | \mathbf{z}_u)] - \text{KL}(q_{\phi}(\mathbf{z}_u | \mathbf{x}_u) \parallel p(\mathbf{z}_u))
\]
Balanced approach enables \ac{VAEs} to capture complex patterns in the data, significantly improving recommendation performance on large datasets \cite{liang2018variational}.

%\subsubsection{Singular Value Decomposition (SVD)}
%, making SVD a powerful tool for addressing missing values in the matrix.
%\todo[inline]{reference for SVD?}

%\subsubsection{Bayesian Personalized Ranking (BPR)}
%BPR aims to maximize the posterior probability:
%
%\begin{equation}\label{BPR-minimization}
%\sum_{(u, i, j) \in D_S} \ln \sigma\left(\hat{x}_{u i j}\right)-\lambda_{\Theta}\|\Theta\|^2
%\end{equation}
%
%In equation \ref{BPR-minimization}, $\hat{x}_{u i j}$
%is a real valued function that represents the relation between user u , item i and item j and is generally calculated by using Matrix factorization model.
%the estimator $\hat{x}_{u i j}$ composed as follow: $\hat{x}_{u i j} = \hat{x}_{u i} - \hat{x}_{u j}$, where $\sigma$ is the logistic sigmoid.
%: $\sigma(x) = \frac{1}{1 + e^{-x}}$.

%\subsubsection{Deep Recommender Systems (DRS)}
%Different applications of deep learning gain a lot of attention in the recent years.
%About recommender systems, deep learning bringing more opportunities to improve the performance.

%\subsection{Privacy-Preserving Recommender Systems}
\subsection{Privacy Challenges and Solutions in Recommender Systems}
%\iw{structure matters. paragraphs in this section are too long and cover multiple topics. please follow best practice in structuring paragraphs, check books on writing for explanations.}
Recommending items based on a user's preferences or past interactions leads to privacy issues, including:
%problems due to the potential inclusion of sensitive information, such as age, gender, and other personal data \cite{ferreira2023recommender}. 
%Data privacy issues include various aspects, including but not limited to: 
(i) personal data breaches, (ii) leaking of information that can enable user tracking, and (iii) the disclosure of user preferences via recommendations \cite{himeur2022latest,ferreira2023recommender}.
For example, an attacker can reveal sensitive information, such as the age of users, from a variational autoencoder recommender (VAE) system \cite{ganhor2022unlearning}.
To this end, once the training of VAE is complete, an attacker network is introduced to the model, which aims to predict sensitive attribute age from the latent vector.  
%Even some approaches that try to improve transparency can be harmfull for user's privacy.\iw{too colloquial}
%For example, as discussed in \cite {ge2024survey}, in social media, when a system tries to be explainable and explains that I recommend this because of your friends' interests, it can leak the privacy of friends.\iw{too colloquial}
Even approaches that are designed to enhance transparency in \ac{RSs} can compromise user privacy.
For instance, in the domain of social media, when a system seeks to provide explanations for its recommendations by citing the interests of friends, it risks violating the privacy of those individuals~\cite{ge2024survey}.

Approaches to protect user privacy include methods based on differential privacy, methods focusing on specific parts of the data, and methods based on recommendation unlearning.
%\iw{in scientific writing, word repetitions (always using the same word for the same concept) are a good thing! In the sentence above, you used methods, approaches, techniques, and strategies, all essentially meaning the same thing. I replaced most with ``methods'', which makes the sentence easier to parse.}
Generally, \ac{DP} proposed to minimize the likelihood of inferring the presence or absence of any specific record in the input.
\ac{DP} provides a theoretical privacy guarantee, the strength of which is governed by the proper privacy budget $\epsilon$, where smaller $\epsilon$ indicates higher privacy protection.
%An algorithm $M$ gives $\epsilon$-differential privacy if for any two adjacent datasets $D_1$ and $D_2$ differing on at most one row, and any possible output $S$:
%\[
%\Pr[M(D) \in S] \leq e^\epsilon \cdot \Pr[M(D') \in S]
%\]\iw{equation is not compiled correctly (square brackets do not show up in the PDF)}
%#\iw{use $\epsilon$, not $\varepsilon$}
%The privacy budget $\epsilon$ measures the level of privacy protection.
$(\epsilon, \delta)\text{-DP}$, called \textit{approximate differential privacy}, is a more relaxed version of DP, where $\delta$ bounds the probability that the privacy guarantee fails, and such failures happen with at most probability $\delta$.
%\iw{this is incorrect! please read up on what $\delta$ actually means. also please mention the name of this variant.}
A randomized mechanism $M$ is considered $(\epsilon, \delta)\text{-DP}$ if the probability of a privacy leakage is no more than $\delta$: 
%\iw{please check round brackets in all equations, they are not typeset correctly.}
\begin{equation}
	\Pr\big[M(D) \in S\big] \leq e^\epsilon \cdot \Pr\big[M(D') \in S\big] + \delta
\end{equation}

There are several strategies for integrating \ac{DP} into \ac{RSs}, each targeting different aspects of the process.
One approach focuses on applying \ac{DP} to input data. For instance, a DP-based method employs k-means clustering and utilizes an exponential mechanism to implement differentially private user-based collaborative filtering \cite{chen2021differentially}. 
Similarly, randomized response is used as a local differential privacy technique to ensure input data privacy \cite{mullner2024impactb}.
Another category incorporates noise directly into the objective function during training to achieve differential privacy.
For example, an auto-encoder-based recommender system that adds noise to the objective function \cite{fang2022differentially}.
In addition to these methods, some approaches utilize embeddings to apply \ac{DP} during the transformation of raw data into feature representations.
Embedding-Aware Noise Addition (EANA) \cite{ning2022eana} selectively adds noise to parameters with non-zero gradients. 
EANA improves the efficiency of training embedding-based neural networks with \ac{DP}, although it provides privacy guarantees only if the adversary has access solely to the final trained model.

In addition to the aforementioned approaches, another method incorporates \ac{DP} mechanisms into the gradients during model training.
Differentially private stochastic gradient descent (DPSGD), introduced for deep learning \cite{abadi2016deep}, is a generic technique that ensures differential privacy for any training process performed with SGD by using the Gaussian mechanism to add noise to the clipped gradients\cite{fang2022differentiallyc}.
A variation of \ac{DPSGD} uses Stochastic Gradient Langevin Dynamics (SGLD) to build a differentially private recommendation system based on matrix factorization \cite{liu2015fast}.
However, SGLD is slower than SGD and complete models have the convergence problem \cite{liu2015fast}. %\iw{pay attention to placement of references relative to punctuation}

One of the main challenges with \ac{DP} is the privacy--utility trade-off, which arises from the noise introduced to ensure privacy. 
To address this issue, several innovative approaches have been proposed to improve privacy without necessarily applying noise.
One such approach is recommendation unlearning, which is used when a recommender system needs to forget certain sensitive data and its complete lineage. 
``Influence function'' is an example to achieve recommendation unlearning without model retraining \cite{zhang2024recommendation}. %\iw{please do not start sentences with "In []". rephrase.}
While this method can reduce the risk, it cannot ensure that all sensitive information linked to a specific user is entirely eliminated.
%\iw{how is unlearning relevant to your work?}
%\iw{use correct latex quotes, i.e., `` and '' for opening and closing}
%\iw{figures should by default be placed at the top of the page. please update all}
\begin{figure}[t]
    \centering
    \begin{subfigure}{0.32\textwidth}
        \centering
        \includegraphics[width=\textwidth]{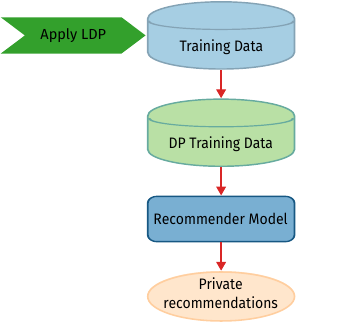}
        \caption{LDP}
        \label{fig:A_LDP}
    \end{subfigure}%
    \begin{subfigure}{0.32\textwidth}
        \centering
        \includegraphics[width=\textwidth]{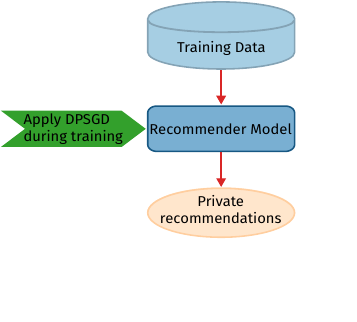}
        \caption{DPSGD}
        \label{fig:B_DPSGD}
    \end{subfigure}%
    \begin{subfigure}{0.32\textwidth}
        \centering
        \includegraphics[width=\textwidth]{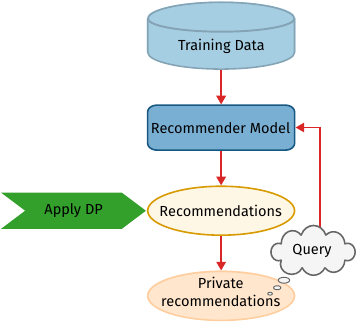}
        \caption{DP}
        \label{fig:C_DP}
    \end{subfigure}
    \caption{Three approaches for creating differentially private recommender systems: applying LDP to training data, applying DPSGD during model training, and applying DP to generated recommendation lists.}
    \label{fig:DPfamily}
    %\iw{this figure has no caption, and it is not referenced in the text. Also, please think about where the figure should be placed.}
\end{figure}

%However, even after implementing this approach, there is no assurance that all sensitive information of a specific user has been completely erased.
%Even some approaches that try to improve transparency can be harmfull for user's privacy.\iw{too colloquial}
%For example, as discussed in \cite {ge2024survey}, in social media, when a system tries to be explainable and explains that I recommend this because of your friends' interests, it can leak the privacy of friends.\iw{too colloquial}
Additionally, another approach focuses on limiting the amount of user interaction history. Therefore, rather than considering a user's entire history, a few carefully selected interactions are sufficient to achieve high personalization quality and thus improve privacy \cite{rendle2023reducing}.
However, reducing a user's history—such as removing ratings for certain items—can increase the system's vulnerability to data poisoning attacks \cite{shams2024evaluating}.
%\iw{when you write ``studies suggest'' or ``studies have demonstrated'': these phrases are too wordy, it is better to rewrite the sentence, removing this type of phrase. For example: ``A few carefully selected interactions are sufficient to achieve high personalization quality, and thus improve privacy \cite{rendle2023reducing}.''}
Similarly, focusing on user interactions rather than content or messages is a means to improve privacy \cite{tomlinson2023targeted}.
Recent research, however, indicates that various types of user-sensitive information—such as age, gender, occupation, and even political orientation—can be inferred from user-item interactions \cite{yuan2020parameterefficient}
Additionally, Federated Learning, as one of the promising approaches, can still leak sensitive user information if \ac{DP} mechanisms are not applied \cite{boenisch2023whena}.
%In \cite{aerni2024evaluations} highlighted that DPSGD provides better privacy protection even at high values of epsilon compared to several newly proposed privacy-enhancing approaches.
%They concluded that DP-SGD could be regarded as the "best-in-class" defense for practical applications.

Considering the challenges and solutions discussed, there are findings \cite{aerni2024evaluations} that DPSGD provides better privacy protection even at high values of epsilon, outperforming several newly proposed privacy-enhancing approaches. 
Therefore \ac{DPSGD} could be regarded as the best-in-class defense for practical applications \cite{aerni2024evaluations}.
%\iw{use acronym package to avoid inconsistent spellings of DPSGD vs DP-SGD. I have mentioned this before, please use this package consistently from now on.} could be regarded as the best-in-class defense for practical applications.
%\iw{this is all one paragraph, covering multiple topics. to the reader, this appears extremely unstructured. use one paragraph per topic to introduce some structure, and pay attention that the sequence of paragraphs is logical. you mix sentences about privacy issues with sentences about privacy-preserving systems, which is confusing because the title of the section only mentions systems, not issues}
%\iw{"it is suggested that", "has shown that", etc. are filler phrases. they do not carry meaning. please rewrite, removing these phrases}

In general, as Figure~\ref{fig:DPfamily} shows, there are three \ac{DP}-based methods.
One option is to introduce controlled noise to the training data before commencing the training process; a technique like randomized response, depicted in Figure \ref{fig:A_LDP}, is an example of this approach and is discussed in greater detail in Section \ref{sec:RSLDP}. 
Another widely-used method involves adding controlled noise during the training phase, which is particularly common for deep models, as shown in Figure \ref{fig:B_DPSGD}.
This approach is discussed in Section \ref{sec:pprs}
The final option, illustrated in Figure \ref{fig:C_DP}, entails training the model without any modifications and subsequently introducing noise to its outputs. 
However, this method has a significant drawback: the privacy loss compounds with each query made to the model, necessitating noise additions that are proportional to the number of queries.
As a result, this can lead to outputs that are effectively unusable.

\subsection{Bias in Recommender Systems}
%\iw{please rewrite with one paragraph per topic, also taking into account all above points}
%In the realm of recommendation systems (RS), there has been a historical emphasis on delivering precise recommendations that align with user preferences.
%However, contemporary research has recognized the significance of incorporating supplementary criteria in the evaluation of perceived quality and utility of recommendation lists.
%However, it is important to recognize that \ac{RS} can amplify different biases.
Recommender systems can amplify biases, including position bias, selection bias, popularity bias, and representation bias \cite{chen2023bias}.
%This can result in the creation of ranked lists that do not provide equal coverage to items throughout the whole popularity range \cite{yalcin2021investigating}. 
These biases can distort recommendation outcomes in various ways.
For instance, they can lead to ranked lists that do not provide equal coverage to items throughout the whole popularity range \cite{yalcin2021investigating}, reinforce filter bubbles \cite{nguyen2014exploring}, and skew user engagement \cite{abdollahpouri2020connection}.
%\iw{reference nguyen2014exploring is missing}
%\iw{we found this paper that presented a taxonomy of fairness/bias types. this may be a good start to this section and provide some structure to the discussion of bias}
%Numerous definitions for fairness have been offered in the field of recommender systems.
%In \cite{steck2018calibrated} if a user has liked 70\% romance and 30\% action movies, a fair recommendation list should contain similar distribution.
%Calibration is a regularization approach that is proposed to reach this purpose.
%Suitable regularization can push the model towards balanced recommendation lists \cite{chenBiasDebiasRecommender2023}.

Bias in recommender systems can be categorized in multiple ways, including group vs individual fairness and single-sided vs.multi-sided bias \cite{deldjoo2024fairness}.
Individual fairness refers to the principle that candidates with similar characteristics should be treated similarly \cite{deldjoo2024fairness}.
With group fairness, the goal is to achieve statistical parity between protected groups.
To evaluate group fairness, the users or items are divided into non-overlapping groups based on some attributes.
For example, users can be divided based on the types or categories of items they consume, and items can be divided based on their popularity.

Research on recommender systems often focuses on their value to  consumers, which corresponds to single-sided fairness \cite{deldjoo2024fairness}.
Multi-sided fairness considers fairness from the viewpoint of multiple stakeholders, such as consumers and item providers \cite{deldjoo2024fairness}.
What may be a fair recommendation for users is, in some way, unfair to the item provider.
Here, items are categorized according to their popularity, and users according to the popularity of items they consume.
%\iw{individual and single-sided are not explained}

%\iw{the following paragraph is very unstructured. it reads like a random list of papers, not like a well thought-out discussion of bias-reducing methods}
In general, approaches for solving bias can be categorized into three main types: those that use de-biasing as a post-processing method, those that incorporate bias mitigation into their objective function, and those that aim to create de-biased representation vectors. %\iw{this would be a better starting sentence for the paragraph}
%\iw{note how I rephrased the following two paragraphs to use active voice instead of passive. Try to apply similar rewrites to avoid phrases like "has been proposed" or "are used" or "is proposed"}
For example, calibration is a post-processing approach that ensures that recommendation lists reflect the distribution of user preferences \cite{steck2018calibrated}.
It achieves by pushing the model towards generating balanced recommendation lists \cite{chen2023bias}.
Similarly, re-ranking techniques can optimize relevance scores while simultaneously reducing both item and consumer bias \cite{rahmani2024personalized}.

To address bias as part of the objective function, off-policy learning combined with an additional cost function can improve item exposure fairness \cite{wang2024constrained}.
%In \cite{wang2024constrained}, as an example of addressing bias as part of the objective function, the authors aimed to improve item exposure fairness using off-policy learning combined with an additional cost function.
To enforce fairness, the Exact-K-Fairness constraint ensures that the proportion of item exposure between head-tailed and long-tailed groups remains statistically indistinguishable. %, within a maximum allowed difference of alpha.
Some methods \cite{li2023explicitly} focus on de-biased representation learning, for example, by making node representations fairer by alleviating the popularity bias in learning the representation of users and items, and tune neighbors' weights based on their popularity levels.
Additionally, when only biased feedback is available, it is possible to focus on mitigating confounding bias through debiased representation learning \cite{liu2023debiased}.

Several metrics help quantify bias in recommender systems. 
For example, higher entropy values indicate better item fairness, while the Gini index measures inequality in the item distribution \cite{deldjoo2024understanding}.
Additionally, the Herfindahl-Hirschman Index (HHI) assesses the concentration of recommendations among items \cite{deldjoo2024understanding}, and  head-tail ratio and the Gini index evaluate fairness \cite{wang2024constrained}.
%The HHI values range from 1/n1/n to 1, where nn represents the total number of items.\iw{this sentence and the next are perhaps too much detail here}
%Lower HHI values indicate better item fairness.
The normalized Gini index assesses the inequality and the average percentage of long-tail items to evaluate the presence of popularity bias in recommendations \cite{malitesta2024formalizing}. 
Additionally, item coverage evaluates how well the recommended items span the entire range of products \cite{malitesta2024formalizing}.
For considering both consumer and product fairness, a division categorizes items into two categories based on their popularity: short-head and long-tail items \cite{rahmani2024personalized}.
The deviation-from-parity exposure metric identifies whether short-head or long-tail items receive disproportionate exposure.
Similarly, a grouping separates users based on their activity levels.
The deviation-from-parity consumer fairness metric evaluates whether fairness holds across different user groups.
The limitation of the proposed metrics are that they are only applicable when there are two user groups.

Popularity bias occurs when popular items receive increased visibility in recommendation lists, leading to a subsequent increase in their consumption rate.
%Additionally, the head-tail ratio and the Gini index evaluate fairness \cite{wang2024constrained}.
Popularity Lift is a metric to evaluate the degree of amplification of popularity bias in the user profile and the generated recommendation lists for different user groups \cite{abdollahpouri2020connection}.
In addition to the approaches discussed earlier, methods such as novelty, miscalibration, and coverage are also used to evaluate the bias state of recommendation lists, which will be explained in detail in \cref{sec:metrics}. %\iw{references to other sections must include the word Section. Just putting the section number is insufficient. you either need to add "Section" manually (check ALL please), or you need to figure out how to use the cleveref package (again, check ALL). This also applies to references for figures and tables. please check them too.}

\subsection{The Privacy-Bias Trade-off in Recommender Systems}
%\iw{what is the point of this section in our methods chapter? I think it should be removed or partially integrated into related work.}
\ac{DP} ensures that algorithms behave similarly on any pair of databases differ only in the data of a single individual~\cite{dwork2014algorithmicc}. 
This definition aligns with fairness, which requires outcomes of the model to remain similar for similar individuals \cite{zemel2013learningb}.
%\iw{avoid starting sentences with "In \cite{}". Rephrase those sentences so that the focus is not on the reference, but on the substance of what the reference says}
Therefore, privacy can promote fairness or decrease bias, and fairness or debiasing, in turn, can enhance privacy.
Also, breaking the user's privacy can be disparately successful against minority groups in some cases \cite{miracle2016deanonymization}.
A violation of privacy can result in unfairness or bias because of an adversary's capability to infer sensitive information about individuals \cite{pessach2022review}.
Some methods proposed to reduce bias also enhance privacy \cite{ganhor2022unlearning}.

However, in general, there has been limited research on how privacy-enhancing methods impact various bias metrics \cite{mullner2023differentialc}.
In machine learning, privacy protections can negatively impact fairness, for example through a disproportionate reduction in model accuracy for underrepresented subgroups \cite{bagdasaryan2019differentialb}.
A recent study examined the correlation between differential privacy and bias in recommender systems~\cite{mullner2024impactb}.
%\iw{I do not want to read the word "authors" (we discussed this). please rephrase how you cite papers.} 
%However, they apply DP to user data rather than to the training process and only consider variants of deep recommender systems.
%Moreover, their analysis primarily considers the impact of DP on popularity bias.
That work applied DP to user data rather than the training process and focused exclusively on deep recommender architectures, with a particular emphasis on how DP influences popularity bias.
In contrast, in this paper, we examine how DPSGD affects utility and bias across four different classes of recommender systems.
Since DP-SGD enforces privacy through modifications to the training procedure, we additionally compare its effects with \ac{ldp} mechanism ~\cite{mullner2024impactb}.

\begin{figure}[t]
	\centering
	\includegraphics[width=0.6\linewidth]{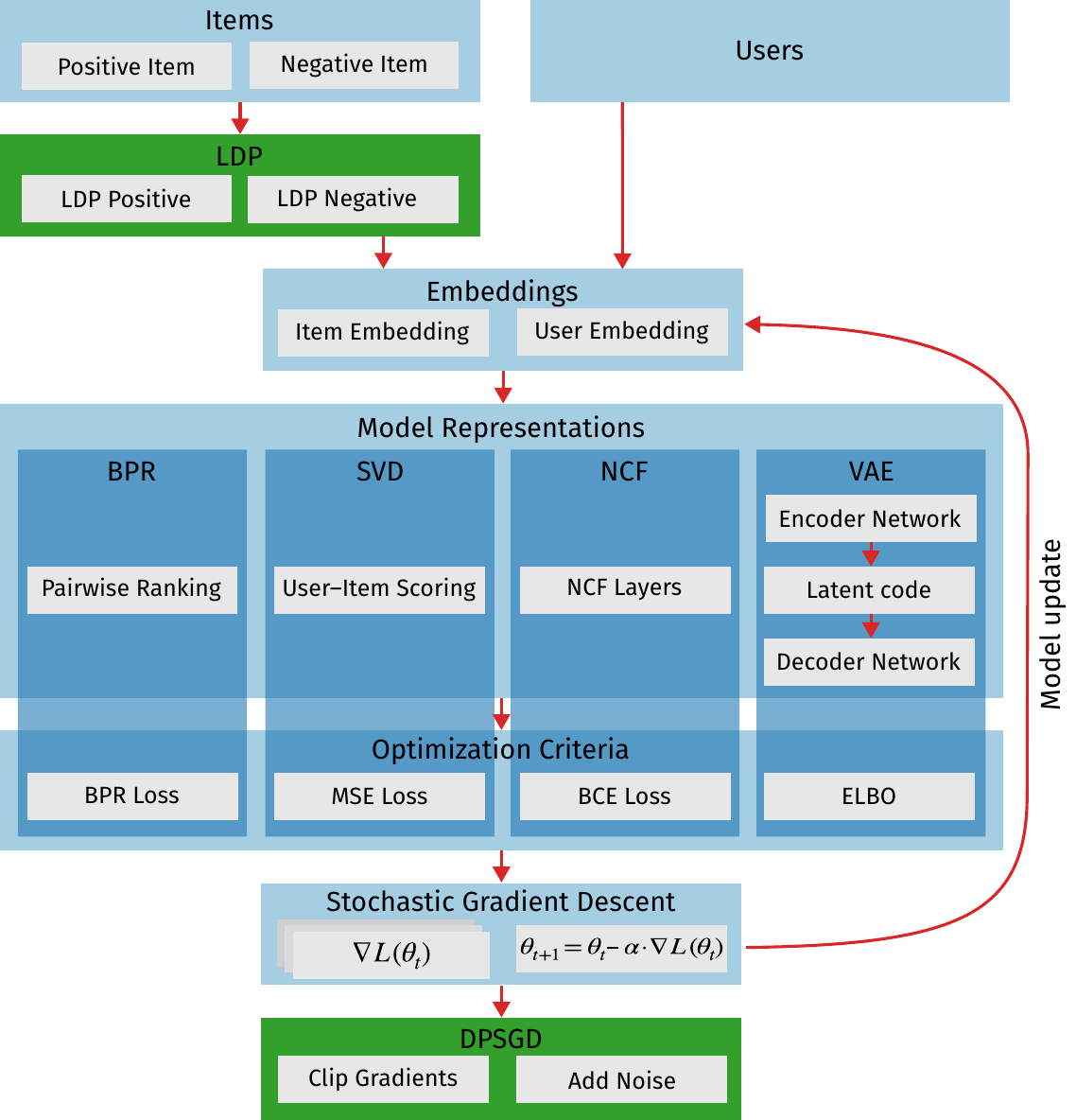}
	\caption{General Process of Applying LDP and DPSGD to Models}
	\label{fig:bpr-plot-apply-dpsgd}
	%\iw{what is the justification for only showing DPSGD/BPR, but none of the other combinations? Could we update the figure to show some the other combinations too (and LDP in addition to DPSGD?) Level of detail in this figure vs Figure 1 is very different.}
\end{figure}

\section{Methods}

Our main goal is to study the trade-offs between privacy, performance, and bias in recommender systems.
We perform this study on four different types of recommender systems, each in a private and non-private version (Section \ref{sec:pprs}), applied to two datasets (Section \ref{sec:datasets}), by using a range of metrics (Section \ref{sec:metrics}) and experimental settings (Section \ref{sec:expsettings}).
Table \ref{tab:notation} summarizes the notation used in this paper.
% using Differential Privacy Stochastic Gradient Descent (DPSGD).
%The objective is to determine the most effective settings in which DPSGD improves privacy while minimizing the adverse effects on the system's predictive accuracy.
%Additionally, we aim to explore whether the implementation of DPSGD influences bias metrics within these models, further assessing its broader implications on the fairness and integrity of recommendat------ion outcomes

\subsection{Privacy-preserving recommender systems}
\label{sec:pprs}
To enhance privacy in recommender systems, two primary approaches leverage differential privacy: \ac{ldp} and \ac{DPSGD}.
\ac{ldp} protects user data at collection by adding randomness, while \ac{DPSGD} ensures privacy during model training by clipping gradients and injecting noise.
%\iw{need an introductory sentence to mention the two approaches and how they differ (on a high level)}
Figure \ref{fig:bpr-plot-apply-dpsgd} shows the integration of \ac{ldp} and \ac{DPSGD} at different stages of the recommendation process.
User and item embedding updates are treated as optimization problems, which enables the application of \ac{DPSGD}.
Figure \ref{fig:bpr-plot-apply-dpsgd} differentiates where each privacy method is applied.
\ac{ldp} is applied before the embedding stage, specifically on the item interaction data.
This means that items are perturbed locally on the client side before they are embedded, resulting in noisy or DP-based embedding representation. 
In contrast, \ac{DPSGD} is applied during the training process.

Each model block in the Figure \ref{fig:bpr-plot-apply-dpsgd} shows its unique approach to generating the recommendation list. 
For example, \ac{NCF} uses a stack of neural network layers (NCF layers) to learn interactions between user and item embeddings.
Similarly, \ac{BPR} and \ac{SVD} rely on pairwise ranking and matrix factorization mechanisms, respectively.
For SVD, we represent users and items using latent embedding vectors and train these embeddings with SGD to facilitate the application of DPSGD.
For BPR, the process of updating user and item embeddings is expressed as a stochastic optimization problem similar to SGD, which iteratively adjusts user and item embeddings.
In each training step, given a positive user–item interaction, the model updates the embeddings to improve the ranking of positive interactions over negative ones. 
\ac{VAE} adopts an encoder-decoder structure with latent variables for probabilistic modeling.
Each model is trained by minimizing its own loss function: \ac{MSE} Loss for \ac{SVD}, BPR Loss for \ac{BPR}, \ac{BCE} Loss for \ac{NCF}, and ELBO for \ac{VAE}.
These loss functions serve as the optimization criteria guiding model updates.
When DPSGD is applied, it modifies this optimization process by introducing noise and clipping gradients, making privacy a built-in part of training rather than a pre-processing or post-processing step.

\subsubsection{Recommender Systems with Local Differential Privacy}
%\iw{need an introductory sentence before jumping into the definition}
\label{sec:RSLDP}
	
One way to apply \ac{DP} to recommender systems is randomized response \cite{mullner2024impactb}.
%The randomized response is one of the well-known %\iw{too colloquial}
%mechanisms of Differential Privacy. %\iw{is randomized response at all relevant to the use of DP in recommender systems?}
Randomize response flips a fair coin to decide whether to report the true response or the opposite \cite{dwork2014algorithmicc}.
Inspired by the 1-Bit mechanism for mean estimation \cite{ding2017collecting}, a Local-DP mechanism can be used to randomize the selection of positive feedback in recommender systems \cite{mullner2024impactb}.
Local-DP mechanism protects user privacy in systems that rely on binary implicit feedback by introducing randomness into the feedback selection process.
Two types of feedback are considered:
\begin{itemize}
    \item \textbf{Positive feedback} ($ D^+ $): Interactions where a user expresses a preference for an item (e.g., a click, like, or purchase).
    \item \textbf{Negative feedback} ($ D^- $): Instances where the user does not interact with an item or provides no feedback.
\end{itemize}
In scenarios with explicit ratings, items can be classified into positive and negative feedback categories using a threshold on the ratings.
Items with ratings exceeding the threshold are categorized as $ D^+ $, while all other items are treated as $ D^- $.
The probability $ P_{D^+} $
%\iw{please be consistent in your use of symbols! this probability has a different symbol in the equation below.}
of including a positive feedback item in the differentially private positive dataset is given by:

\begin{equation}
	P_{D^+} = \frac{1}{e^\epsilon + 1} + \frac{e^\epsilon - 1}{e^\epsilon + 1}
\end{equation}

%This formula determines the probability that a positive feedback $ f_{u,i} $ (where user $ u $ interacts with item $ i $) will be included in the differentially private dataset.
%This probability is designed to balance privacy protection and data utility. 
The first term introduces randomness, while the second term adjusts this randomness based on the privacy budget, where 
%The overall probability will always be between 0 and 1, with 
higher values of $\epsilon$ lead to a higher chance of including positive feedback, resulting in a dataset that is less private but more accurate.

Similarly, the probability that a negative or missing feedback item will be included in the new DP positive dataset is calculated as follows:
\begin{equation}
	P_{D^-} = \frac{1}{e^\epsilon + 1}
\end{equation}
This probability is lower than the probability for positive feedback, ensuring that negative or missing feedback data is added with lower likelihood.
%\iw{most of this section does not belong in a section titled "Differential Privacy", but rather in a section titled "Recommender Systems with Local Differential Privacy"}
In this approach, only positive feedback is perturbed, as it represents explicit user preferences (e.g., what a user explicitly likes or purchases) that are more likely to reveal personal information.
Because negative feedback is implicit (e.g., lack of interaction) and less informative, it may not require the same level of perturbation to achieve privacy guarantees.

\subsubsection{Recommender Systems with Differential Private Stochastic Gradient Descent} %Differentially Private Stochastic Gradient Descent}

\ac{DPSGD} is a technique to ensure privacy protection during the training process of deep learning models \cite{abadi2016deep}.
And is therefore a natural fit for deep recommender systems.
Consider a user $ u $ with $ k $ positive interactions represented as $ u_p = {x_1, x_2, \dots, x_k}$.
%$\iw{where does this notation come from? is it standard notation? if so, give the reference. if not, please change it! $xi$ as a single variable name is extremely confusing, the $i$ should be a sub-/superscript}, where $ i \subseteq I $ and $ I $ is the set of of all $ N $ items.
Suppose we have a recommender system designed to optimize a loss function $L$, such as \ac{MSE}, and \ac{BCE}.
%\iw{give 1-2 examples}.
The primary goal of such systems is to predict items that a user is likely to be interested in based on their past interactions $ u_p $.

Using \ac{SGD} on a recommender system involves iteratively minimizing the loss function $L$.
%During each iteration, gradients are computed based on a batch of training examples, aggregated, and used to update the model parameters accordingly.
%\iw{define acronyms using the package and always use one of the commands from the package (\ac{SGD} or \acf{SGD}), never type the acronym itself.}, 
\ac{DPSGD} introduces a privacy-preserving approach to this process \cite{abadi2016deep}. %\iw{punctuation}
At each iteration, a batch of user interactions $ x_i $ is sampled from the training dataset.
This batch includes a subset of positive interactions, and for some models, both positive and negative interactions $ u_p $ from each user. %\iw{it is not clear how mini-batches are different from the batches in SGD, and why the distinction is important.}
The size of the batch is typically fixed and is determined based on the system's capacity and the desired trade-off between computation and privacy.
%For each user interaction $ x_i $ in the mini-batch, the gradient of the loss function $ L(\theta, x_i)$
%$\iw{please check all your equations. they do not seem to conform to standard latex equation typesetting. here, the brackets are missing from the PDF.} 
%with respect to the model parameters $ \theta $ is computed.%\iw{here, $x_i$ is a data point, above, $xi$ was a positive interaction. this is too confusing for readers, please pay attention to the symbols you use.}
%This gradient tells us how to adjust the model parameters to minimize the loss.\iw{make sure to focus on the differences between DPSGD and SGD. this point is exactly the same for the two, so no need to mention it here.}
To control the influence that any single data point has on the model, each gradient $ \nabla_\theta L(\theta, x_i) $ or $g$ is clipped by using the $ \ell_2 $-norm.
Specifically, the gradient is rescaled to ensure that its $ \ell_2 $-norm does not exceed a predefined threshold $ C $.
The clipping norm $ C $ determines the maximum allowed $ \ell_2 $-norm of the gradients.

%\iw{update to match notation in abadi paper}

\begin{equation}
	\bar{g} (x_{i})= \nabla_\theta L(\theta, x_i) / \max\left( 1, \frac{\|\nabla_\theta L(\theta, x_i)\|_2}{C} \right)
\end{equation}

This step prevents any individual example from disproportionately affecting the model.
Once the gradients for each data point in the batch are computed and clipped, the average of the gradients is calculated:
%Once the gradients for each data point in the batch are computed and clipped, they are averaged and used to update the model parameters:
%\iw{the order of actions in your explanation is wrong. noise addition is explained after update to model parameters is mentioned, but it needs to come before.}
\begin{equation}
	\bar{g} = \frac{1}{B} \sum_{i=1}^B \bar{g}(x_{i})
\end{equation}
where $ B $ is the size of the batch.
To ensure differential privacy, noise is added to the aggregated gradient.
The noise is drawn from a Gaussian distribution with mean zero and a variance scaled by $ \sigma^2 C^2 $.
The amount of noise is controlled by the noise multiplier $ \sigma $, which governs the trade-off between privacy and utility.
\begin{equation}
	\tilde{g} = \bar{g} + \mathcal{N}(0, \sigma^2 C^2 I)
\end{equation}

Finally, the model parameters $\theta$ are updated in the opposite direction of the noisy gradient.
The update is performed by subtracting the noisy gradient scaled by the learning rate $ \eta $:

\begin{equation}
	\theta = \theta - \eta \cdot \tilde{g}
\end{equation}
The algorithm repeats the process for a set number of iterations or until convergence, where convergence is typically determined by the stability of the loss function or parameter updates.
%This process ensures that, even though the model is trained using user data, the privacy of individual users is protected by the noise addition and gradient clipping.
The privacy cost of the entire training process is tracked by a privacy accountant, such as the moments accountant \cite{abadi2016deep}, Rényi Differential Privacy (RDP) accountant, or Privacy Random Variables (PRV) accountant \cite{gopi2021numerical} that tracks the privacy loss more accurately than RDP or the classical moments accountant.
%, all of which offer tighter or more flexible accounting methods.
These techniques ensures an overall privacy bound of $O(q\epsilon T, \delta)$-differential privacy, where $T$ represents the number of gradient descent steps, and $q$ the batch sampling probability.

\subsection{Datasets}
\label{sec:datasets}

\begin{table}[t]
	\centering
	\begin{tabular}{lcc}
		\toprule
		 & \textbf{Yelp} & \textbf{MovieLens 1M} \\
		\midrule
		Total items & 7,156 & 3,258 \\
		Total users &  9,872& 6,038\\
		\# of interactions &  175,767& 835,614\\
		\% of interactions & 0.24\%& 4.22\%\\
		\# of item categories &1030  &  19\\
		Min categories per item & 1 & 1 \\
		Max categories per item &  25& 7 \\
		Mean categories per item & 4.36 & 2.11 \\
%		Median categories/item & 4.0 & 2.0 \\
		\bottomrule
	\end{tabular}
	\caption{Comparison of item statistics between Yelp and MovieLens 1M datasets.}
	\label{tab:genre_stats}
\end{table}

The MovieLens 1M dataset has a total of 1,000,209 ratings provided by 6,040 individuals for 3,706 movies \cite{harper2016movielens}.
Movies are categorized in 20 genres.
%, and ratings are in the range (1-5).
%MovieLens is often used as a benchmark dataset in the research community.
% 
%\todo[inline]{how many genres are there?}
The Open Yelp dataset contains 6,990,280 reviews for 150,346 businesses, categorized in 1030 business types.
%Open Yelp dataset is a subset of Yelp data intended for educational purpose.
% and aggregated check-ins over time for each of the businesses.
%\todo[inline]{is this the entire Yelp dataset, or just the part focusing on restaurants? we need to say that we select only businesses that are restaurants.}
%For the calibration purpose we consider categories as our calibration feature.\todo[inline]{we don't do calibration, so this sentence is out of place.}
%The total number of features on Yelp is 591,  a considerable number that makes it unsuitable for the purpose of calibration.
%We consider major categories as our calibration feature.
%We select businesses in the \textit{restaurant} category, which is categorized in 313 restaurant types.
Both Open Yelp and MovieLens 1M use ratings between 1 (lowest) and 5 (highest).
%Yelp includes a total of 22 primary categories.
%We map each category to its root or major category. 
%Then, we only consider the restaurant category.
%And resturan category is used to calculate miscalibration.

%\iw{fix structure of following paragraph (it is too long and covers multiple topics). also, popular items are defined twice.}
We clean both datasets in line with prior work.
%\iw{please make spelling consistent: yelp vs Yelp}
In particular, we exclude all ratings below three, all items with less than five ratings, and all users who have rated less than five items \cite{he2017neural,curmei2023private}.
In addition, following the common practice of restricting the Yelp dataset to specific regions \cite{leskovec2010yelp}, we limit our analysis to businesses in Arizona (AZ area).
After preprocessing, we have 835,614 ratings in the 1M dataset (3,258 movies and 6,038 users) and 175,767 ratings in the Yelp dataset (7,156 businesses and 9,872 users).
Table \ref{tab:genre_stats} shows the statistics of these two datasets after all pre-preprocessing steps. 
Compared to 1M, the Yelp dataset is significantly more sparse.
Despite having more users and items, Yelp contains fewer overall interactions, which results in an interaction density of just 0.24\%, in contrast to 4.22\% in 1M.
Moreover, Yelp includes a much larger and more diverse set of item categories (1,030 vs. 19), with a higher average number of categories per item (4.36 vs. 2.11).

%Yelp includes 313 distinct restaurant styles, in contrast to the 20 genres found in MovieLens.
%We split the datasets into training (80\%) and testing (20\%) parts on a \textit{per-user} basis.

We employ an 80/10/10 data split for training, testing, and validation, splitting on a \textit{per-user} basis.
%\iw{please use present tense throughout.}
This ensures that each user in the test/validation sets has a history in the training set. 
%\todo[inline]{how? please be more specific: if I remember correctly, the split was not done fully randomly, so please explain.}
%To ensure the possibility of calibration in our recommender system, the partitioning of the dataset into training and test sets was strategically designed to maintain a user history in the training data for each user present in the test set.
%This approach guarantees that every user evaluated in the test phase has corresponding historical data in the training set.
%\todo[inline]{We made those changes in order to make our dataset suitable for calibration. But in this paper, we do not have a calibration part; should I explain it?}
%\todo[inline]{what does "in a user's profile" mean? all items that the user has rated, or items that the user has rated highly?}

\subsubsection{User types}
%To categorize users and items into different groups, we need to know the frequency or popularity of items. 
We divide users into three categories -- niche, blockbuster, and diverse -- based on the popularity of items they have rated \cite{abdollahpouri2020connection}.
Items are called \textit{popular} if they are in the top 20\% of items sorted by popularity.
%%\iw{I would introduce item popularity before user categories, then we do not have to repeat the definition of popular}
Niche users prefer unpopular items, i.e., less than 50\% of items in their profiles are popular, whereas blockbuster users have more than 85\% of popular items in their profiles.
All other users are called diverse.
The distribution of user types (see Figures \ref{fig:MovieLens1M} and \ref{fig:yelp}) shows that the Yelp dataset contains significantly more niche users, and fewer blockbuster users, than 1M.

\subsubsection{Item popularity}
\begin{figure}[t]
	\centering
	\begin{subfigure}[t]{0.40\linewidth}
		\includegraphics[width=\textwidth]{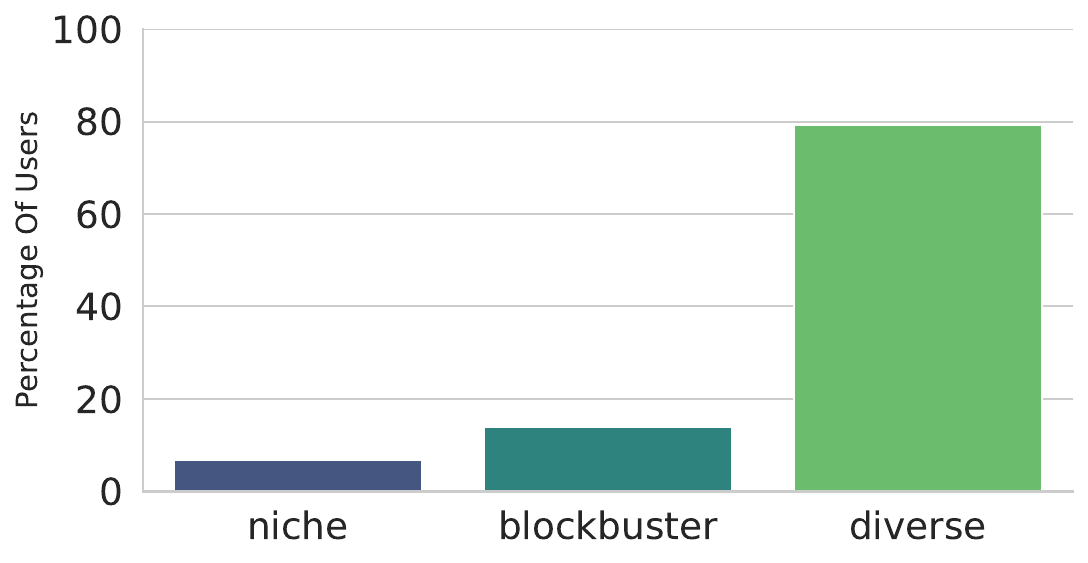}
		\caption{Distribution of user types in MovieLens 1M}
		\label{fig:MovieLens1M}
	\end{subfigure}
	\hspace{0.05\linewidth}
	\begin{subfigure}[t]{0.40\linewidth}
		\includegraphics[width=\textwidth]{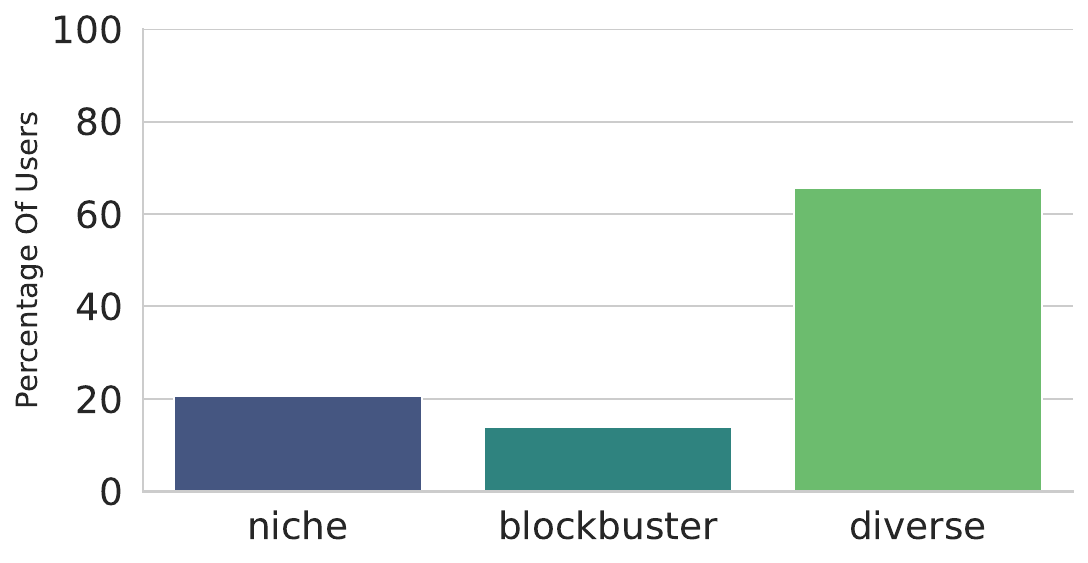}
		\caption{Distribution of user types in Yelp}
		\label{fig:yelp}
	\end{subfigure}
	%\iw{make captions consistent with what the figure shows: number of users or percentage? also, the figure shows a percentage of users, not a percentage of user groups}
  
	\begin{subfigure}[t]{0.40\linewidth}
		\includegraphics[width=\textwidth]{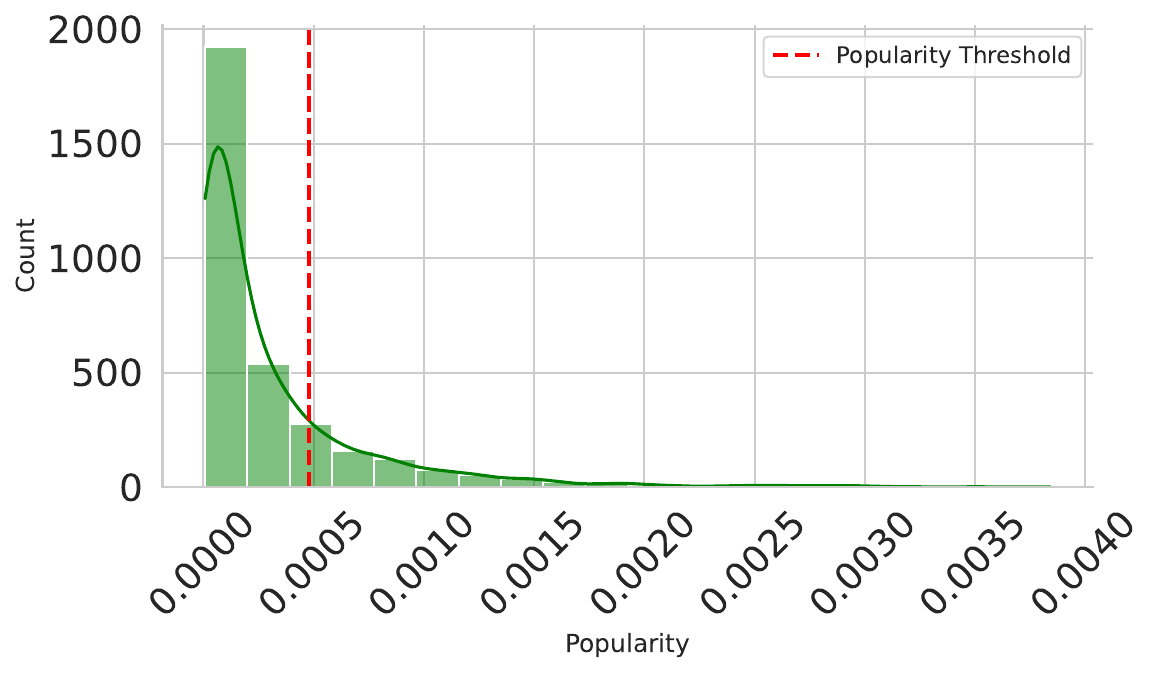}
		\caption{Distribution of item popularity in MovieLens 1M}
		\label{fig:dist:1M}
	\end{subfigure}
	\hspace{0.05\linewidth}
	\begin{subfigure}[t]{0.40\linewidth}
		\includegraphics[width=\textwidth]{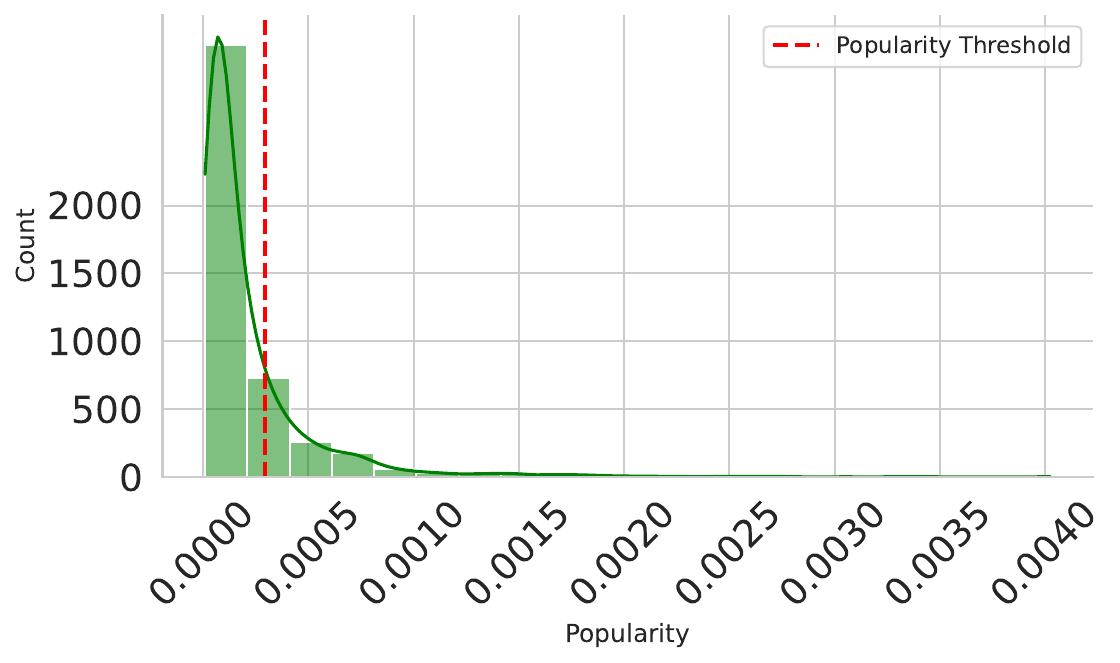}
		\caption{Distribution of item popularity in Yelp}
		\label{fig:dist:Yelp}
		%\iw{use the same figures size as for subfigures a and b, same size of axis labels, despine(), etc}
	\end{subfigure}
	
	\caption{ Distribution of user types and item popularity in Yelp and MovieLens  datasets}
	\label{fig:movie+yelp}
  \end{figure}

In addition to examining different user categories, we also evaluate the behavior
of recommender systems on two categories of items: the short head and the long tail, which are defined based on item popularity \cite{abdollahpouri2019unfairnessa,rahmani2024personalized}.
%We divide items into two categories, $I_1$ or short head and $I_2$ or long tail, defined based on the frequency of being rated by users or their popularity among users. 
Each item belongs exclusively to either $I_1$ or $I_2$, so $I_1 \cap I_2 = \emptyset$.
The top 20\% of items are classified as $I_1$, while the remaining 80\% are considered as long-tail or unpopular items ($I_2$).
By categorizing the items into short head and long tail, we can assess the performance of recommender systems in terms of how they handle the recommendation of popular versus unpopular items.
Figures \ref{fig:dist:1M} and \ref{fig:dist:Yelp} show the popularity distribution of items in each dataset.
The majority of items are to the left of the threshold line, classified as unpopular items $I_2$, while those to the right are considered as $I_1$.
%Figure \ref{fig:movie+yelp} shows the majority of items fall into the unpopular category.

%\todo[inline]{Explain about different items}

\subsection{Metrics}
\label{sec:metrics}
%\iw{need introductory sentence: what do we need metrics for?}
%\todo[inline]{Add coverage and novelty and DPF}
We use a set of metrics to comprehensively evaluate utility, bias, and privacy of \ac{RSs}.
%The comprehensive evaluation of a recommender system is a multifaceted and intricate process. 
Our selection ensures that the evaluation covers different aspects of how an RS performs \cite{zangerle2022evaluating} and includes the most common metrics in \ac{RS} evaluation~\cite{bauer2024exploring}.
%Here, we investigate different metrics to cover various dimensions of recommender systems. 
%Our different metrics, such as Novelty, Coverage, and NDCG, are among the most used metrics to evaluate \ac{RS} \cite{zangerle2022evaluating,bauer2024exploring}.

\begin{table}[ht]
	\centering
	\caption{Notation used throughout this paper}
	\begin{tabular}{cp{11.3cm}}
	\textbf{Symbol} & \textbf{Meaning} \\
	\midrule
	$\epsilon$ & Privacy Budget \\
	$\delta$ & Probability of the privacy leakage \\
	$D^+$ /$D^-$  & Positive feedback/ Negative feedback \\
	$P_{D^+}$/ $P_{D^-}$ & Probability of including a positive/negative feedback item in the new \ac{DP} positive dataset \\
	$C$ & clipping norm  \\
	$ \sigma$ & Noise Multiplier \\
	$ \theta$ & Model weights\\
	%$ \nabla_\theta$ & Gradient of $\theta$\\
	$\bar{g}$ & Average of Gradient\\
	$B$ & Batch Size\\
	$L(\theta, x_i)$ & Loss function evaluated for input $x_i$ and model parameters $\theta$ \\
	$I$, $I_{1}$, $I_{2}$& Set of the items, Popular Items, Unpopular Items \\
	$q(z|u)$ & Distribution of genres/business-styles $z$  in the list of movies/businesses recommended to user $u$ \\ 
	$p(z|u)$ & Distribution of genres/business styles $z$ in the set of movies/businesses $H$ previously rated by user $u$\\
	$nDCG_{p}$   & Normalized Discounted Cumulative Gain \\
%	$rel_{i}$ & Relevance score of the item at position $i$ \\
	%$iDCG_{p}$ & Ideal DCG \\
	%$nDCG_{p}$& Normalized Discounted Cumulative Gain\\
%	$r_{i}$ & Rank of item i in the recommendation list\\
%	$w_{u(i)}$ & The weight of item i based on its rank $r_{i}$ in the recommendation list \\
	$C_\text{KL}(p, q)$ & Miscalibration based on KL divergence\\ 
	$GAP(G)$ & Group Average Popularity\\
	$PL(G)$ & Popularity Lift\\
	$\pi_{i}$ & Popularity of item i \\
	$G$ & Users' group, such as niche or blockbuster \\
	$R$ & The recommended list \\
%	$|R|$ & the length of the recommended list \\
	$Novelty (R)$ & Novelty of $R$ recommended items\\
	$Coverage$ & Percentage of total items $I$ that are recommended \\
	$DPF$ & Deviation from Producer Fairness \\
%	$\text{Item\_Exp}(i, L)$ &  Quantifies binary item groups' exposition\\
	\bottomrule
	\end{tabular}
	\label{tab:notation}
	\end{table}

\subsubsection{Utility Metrics}

We use Recall, Normalized Discounted Cumulative Gain (NDCG@10), and Mean Reciprocal Rank (MRR) to evaluate the ranking quality, or utility, of recommender algorithms.
However, due to the similarity in trends across these metrics, we only report results for NDCG. 

%\iw{the heading does not correspond to what the sentence says}
\paragraph{Normalized Discounted Cumulative Gain}
NDCG measures how closely the predicted ranking of items aligns with the ideal ranking, taking into account the relevance of each item \cite{jadon2025comprehensive}.
The DCG for the first \(p\) items is given by:

\[
DCG_p = \sum_{i=1}^{p} \frac{2^{\mathrm{rel}_i} - 1}{\log_2(i + 1)}
\]
where \(\mathrm{rel}_i\) is the relevance score of the item at position \(i\), and \(p\) is the number of ranked items.
The Normalized Discounted Cumulative Gain is obtained by normalizing \(DCG_p\) with the ideal DCG (\(iDCG_p\)).
%
%\[
%nDCG_p = \frac{DCG_p}{iDCG_p}
%\]
%
A perfect ranking results in \(nDCG = 1\), with any deviation leading to a score less than 1.
%For simplicity, we refer to this metric as NDCG throughout the rest of the paper

%\paragraph{Mean Reciprocal Rank (MRR)}
%MRR \cite{jadon2025comprehensive} is a metric used to evaluate the performance of recommendation or information retrieval systems, focusing on the rank of the first correct answer.
%For a set of \(Q\) queries, MRR is calculated as:

%\[
%MRR = \frac{1}{|Q|} \sum_{i=1}^{|Q|} \frac{1}{\text{rank}_i}
%\]

%where \(|Q|\) is the number of queries and \(\text{rank}_i\) is the position of the first relevant item for the \(i\)-th query.

%\paragraph{Recall}
%Recall@k \cite{jadon2025comprehensive} is a metric used to evaluate recommendation systems based on the proportion of relevant items found in the top-\(k\) recommendations.
%It is calculated as:

%\[
%\text{Recall@k} = \frac{|\{ \text{Relevant items} \} \cap \{ \text{Top-k recommended items} \}|}{|\{ \text{Relevant items} \}|}
%\]

%where \(|\{ \text{Relevant items} \}|\) is the total number of relevant items, and \(|\{ \text{Relevant items} \} \cap \{ \text{Top-k recommended items} \}|\) is the number of relevant items found in the top-\(k\) recommendations.
%\iw{we could drop the details for MRR and recall because we do not show results for them}

%\iw{please do not use \textbf{} to create fake unnumbered section headings. instead, think about a logical grouping of metrics and use a subsubsection for each group}

%\subsubsection{Miscalibration}
\subsubsection{Bias Metrics}
%\iw{add an introductory sentence for bias metrics}
To evaluate biases in recommender systems, we used five different bias metrics to assess different aspects of bias: group fairness (miscalibration), popularity bias, novelty, item coverage, and item fairness (deviation from producer fairness).
%The following sections provide a detailed introduction to the metrics that we used.

\paragraph{Miscalibration}
Miscalibration measures to what extent the item category distribution in a user's history deviates from the distribution in the recommended list.
Miscalibration occurs when this alignment is poor, indicating that the recommender fails to reflect the user's true distribution of interests.
Miscalibration can be seen as a measure of group fairness, even if item categories do not explicitly represent protected attributes~\cite{deldjoo2024fairness}.
%Calibration evaluates if different groups experience different levels of miscalibration in their recommended lists~\cite{deldjoo2024fairness}.

We use the Kullback-Leibler Divergence (KLD) to measure the miscalibration of recommended item lists \cite{steck2018calibrated}.
In general, KLD is a measure for the distance between two distributions.
Here, we consider the distribution of genres/business-styles in the list of recommended items and in the user's rating history.
%\todo[inline]{add one sentence to explain the intuition why KL measures miscalibration. Readers probably know that KL expresses the distance between two distributions, so you can say what the two distributions are.}
%This formula is proposed by \cite{steck2018calibrated}.
The distribution of genres/business-styles $z$ in the list of movies/businesses recommended to user $u$ is denoted as $q(z|u)$:
\begin{equation}
	q(z \mid u) = \frac{\sum_{i \in R} w_{u(i)} p(z\mid i)}{\sum_{i \in R} w_{r(i)}}
\end{equation}

Similarly, the distribution of genres/business styles $z$ in the set of movies/businesses $H$ previously rated by user $u$ is denoted as $p(z|u)$:
%%\todo{missing reference}
\begin{equation}
	p(z \mid u) = \frac{\sum_{i \in H} w_{u,i} p(z\mid i)}{\sum_{i \in H} w_{u,i}}
	\end{equation}

$I$ is the set of recommended items, and $w_{u(i)}$ is the weight of item i based on its rank $r_{i}$ in the recommendation list.   
And finally, the calibration metric $ C_\text{KL}$ based on KL divergence is calculated as follows:
\begin{equation}
   C_\text{KL}(p, q) = \text{KL}(p \| \tilde{q}) = \sum_{g} p(z|u) \log \frac{p(g|u)}{\tilde{q}(z|u)} 
\end{equation}
%\iw{$\tilde{q}(g|u)$ is never defined. CKL is never defined either.}
Where $w_{u,i}$ is the weight of movie $i$, e.g., how recently it was played by user $u$. 
If $q(g|u)$ and $p(g|u)$ are similar, $C_{KL}$ will take on small values.
Given that th $KL$ divergence becomes undefined when $q(z|u)=0$ and $p(z|u)>0$ for a genre/business z, we instead use:
\begin{equation}
	\tilde{q}(z\mid u) = (1 - \alpha) \cdot q(z \mid u) + \alpha \cdot p(z \mid u)
\end{equation}
	
By choosing a small and positive $\alpha$, we ensure that $q \approx \tilde{q}$ while avoiding zero probabilities in $q$.
In our experiments, we set $\alpha=0.01$ \cite{steck2018calibrated}.

%\subsubsection{Popularity Lift}
\paragraph{Popularity Bias} To measure popularity bias, popularity lift $PL$ expresses to what extent the popularity of items in a recommended list of items differs from the popularity of items in a user profile~\cite{abdollahpouri2020connection,naghiaei2022unfairness}, here a positive $PL$ indicates that the algorithm amplifies popularity bias.
%measures is used to examine the degree of amplification of popularity bias in the user profile and the generated recommendation lists for different user groups\cite{abdollahpouri2020connection}.
Popularity lift is computed by comparing the Group Average Popularity $GAP(G)$ items in a user's profile $p$ and items in a recommended list $q$:
\begin{equation}
PL(G)=\frac{G A P_{q}(G)-G A P_{p}(G)}{G A P_{p}(G)}
\end{equation}
%\iw{please fix typesetting of equations}
The $GAP(G)$ takes into account the popularity $\pi_{i}$ of all items $i$ in the item lists for all users $u$ in a group $G$ (such as niche or blockbuster) \cite{naghiaei2022unfairness}:
%, with $\theta(i)$ indicating the popularity of a specific item :
\begin{equation}
GAP_{p}(G) = \frac{\sum_{u \in G} \frac{\sum_{i \in p_u} \pi_{i}}{|p_u|}}{|G|}
\end{equation}

\paragraph{Novelty}
Novelty evaluates how unexpected the recommended items are to users.
If a recommender system heavily relies on a user's past behavior, it may recommend items that are similar to those that the user interacted with before.
This can create a feedback loop where users are repeatedly exposed to similar content, reinforcing existing preferences and potentially leading to bias.
The concept of novelty refers to having novel elements in the recommended list or the average information surprisal in the recommended item list \cite{wagner2018technical}.
%A novel item is an item unfamiliar to the user \cite{ricci2011recommender} or about which the user knows little or has limited information \cite{ge2010accuracy}. 
Using popularity as a basis for calculating novelty \cite{zhang2012auralist,zhou2010solvinga,dasilva2021exploiting}, novelty is the average popularity of the N top items:
\begin{equation}
	\text { Novelty }(R)=\frac{1}{|R|} \sum_{i \in R} POP_{i}=\frac{1}{|R|} \sum_{i \in R}-\log (\pi_{i})
	\label{nn}
	\end{equation}
In the equation \ref{nn}, $|R|$ represents the length of the recommended list, and $\pi_{i}$ represents the popularity of item $i$, i.e., the fraction of users who have rated it positively (for example, ratings greater than a threshold value).
%\iw{the notation seems to be a mess. popularity here is  $pop(i)$. in the previous equation it is $\theta(i)$. we do need to use consistent notation throughout this paper. Please create a notation table that lists all symbols in the paper and their meaning.}

%\iw{link this to information surprisal: novelty = average information surprisal in the recommended item list \cite{wagner2018technical}}
	
\paragraph{Coverage}
%To assess item distribution, we report on Coverage and Exposure, which measure how broadly the recommender system utilizes the item space. 

Coverage in a recommender system refers to the percentage of items from the total item set $I$ that contained in the list of items $R$ recommended by the system \cite{ge2010accuracy}. Formally:
%\iw{what does it mean to be "successfully recommended"? how would a recommendation be unsuccessful?}
\begin{equation}
\text{Coverage}(I, R) = \frac{\left| \{i \in I : i \in R\} \right|}{|I|}
\end{equation}
%\iw{why the $\times 100$? the y axes for our coverage figures report a number in $[0,1]$, not in $[0,100]$}

To calculate coverage, we consider the entire set of items $ I $ that the recommender system could recommend.
Coverage evaluates the items that appear in the combined recommendation lists across all users, then calculate the fraction of $ I $ represented by these recommended items.

\paragraph{Deviation from Producer Fairness}
To evaluate producer or item fairness, we use Deviation from Producer Fairness (DPF) \cite{rahmani2024personalized}.
Intuitively, DPF measures whether popular (short-head) items and unpopular (long-tail) items are getting equal visibility in the recommendation lists.
If both groups are shown equally, DPF is close to zero, meaning fair exposure between the two item groups, while positive (negative) DPF means over exposure of popular (unpopular) items.

\begin{equation}
	\text{DPF}(I_1, I_2, R) = \text{Coverage}(I_1, R) - \text{Coverage}(I_2, R)
\end{equation}

%\begin{equation}
%	\text{DPF}(R, I_1, I_2) = \mathbb{E} \left[ \text{Item\_Exp}(i, R) \mid i \in I_1 \right] - \mathbb{E} \left[ \text{Item\_Exp}(i, R) \mid i \in I_2 \right]
%	\end{equation}
%
%In this equation, $\mathbb{E} \left[ \text{Item\_Exp}(i, R) \mid i \in I_1 \right]$ quantifies the average binary item groups' exposure in the recommendation list $R$, and it is calculated as follows:
%%\iw{exposition or exposure?}
%
%\begin{equation}
%    \mathbb{E} \left[ \text{Item\_Exp}(i, R) \mid i \in I_1 \right] = \frac{\text{Number of items from } I_1 \text{ recommended to users}}{|I|}
%\end{equation}
%This is identical to calculating the coverage for I1.
%\begin{equation}
%    \mathbb{E} \left[ \text{Item\_Exp}(i, R) \mid i \in I_2 \right] = \frac{\text{Number of items from } I_2 \text{ recommended to users}}{|I|}
%\end{equation}
%%\iw{$\mathbb{E}$ is never defined. }
%	
%$|I|$ represents the total number of items regardless of their categories.
%Exposure relates to the degree to which items or item groups are presented uniformly to all users or different user groups \cite{deldjoo2024fairness}.
%%We also examined item exposure, particularly for popular and unpopular items, and found that the results are highly consistent with DPF trends.
%Since the results for exposure largely mirror the DPF results, we omit detailed exposure plots from the paper for conciseness.

\subsubsection{Privacy}

The privacy budget $\epsilon$ quantifies the level of privacy protection provided by a \ac{DP} mechanism.
Privacy budget specifies an upper-bound on our privacy loss.
%We can quantitatively compare mechanisms by comparing their privacy budgets.
A smaller privacy budget requires the output to be more noisy, often leading to reduced utility.
For the LDP mechanism, the value of $\epsilon$ is determined before the training process begins, as explained in Section \ref{sec:RSLDP}.

For \ac{DPSGD}, the privacy guarantee at each gradient descent step is characterized by the noise multiplier $\sigma = \frac{\sqrt{2 \log\left(\frac{1.25\delta}{\epsilon}\right)}}{\epsilon}$.
This choice of $\sigma$ ensures that each step satisfies $(\epsilon, \delta)$-differential privacy \cite{abadi2016deep}.
%\iw{please be precise with your language and the mathematical notation. $\sigma$ is not a privacy guarantee!}
The total privacy cost across the entire training process is analyzed using a privacy accountant technique.
%In this paper, we use the RDP accountant, as it offers a balance between computational efficiency (in terms of runtime and memory) and accuracy in estimating privacy loss.
It is important to note that the $\epsilon$ values for LDP and DPSGD may not be directly comparable. 
This means that the same $\epsilon$ does not necessarily provide the same level of privacy protection across different mechanisms.

%\iw{add explanation that epsilons between LDP and DPSGD may not be comparable}

% \quad 
%  \text{and} \quad GAP_{q}(G) = \frac{\sum_{u \in G} \frac{\sum_{i \in q_{u}} \theta(i)}{|q_{u}|}}{|G|}  

%$G$ indicates the niche/blockbuster/diverse group
%users $u$ 

%$\theta(i)$ in $GAP_{p}(G)$ represents the popularity value for item $i$ in the user history.
%To determine its popularity level, we look at the percentage of users who have rated an item.
%Also, $\theta$ in $GAP_{q}(G)$ represents the popularity value for item $i$ in the recommendation list to user $u$ in group $g$.
%Ultimately, we calculate the value of "popularity lift" by comparing the average popularity of the group's profile and the average popularity of the recommendations for group "g" as follows:

%A negative $PL$ suggests that the recommendations focus less on popular items compared to the user's profile.
%If $Pl$ is zero, there is no amplification of popularity bias.

%\todo[inline]{do our results show this too?}
\begin{table*}[t]
	\caption{Average training time and epochs for different models on 1M and Yelp over a set of noise multipliers: (0.2, 0.4, 0.8, 2, 4). These noise multipliers result in different epsilon for different models; the relevant epsilon values do not significantly impact training time per epoch.  (Values are averaged, and near-zero standard deviations are not reported)}
	\centering
	\setlength{\tabcolsep}{6pt}
	\begin{adjustbox}{max width=\textwidth}
		\begin{tabular}{cccccc}
			\textbf{Privacy} & \textbf{Model} & \multicolumn{2}{c}{\textbf{1M}} & \multicolumn{2}{c}{\textbf{Yelp}} \\
			\cmidrule(lr){3-4} \cmidrule(lr){5-6}
			& & Time/Epoch (s) & Early Stop Epoch
			& Time/Epoch (s) & Early Stop Epoch \\
			\midrule
			\multirow{4}{*}{Non-Private} 
			& NCF & $70.3 \pm 13.2$&$21\pm23$ & $27.7\pm0.0$ & 8.0 \\
			& VAE & $7.6 \pm 0.2$ & 133.0 & $52.6\pm6.7$ & 14.0 \\
			& SVD & $1183.6\pm 754.7$ & $21.3 \pm 12.5$ & $5451.3\pm 1359.4$ & 46.0 \\
			& BPR & $105.9\pm 31.2$ & 48.0 & $182.0\pm27.1$ & 43.0 \\
			
			\midrule
			\multirow{4}{*}{DPSGD for $\epsilon$}
			& NCF & $106.8\pm1.64$ &$18\pm10$ & $1265.7\pm 33.1$ & $8.0\pm1.2$ \\
			& VAE & $21.7\pm0.44$ & 11.0 & $55.32\pm 6.3$ & 11.0 \\
			& SVD & $1487.8\pm702.5$ & 400.0 &  $29559.8\pm 4070.1$& 400.0 \\
			& BPR & $91.1\pm2.7$ & 400.0 & $24078.1\pm 1944.4 $ & 400 \\
			\bottomrule
		\end{tabular}
	\end{adjustbox}
	\label{tab:training_time_comparison}
\end{table*}

\subsection{Implementation and Experimental Settings}
\label{sec:expsettings}

%\iw{add detail for the implementation: what did we implement ourselves, which libraries did we rely on? also comment on whether code will be open source}

We evaluate four recommender algorithms on a range of privacy settings.
The \acp{RS} are configured by varying the noise multiplier for \ac{DPSGD} in Opacus, which determines the standard deviation of the Gaussian noise added to the gradients.
%For the privacy parameters, we consider a range of noise multipliers.
%The noise multiplier determines the standard deviation of the Gaussian noise added to the gradients.
High noise multipliers lead to smaller privacy budgets $\epsilon$.
To calculate the privacy loss of training our models, we adopt PRV approach \cite{gopi2021numerical} implemented in Opacus.
%\iw{please be consistent. in the section directly above, you claim that DPSGD uses the moments accountant technique -> fix the section above}
For \ac{ldp}, we evaluate the recommendation algorithms by varying the $\epsilon$. 
%This approach is implemented in Opacus  
%\todo[inline]{need to explain that noise multiplier influences final measured epsilon, and how the dependency works}

%For each combination of privacy setting, recommender algorithm, and dataset, we tune the hyperparameters batch size, learning rate, and clipping bound.
To ensure that all models are evaluated under optimized conditions, we performed hyperparameter tuning for every model, dataset, and privacy mechanism.
This tuning acts as an ablation study over the key parameters including batch size, learning rate, clipping bound (gradient norm C), Weight decay and Latent dimensionality.
We implement all algorithms in Python/PyTorch.
For SVD and BPR, we define user and item embeddings, then train the models using SGD to facilitate the application of DPSGD.
%(\iw{add Git link -- we should review the code to prepare for publishing it. Please add a README.md to the code repository to explain what the repo does and how the experiments can be reproduced.}).
%This selection was guided by the highest observed performance from a tested subset of possible values. 
%\todo[inline]{optimal is a strong word. how did we determine that parameters are indeed optimal?}
For each \ac{RS} model, we conducted a grid-based ablation study, training multiple configurations and selecting hyperparameters based on utility (NDCG) on the held-out validation set.
This approach aligns with common practice in the field, where tuning decisions are typically made based on utility metrics.
To ensure reproducibility of our results, our code is available at \url{https://gitlab.com/dmi-pet-public/parsarad2025privacy}.

For BPR, SVD and NCF, we found that a batch size of 256 and an L2 norm regularization with a weight decay of $10^{-5}$ provide better utility results. 
For the VAE model, a higher weight decay of 0.01 shows better utility and stability during training.
%\iw{we need to justify parameter settings}
VAE also used a batch size of 500 for the 1M dataset and 10,000 for the Yelp dataset in the DPSGD state; for both the LDP and non-private settings, a batch size of 5000 was used.

The initial learning rate for BPR and SVD was set to 0.0005 and reduced by a factor of 0.1 if the test loss remained unchanged over four consecutive epochs.
For \ac{VAE}, we used an initial learning rate of 0.001, which similarly reduced it if the test loss stagnated over four consecutive epochs.
For NCF, the learning rate was set to 0.0005 for Movielens and 0.0001 for Yelp.
The NCF learning rate was scheduled  using a StepLR scheduler, which multiplicatively reduces the learning rate by a factor of 0.9 every four epochs.

The latent dimension For the BPR and SVD algorithms was set to 5 for 1M and 8 for Yelp, based on validation performance.
%For BPR, an optimal latent dimension of 8 was used for Yelp.  \todo{and BPR/Yelp?}
%The DRS algorithm used different settings for learning rates: 0.0005 for Movielens and 0.0001 for Yelp, managed via a StepLR scheduler that reduces \todo{reduces sounds like 0.9 is subtracted. what you mean is multiplicative?} the rate by 0.9 every four epochs.
For NCF, the matrix factorization dimension was fixed at 8, and the MLP structure used a 3-layers structure with hidden size of 16, 8, and 4.
To prevent overfitting and improve model generalization, we use a dropout rate of 0.5 for NCF.

%\iw{what does "at least 400" mean? were there cases with more than 400?}
We train all models for 400 epochs and used early stopping to prevent overfitting: training is stopped if no improvement observed on the validation set for six consecutive epochs.
%\todo[inline]{validation set was not mentioned in the 80/20 train/test split. Is early stopping is based on the test set (this would not be best practice because test set is not supposed to be used during training)? In the future, we definitely need to use a train/test/validation split.}
%\todo[inline]{It was based on validation set (All decision-making in the change training process was based on validation data.): updated in text }
DP noise introduces randomness during training, making single-run results potentially unreliable.
Therefore, we executed multiple independent training runs for each model–dataset–privacy configuration to obtain stable and representative averages, to the extent feasible given the substantial training times.
Specifically, we perform three runs for BPR/1M and SVD/1M, two runs for BPR/Yelp, SVD/Yelp, and VAE/Yelp, five runs for VAE/1M, and six runs for NCF.
Table \ref{tab:training_time_comparison} shows the average training times for each model/dataset combination as well as the early stopping epoch. 
Note that there was no significant difference in training time depending on $\epsilon$.

\begin{figure}[t]
	\centering
		\subfloat[NCF, 1M]{\includegraphics[width=0.23\textwidth]{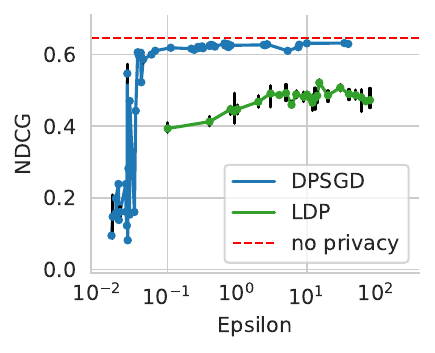}\vspace{-2mm}\label{fig:plot1_NDCG}}
		\subfloat[BPR, 1M]{\includegraphics[width=0.23\textwidth]{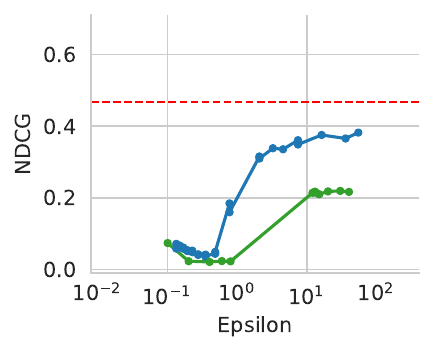}\vspace{-2mm}\label{fig:plot5_NDCG}}
		\subfloat[SVD, 1M]{\includegraphics[width=0.23\textwidth]{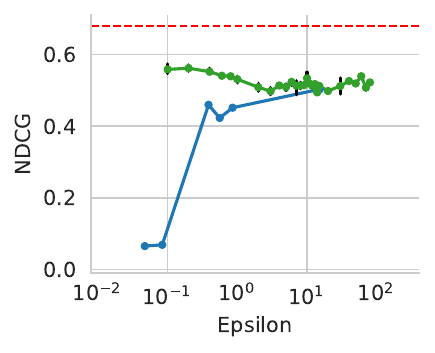}\vspace{-2mm}\label{fig:plot9_NDCG}}
		\subfloat[VAE, 1M]{\includegraphics[width=0.23\textwidth]{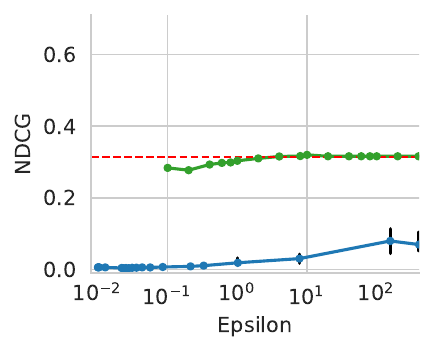}\vspace{-2mm}\label{fig:plot13_NDCG}}\\

		\subfloat[NCF, Yelp]{\includegraphics[width=0.23\textwidth]{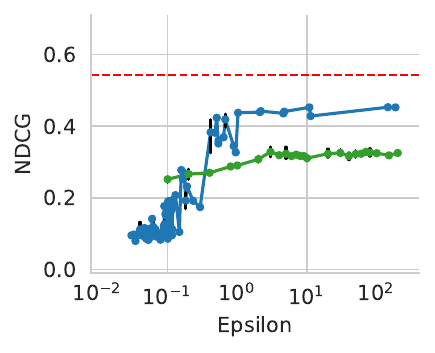}\vspace{-2mm}\label{fig:plot3_NDCG}}
		\subfloat[BPR, Yelp]{\includegraphics[width=0.23\textwidth]{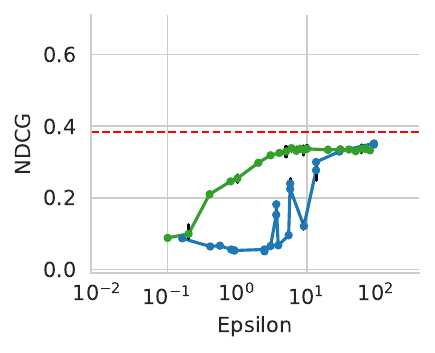}\vspace{-2mm}\label{fig:plot7_NDCG}}		
		\subfloat[SVD, Yelp]{\includegraphics[width=0.23\textwidth]{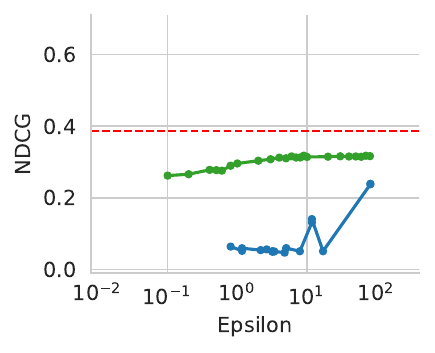}\vspace{-2mm}\label{fig:plot11_NDCG}}		
		\subfloat[VAE, Yelp]{\includegraphics[width=0.23\textwidth]{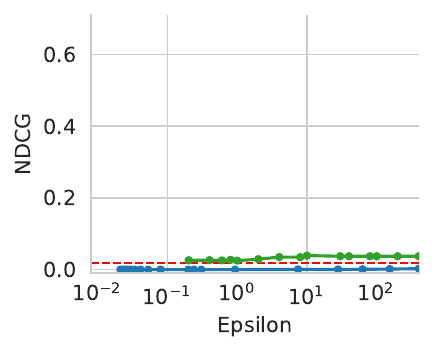}\vspace{-2mm}\label{fig:plot15_NDCG}} \\

	\caption{NDCG Across Different Models and Two Datasets with DPSGD and LDP}
	\label{fig:plots_NDCG_16}
\end{figure}

\section{Results}
\label{sec_results}

In this section, we present results for privacy, utility, and bias across four recommender systems on the MovieLens 1M and Yelp datasets.
We first show results for the privacy-utility trade-off (Section \ref{sec:privacy-utility}), followed by an analysis of privacy-bias trade-offs (Section \ref{sec:privacy-bias}) and utility-bias trade-offs (Section \ref{sec:utility-bias}).
In each case, we first show overall results for all item and user types combined and then separately analyze popular/unpopular items and blockbuster/diverse/niche users.
Finally, we 
%discuss trade-offs between utility, privacy, and bias (Section \ref{sec:results-tradeoffs}), and 
synthesize insights across these dimensions (Section \ref{sec:results-discussion}).

%\iw{need an introductory paragraph that gives an overview of the results section}
%\todo[inline]{text in results section does not yet refer to any of the figures.}
%\todo[inline]{It appears that DP performs better than DPSGD when epsilon is small. However, as epsilon increases, DPSGD improves while DP remains almost unchanged.}
%\todo[inline]{Having more iterations or batch size in each model? impact on DP and DPSGD}

\subsection{Privacy-Utility Trade-off}
\label{sec:privacy-utility}

\paragraph{Ranking quality (NDCG)}
Analyzing the ranking quality in terms of NDCG for different recommender algorithms with privacy protections (Figure \ref{fig:plots_NDCG_16}), we see that ranking quality decreases with lower $\epsilon$, as expected, in almost all cases.
%Similar trends are observed for recall and MRR, and thus are not presented here.
Models trained with DPSGD show high sensitivity to $\epsilon$, with a marked decrease in ranking quality for smaller $\epsilon$ values.
In contrast, models trained with LDP show less variation in ranking quality across privacy levels, indicating a more stable but potentially less privacy-sensitive response.

Notably, NCF trained with DPSGD achieves performance comparable to the non-private baseline, even under high privacy constraints ($\epsilon < 1$, see Figures \ref{fig:plot1_NDCG} and \ref{fig:plot3_NDCG}). 
This suggests that NCF, when trained with DPSGD, maintains its ranking effectiveness despite differential privacy noise.
%\iw{please ensure consistent upper-case spelling of the word Figure when you use it to refer to a Figure -- 20250701 this is still relevant. Please use a case-sensitive search to find spellings you need to correct.}

LDP outperforms DPSGD in several dataset-model combinations, namely Yelp/SVD (Figure \ref{fig:plot11_NDCG}), Yelp/BPR (Figure \ref{fig:plot7_NDCG}), 1M/SVD (Figure \ref{fig:plot9_NDCG}), and 1M/VAE (Figure \ref{fig:plot13_NDCG}).
In terms of absolute NDCG, NCF consistently outperforms other models across most datasets and privacy mechanisms.
VAE/Yelp, as shown in Figure \ref{fig:plot15_NDCG}, has low NDCG in private and non-private settings.
%\iw{we need to look at the VAE code. why is performance so bad compared to other algorithms and also compared to the other dataset? we need at least a good explanation for this}
%\shp{For 1M, we already checked the performance while the NDCG was fixed as 50 (same as the main paper), and the performance was the same as the main paper.
	%Here, we used NDCG@10, which is more common; therefore, the performance drop is standard}%\iw{please fix references to figures - they point to KLD instead of NDCG}
%\shp{For Yelp, We can look at these paper: \cite{antognini2021fast},\cite{karamanolakis2018item}, and \cite{carraro2020conditioneda}, based on these paper, different performance are reported for VAE + Yelp (Please check my detailed information in the 2023-Shiva-Privacy-enhancing-approaches-in-RS/VAE\_Yelp/README.md )}
%\shp{The following lines are added based on my experiments}
Our results show that VAE performs worse on Yelp than on 1M and compared to other algorithms.
This weaker performance is consistent with prior studies \cite{antognini2021fast}, \cite{karamanolakis2018item}, \cite{carraro2020conditioneda}.
A key reason is data sparsity: in our filtered Yelp set, the interaction ratio is only 0.24\%, compared to 4.22\% for 1M, which weakens the user–item signal\footnote{Note that detailed comparisons with prior work are complicated by different preprocessing approaches.
	Some studies binarize ratings above 4.5 while restricting to restaurants without clear frequency thresholds \cite{antognini2021fast}, while others keep the 20 most popular categories and apply 4/10 user–item constraints \cite{carraro2020conditioneda}. 
	Our filtering (ratings $\geq 3$, $\geq 5$ reviews per user/item) yields different distributions, and the continuously updated Yelp dataset further increases sparsity compared to the smaller, denser versions used in 2020--2021.}.

Differences in evaluation also affect outcomes: prior work often reports NDCG@50 or Recall@100, where VAE performs better, while our stricter NDCG@10 penalizes missing top-ranked items.
In addition, Yelp spans a diverse item space (restaurants, shops, services) with strong regional variation, making collaborative filtering harder for generative models like VAE that assume more uniform interaction patterns.
Overall, encoding user representations from sparse and diverse data remains a challenge for VAEs \cite{liang2024survey}.

\begin{figure}[t]
	\centering
	\subfloat[NCF, 1M]{\includegraphics[width=0.23\textwidth]{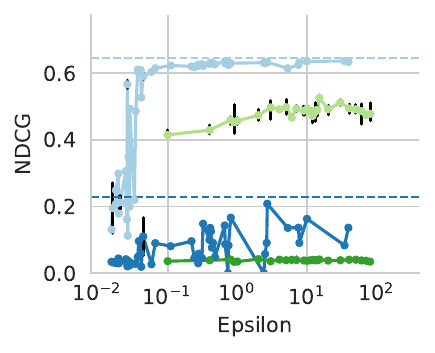}\vspace{-2mm}\label{fig:plot1_NDCG_per_category}}
	\subfloat[BPR, 1M]{\includegraphics[width=0.23\textwidth]{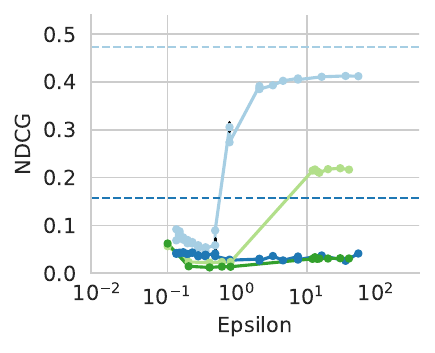}\vspace{-2mm}\label{fig:plot5_NDCG_per_category}}
	\subfloat[SVD, 1M]{\includegraphics[width=0.23\textwidth]{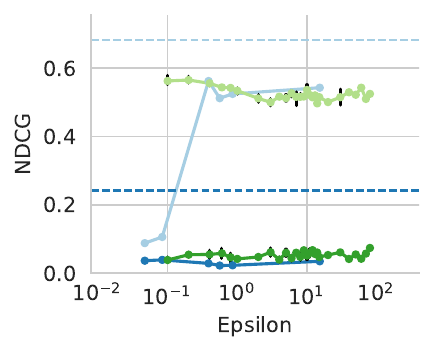}\vspace{-2mm}\label{fig:plot9_NDCG_per_category}}
	\subfloat[VAE, 1M]{\includegraphics[width=0.23\textwidth]{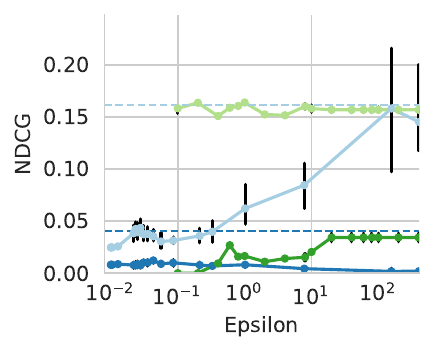}\vspace{-2mm}\label{fig:plot13_NDCG_per_category}}
	
	\subfloat[NCF, Yelp]{\includegraphics[width=0.23\textwidth]{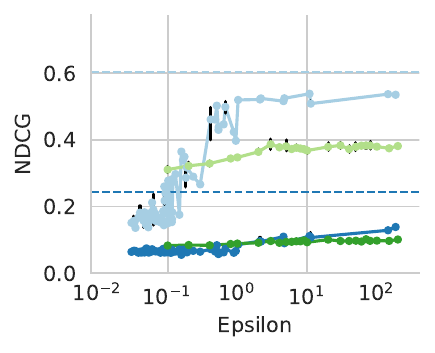}\vspace{-2mm}\label{fig:plot3_NDCG_per_category}}
	\subfloat[BPR, Yelp]{\includegraphics[width=0.23\textwidth]{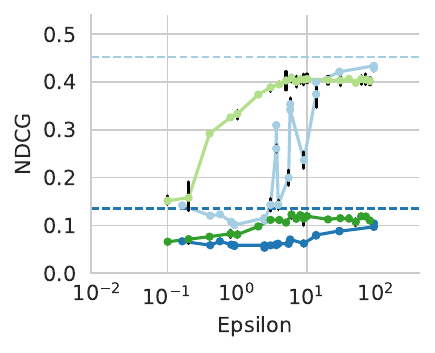}\vspace{-2mm}\label{fig:plot7_NDCG_per_category}}
	\subfloat[SVD, Yelp]{\includegraphics[width=0.23\textwidth]{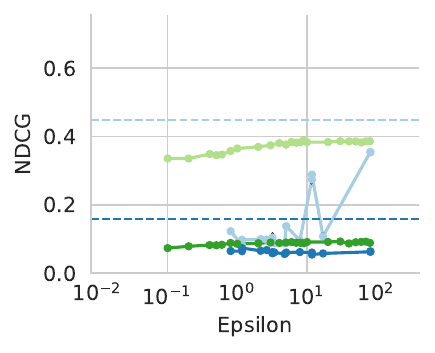}\vspace{-2mm}\label{fig:plot11_NDCG_per_category}}
	\subfloat[VAE, Yelp]{\includegraphics[width=0.23\textwidth]{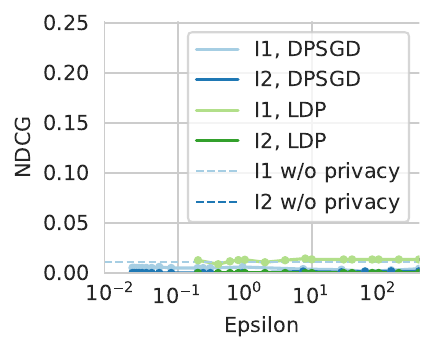}\vspace{-2mm}\label{fig:plot15_NDCG_per_category}}

	\caption{NDCG for popular items (I1, light colors) and unpopular items (I2, dark colors)}
	%\iw{this caption does not make a lot of sense (except for the first word). Take this opportunity to carefully think about and revise \textit{all} captions in the paper}
	\label{fig:plots_NDCG_per_category_16}
	%\iw{move legend to the VAE/Yelp plot}
\end{figure}

\paragraph{NDCG for Popular vs. Unpopular Items}
Analyzing NDCG separately for popular items (I1) and unpopular items (I2) in Figure \ref{fig:plots_NDCG_per_category_16} shows that changes in NDCG with varying $\epsilon$ differ between the two item groups.
For unpopular items, NDCG is low, and remains largely unaffected by the privacy level ($\epsilon$) under both LDP and DPSGD,  suggesting that privacy mechanisms do not significantly alter ranking effectiveness for less frequently recommended items (\cref{fig:plots_NDCG_per_category_16}).

For popular items, NDCG improves as privacy constraints are relaxed.
This means that the improvement in overall NDCG observed in Figure \ref{fig:plots_NDCG_16} is attributable to an increase in NDCG for popular items, rather than changes in the ranking quality of unpopular items.

\begin{figure}[t]
	\centering
	\subfloat[NCF with DPSGD]{\includegraphics[width=0.23\textwidth]{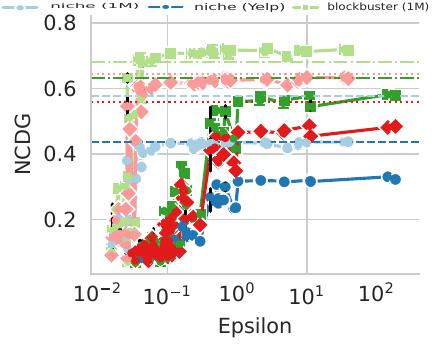}\label{fig:plot1_ndcg_user_based}} 
	\subfloat[BPR with DPSGD]{\includegraphics[width=0.23\textwidth]{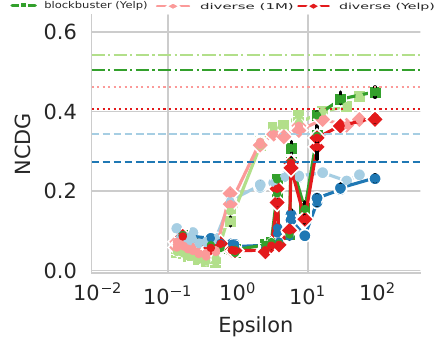}\label{fig:plot3_ndcg_user_based}} 
	\subfloat[SVD with DPSGD]{\includegraphics[width=0.23\textwidth]{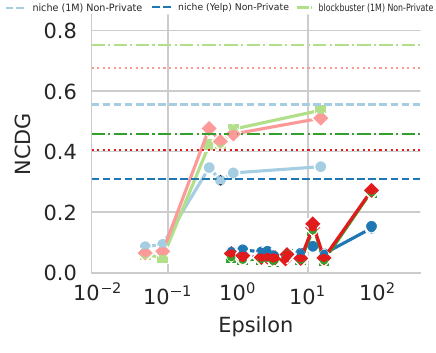}\label{fig:plot5_ndcg_user_based}}
	\subfloat[VAE with DPSGD]{\includegraphics[width=0.23\textwidth]{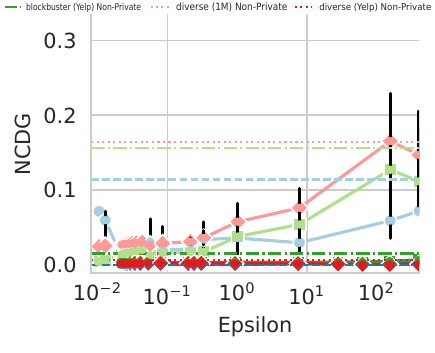}\label{fig:plot7_ndcg_user_based}} 
	
	\subfloat[NCF with LDP]{\includegraphics[width=0.23\textwidth]{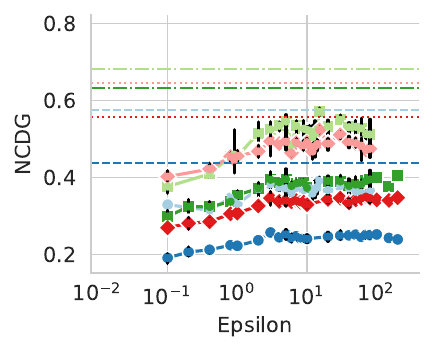}\label{fig:plot2_ndcg_user_based}} 
	\subfloat[BPR with LDP]{\includegraphics[width=0.23\textwidth]{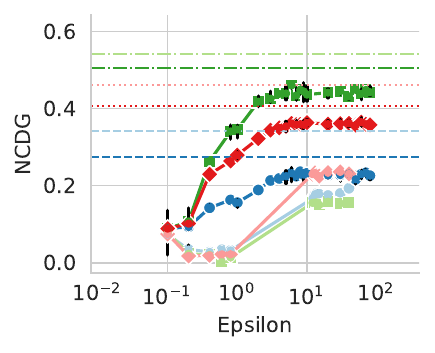}\label{fig:plot4_ndcg_user_based}}
	\subfloat[SVD with LDP]{\includegraphics[width=0.23\textwidth]{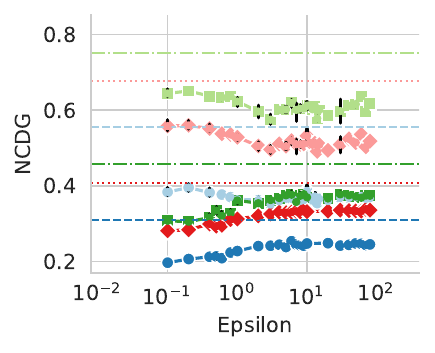}\label{fig:plot6_ndcg_user_based}}
	\subfloat[VAE with LDP]{\includegraphics[width=0.23\textwidth]{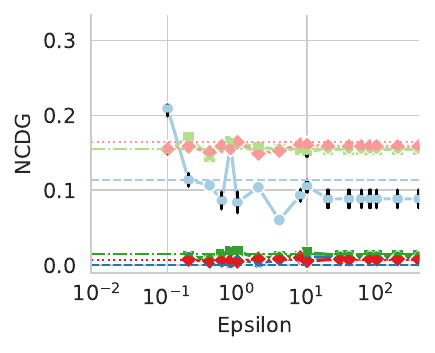}\label{fig:plot8_ndcg_user_based}}
	
	\caption{NDCG by user type (niche (blue circles), diverse (red diamonds), and blockbuster (green squares)) and dataset (Yelp (dark colors) and 1M (light colors))}
	\label{fig:plots_ndcg_user_based_8}
	%\iw{add legend to this figure: try VAE/DPSGD, otherwise use inkscape, spread it out over figures (a)-(d)}
\end{figure}

\paragraph{NDCG by User Type}
Figure~\ref{fig:plots_ndcg_user_based_8} presents the NDCG values obtained under \ac{ldp} and \ac{DPSGD} across different user types.
In most cases, NDCG differs substantially between user groups, indicating that privacy mechanisms affect users with distinct interaction behaviors in different ways.
Across all models and datasets, niche users consistently achieve the lowest NDCG values, and their performance plateaus rapidly as $\epsilon$ increases.
For niche users, NDCG is generally higher on 1M than on Yelp (e.g., approximately 0.4 vs.~0.3 for NCF with \ac{DPSGD}), with one notable exception: for BPR with \ac{ldp} (Figure~\ref{fig:plot3_ndcg_user_based}), lower $\epsilon$ yields slightly higher NDCG on Yelp (0.22 vs.~0.18).
For diverse users, NCF (Figures~\ref{fig:plot1_ndcg_user_based} and~\ref{fig:plot2_ndcg_user_based}) achieves the highest NDCG scores, reaching about 0.7 with \ac{DPSGD} and 0.6 with \ac{ldp}.
In contrast, for blockbuster users, BPR and SVD (Figures~\ref{fig:plot3_ndcg_user_based}–\ref{fig:plot6_ndcg_user_based}) obtain the best performance, typically exceeding 0.4.
Finally, VAE (Figures~\ref{fig:plot7_ndcg_user_based} and~\ref{fig:plot8_ndcg_user_based}) performs worst overall, with the most pronounced degradation observed for niche users ($\mathrm{NDCG} \approx 0$).

\subsection{Privacy-Bias Trade-off}
\label{sec:privacy-bias}

\begin{figure}[t]
	\centering
	\subfloat[NCF, 1M]{\includegraphics[width=0.23\textwidth]{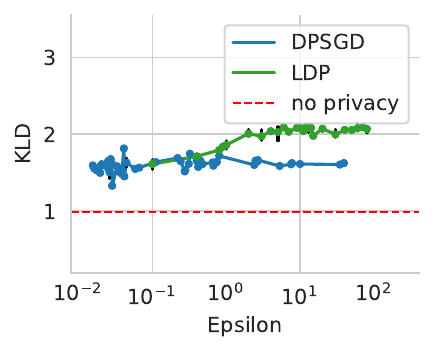}\vspace{-2mm}\label{fig:plot1_kld}}
	\subfloat[BPR, 1M]{\includegraphics[width=0.23\textwidth]{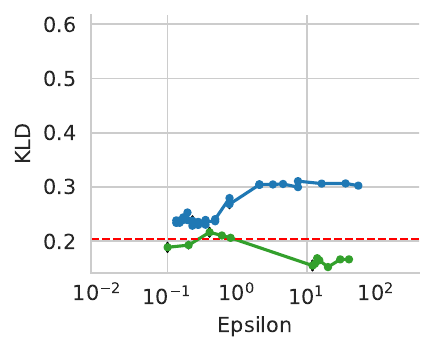}\vspace{-2mm}\label{fig:plot5_kld}}
	\subfloat[SVD, 1M]{\includegraphics[width=0.23\textwidth]{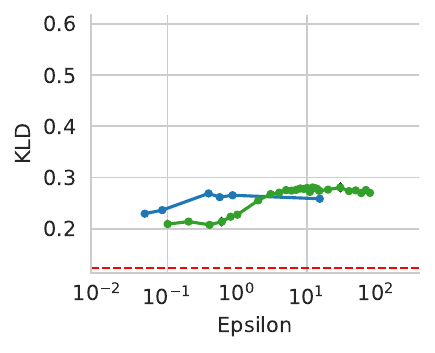}\vspace{-2mm}\label{fig:plot9_kld}}
	\subfloat[VAE, 1M]{\includegraphics[width=0.23\textwidth]{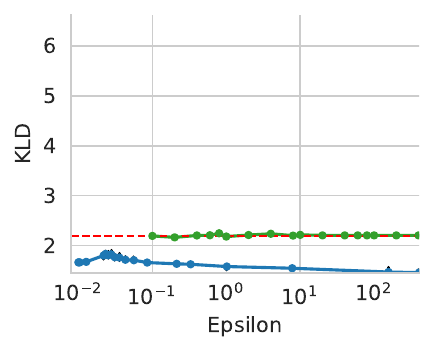}\vspace{-2mm}\label{fig:plot13_kld}}
	
	\subfloat[NCF, Yelp]{\includegraphics[width=0.23\textwidth]{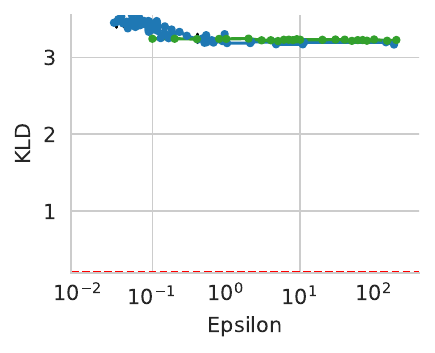}\vspace{-2mm}\label{fig:plot3_kld}}
	\subfloat[BPR, Yelp]{\includegraphics[width=0.23\textwidth]{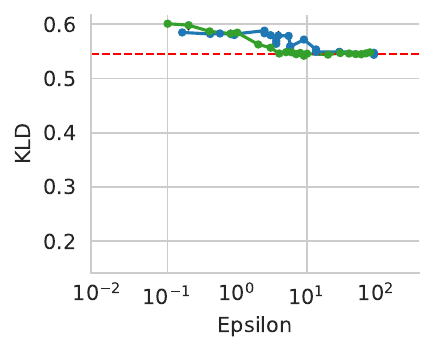}\vspace{-2mm}\label{fig:plot7_kld}}
	\subfloat[SVD, Yelp]{\includegraphics[width=0.23\textwidth]{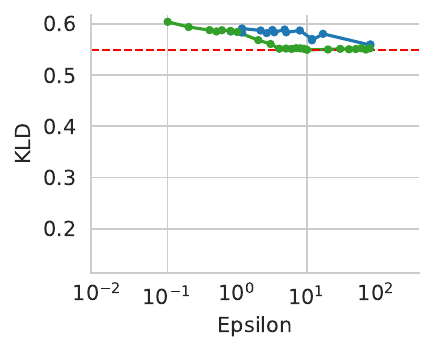}\vspace{-2mm}\label{fig:plot11_kld}}
	\subfloat[VAE, Yelp]{\includegraphics[width=0.23\textwidth]{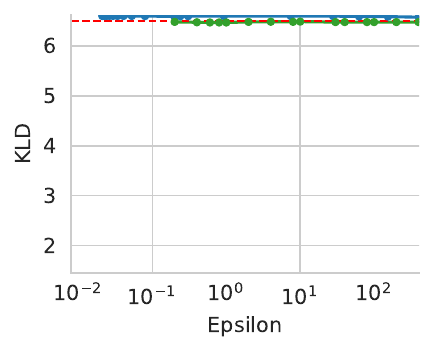}\vspace{-2mm}\label{fig:plot15_kld}}
	\caption{KLD Across Different Models and Two Datasets with DPSGD and LDP}
	\label{fig:plots_kld_16}
\end{figure}

\paragraph{Miscalibration (KLD)}
Figure \ref{fig:plots_kld_16} shows the levels of miscalibration, measured by KLD, in our experiments.
We can see that whether the use of privacy protections affects miscalibration levels depends on the recommender algorithm.
Specifically, miscalibration increases for NCF when privacy mechanisms are applied (Figures \ref{fig:plot1_kld} and \ref{fig:plot3_kld}),
%see Figures \ref{fig:plot1_kld}, \ref{fig:plot2_kld},\ref{fig:plot3_kld},\ref{fig:plot4_kld}),
while miscalibration for other algorithms remains similar to that of the non-private baseline.
For all algorithms, there is no significant correlation between the degree of miscalibration and the level of privacy ($\epsilon$).
%Additionally, the level of privacy ($\epsilon$) does not have a significant impact on the degree of miscalibration.
%\iw{what does it mean?}
%\shp{the $\epsilon$ values does not change the distribution of item categories (genres or businesses) in the recommended list, Calibration measures how well the genre distribution of the recommended list matches that of the user’s history.
	%DP affects model training gradient or training data noise,  but it doesn't directly optimize or constrain the output genre distribution. 
	%Genre categories are aggregated statistics, and small changes (due to $\epsilon$) in the recommendation list don't shift these dramatically.}

\begin{figure}[t]
	\centering
	\subfloat[NCF, 1M]{\includegraphics[width=0.23\textwidth]{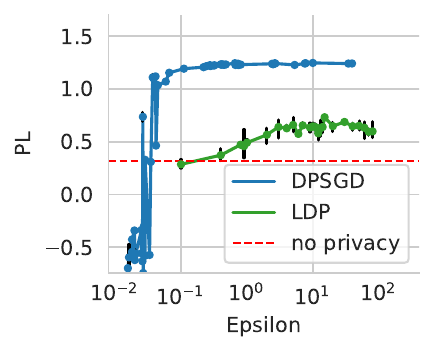}\vspace{-2mm}\label{fig:plot1_PL}}
	\subfloat[BPR, 1M]{\includegraphics[width=0.23\textwidth]{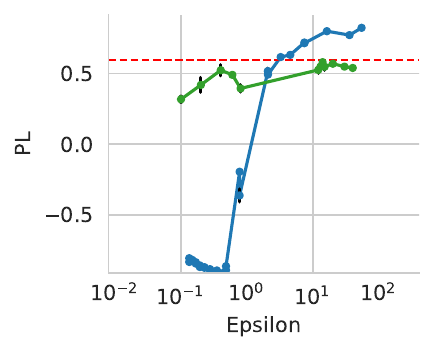}\vspace{-2mm}\label{fig:plot5_PL}} 
	\subfloat[SVD, 1M]{\includegraphics[width=0.23\textwidth]{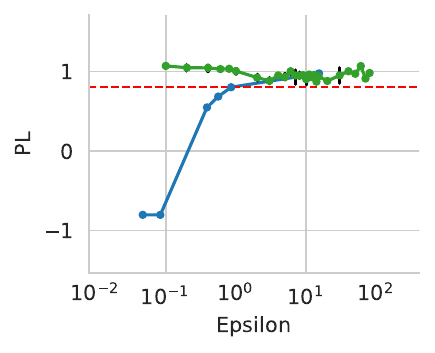}\vspace{-2mm}\label{fig:plot9_PL}} 
	\subfloat[VAE, 1M]{\includegraphics[width=0.23\textwidth]{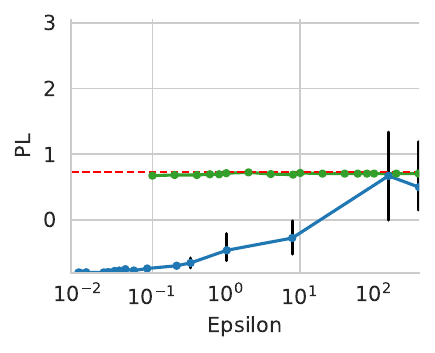}\vspace{-2mm}\label{fig:plot13_PL}} 
	
	\subfloat[NCF, Yelp]{\includegraphics[width=0.23\textwidth]{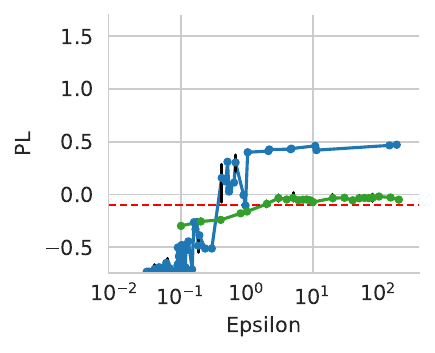}\vspace{-2mm}\label{fig:plot3_PL}} 
	\subfloat[BPR, Yelp]{\includegraphics[width=0.23\textwidth]{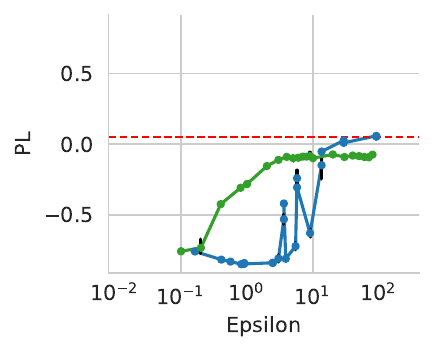}\vspace{-2mm}\label{fig:plot7_PL}} 
	\subfloat[SVD, Yelp]{\includegraphics[width=0.23\textwidth]{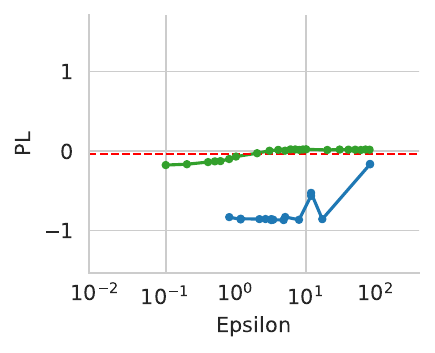}\vspace{-2mm}\label{fig:plot11_PL}} 
	\subfloat[VAE, Yelp]{\includegraphics[width=0.23\textwidth]{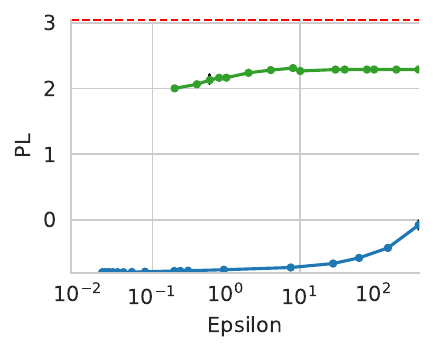}\vspace{-2mm}\label{fig:plot15_PL}} 
	\caption{Popularity Lift Across Different Models and Two Datasets with DPSGD and LDP}
	\label{fig:plots_PL_16}
\end{figure}

\paragraph{Popularity Lift}
%\iw{please check how I have restructured the two previous result sections: starting with an introduction, discussing general observations first, then going to more specific observations. please restructure the remaining paragraphs in a similar way.}

In Figure~\ref{fig:plots_PL_16}, we report the \ac{PL}, showing that changes in \ac{PL} are influenced by both the type of privacy mechanism and the level of privacy applied.
Specifically, at high privacy levels, DPSGD tends to produce negative \ac{PL} values, suggesting that it suppresses recommendations for popular items.
While PL generally remains close to the baseline at lower privacy levels, an exception occurs with NCF/DPSGD, which shows a higher PL, as depicted in Figures \ref{fig:plot1_PL} and \ref{fig:plot3_PL}.
In contrast, LDP shows minimal deviation from the baseline across all models and privacy levels, indicating that it does not introduce substantial bias toward or against popular items. 
Overall, similar to the baseline, PL for LDP remains consistently positive, indicating that recommendations continue to favor popular items.
Popularity lift is generally lower for the Yelp dataset.
This may be because the Yelp dataset has a larger item space with sparser interactions. 
This means that each user interacts with fewer items, and even popular items have fewer total interactions, resulting in less strong signals for the model to confidently overfit to, especially under DP noise.

\begin{figure}[t]
	\centering
	\subfloat[NCF, 1M]{\includegraphics[width=0.23\textwidth]{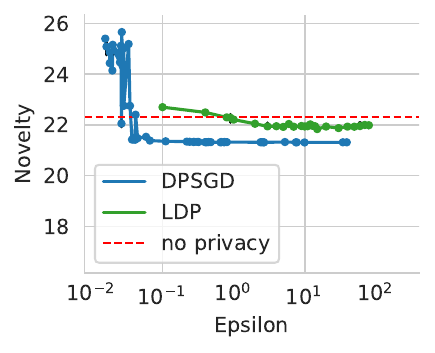}\vspace{-2mm}\label{fig:plot1_Novelty}} 
	\subfloat[BPR, 1M]{\includegraphics[width=0.23\textwidth]{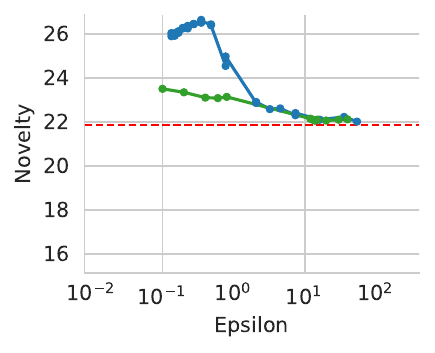}\vspace{-2mm}\label{fig:plot5_Novelty}} 
	\subfloat[SVD, 1M]{\includegraphics[width=0.23\textwidth]{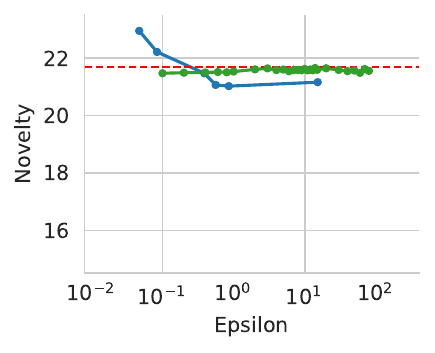}\vspace{-2mm}\label{fig:plot9_Novelty}} 
	\subfloat[VAE, 1M]{\includegraphics[width=0.23\textwidth]{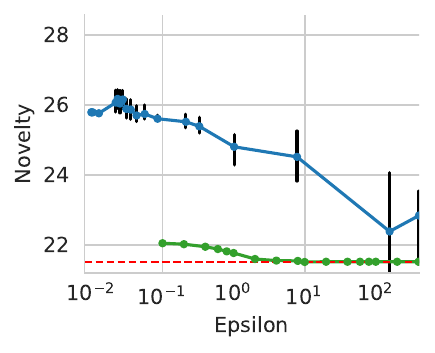}\vspace{-2mm}\label{fig:plot13_Novelty}} 
	
	\subfloat[NCF, Yelp]{\includegraphics[width=0.23\textwidth]{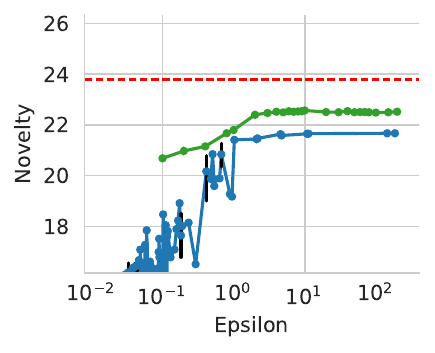}\vspace{-2mm}\label{fig:plot3_Novelty}} 
	\subfloat[BPR, Yelp]{\includegraphics[width=0.23\textwidth]{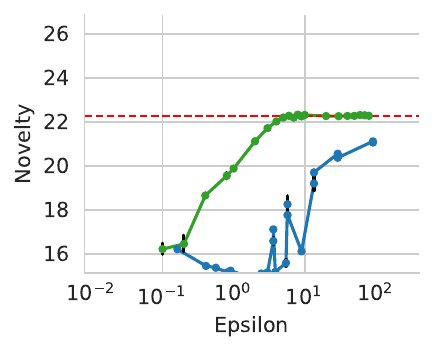}\vspace{-2mm}\label{fig:plot7_Novelty}} 
	\subfloat[SVD, Yelp]{\includegraphics[width=0.23\textwidth]{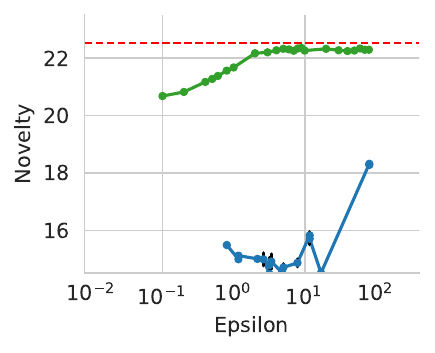}\vspace{-2mm}\label{fig:plot11_Novelty}}		
	\subfloat[VAE, Yelp]{\includegraphics[width=0.23\textwidth]{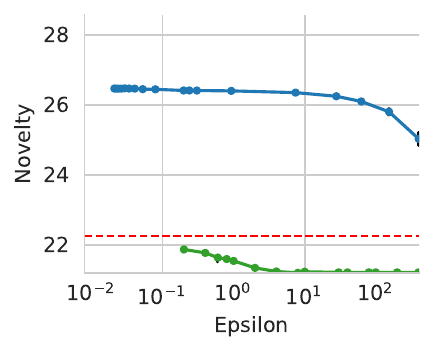}\vspace{-2mm}\label{fig:plot15_Novelty}} 
	
	\caption{Novelty Across Different Models and Two Datasets with DPSGD and LDP}
	\label{fig:plots_Novelty_16}
\end{figure}

\paragraph{Novelty}

The Novelty results in Figure~\ref{fig:plots_Novelty_16} suggest that privacy constraints may have a dataset-specific influence on novelty.
In particular, on the Yelp dataset (bottom row in Figure~\ref{fig:plots_Novelty_16}), lower privacy increases novelty, implying that as privacy constraints are relaxed, the model recommends more diverse and less popular items.
Conversely, for the 1M dataset (top row in Figure~\ref{fig:plots_Novelty_16}), lower privacy generally decreases novelty.

\begin{figure}[t]
	\centering
	\subfloat[NCF, 1M]{\includegraphics[width=0.23\textwidth]{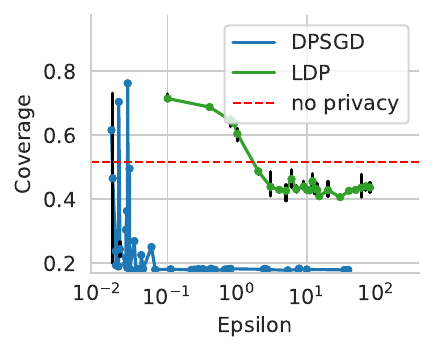}\vspace{-2mm}\label{fig:plot1_Coverage}} 
	\subfloat[BPR, 1M]{\includegraphics[width=0.23\textwidth]{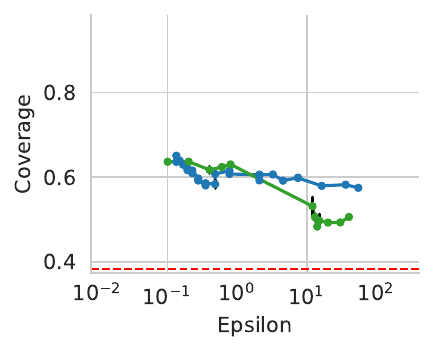}\vspace{-2mm}\label{fig:plot5_Coverage}} 
	\subfloat[SVD, 1M]{\includegraphics[width=0.23\textwidth]{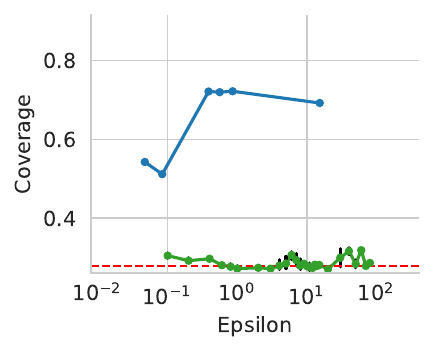}\vspace{-2mm}\label{fig:plot9_Coverage}} 
	\subfloat[VAE, 1M]{\includegraphics[width=0.23\textwidth]{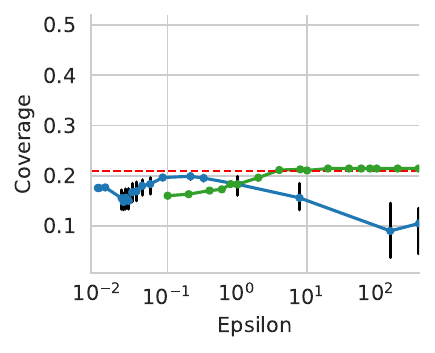}\vspace{-2mm}\label{fig:plot13_Coverage}} 
	
	\subfloat[NCF, Yelp]{\includegraphics[width=0.23\textwidth]{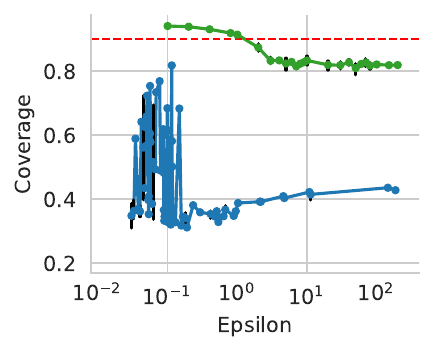}\vspace{-2mm}\label{fig:plot3_Coverage}} 
	\subfloat[BPR, Yelp]{\includegraphics[width=0.23\textwidth]{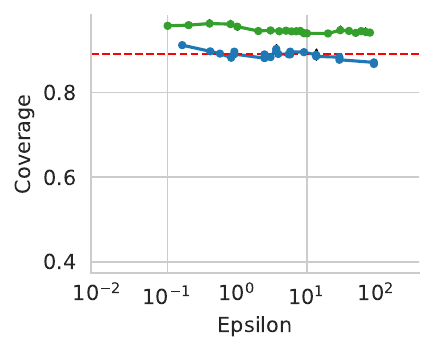}\vspace{-2mm}\label{fig:plot7_Coverage}} 		
	\subfloat[SVD, Yelp]{\includegraphics[width=0.23\textwidth]{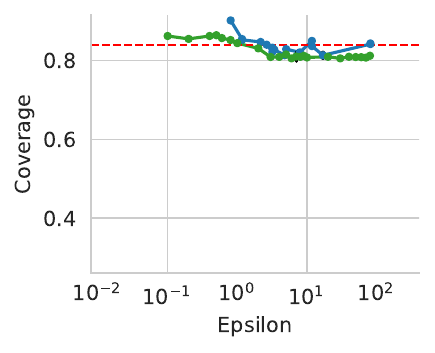}\vspace{-2mm}\label{fig:plot11_Coverage}}		
	\subfloat[VAE, Yelp]{\includegraphics[width=0.23\textwidth]{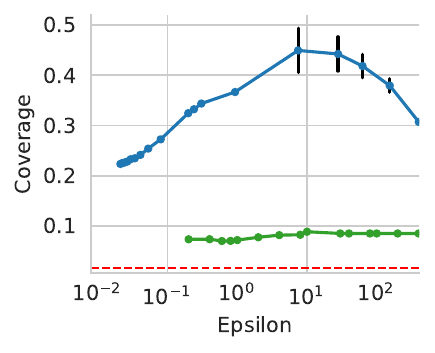}\vspace{-2mm}\label{fig:plot15_Coverage}}
	\caption{Coverage Across Different Models and Two Datasets with DPSGD and LDP}
	\label{fig:plots_Coverage_16}
\end{figure}

\paragraph{Coverage}

Figure~\ref{fig:plots_Coverage_16} presents the item coverage results, highlighting that the impact of privacy on item coverage is highly model- and dataset-dependent.
In particular, NCF/DPSGD (Figures \ref{fig:plot1_Coverage} and \ref{fig:plot3_Coverage}) significantly reduces coverage, indicating that this combination leads to a concentration of recommendations on a smaller subset of items.
In contrast, coverage increases for SVD/DPSGD/1M (Figure \ref{fig:plot9_Coverage}), BPR/1M (Figure \ref{fig:plot5_Coverage}), and VAE/Yelp (Figure \ref{fig:plot15_Coverage}) regardless of privacy levels.
For almost all other combinations, coverage remains close to the baseline, suggesting minimal impact.

\begin{figure}[t]
	\centering
	\subfloat[NCF, 1M]{\includegraphics[width=0.23\textwidth]{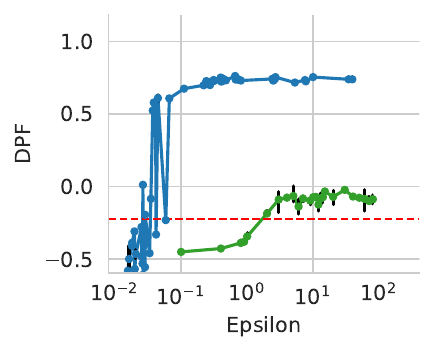}\vspace{-2mm}\label{fig:plot1_DPF}} 
	\subfloat[BPR, 1M]{\includegraphics[width=0.23\textwidth]{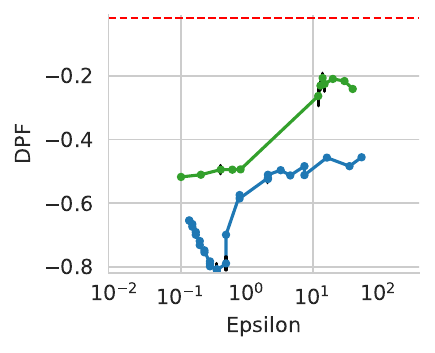}\vspace{-2mm}\label{fig:plot5_DPF}} 
	\subfloat[SVD, 1M]{\includegraphics[width=0.23\textwidth]{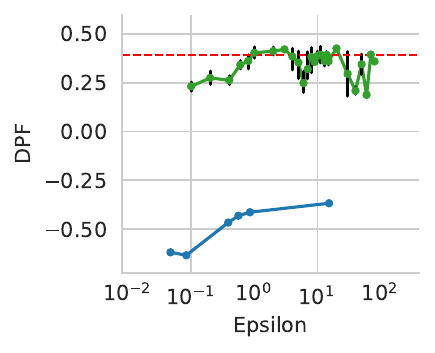}\vspace{-2mm}\label{fig:plot9_DPF}} 
	\subfloat[VAE, 1M]{\includegraphics[width=0.23\textwidth]{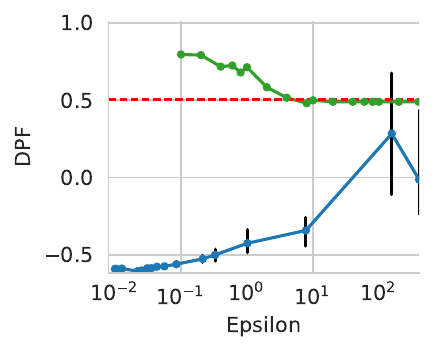}\vspace{-2mm}\label{fig:plot13_DPF}}
	
	\subfloat[NCF, Yelp]{\includegraphics[width=0.23\textwidth]{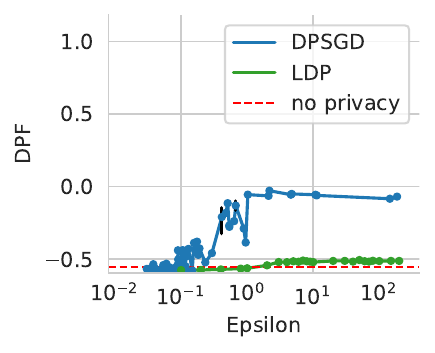}\vspace{-2mm}\label{fig:plot3_DPF}} 
	\subfloat[BPR, Yelp]{\includegraphics[width=0.23\textwidth]{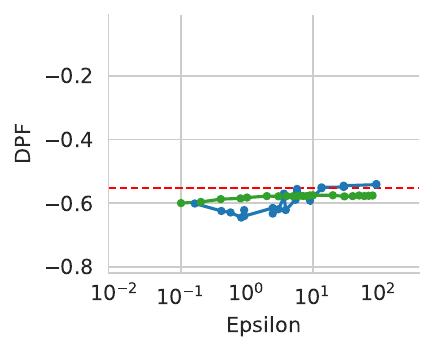}\vspace{-2mm}\label{fig:plot7_DPF}} 
	\subfloat[SVD, Yelp]{\includegraphics[width=0.23\textwidth]{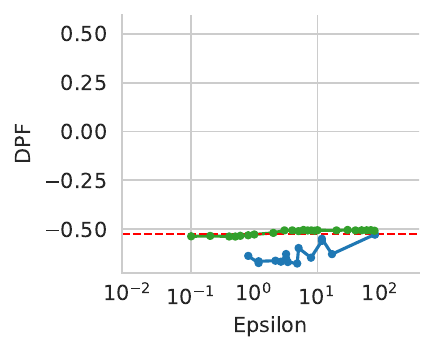}\vspace{-2mm}\label{fig:plot11_DPF}} 		
	\subfloat[VAE, Yelp]{\includegraphics[width=0.23\textwidth]{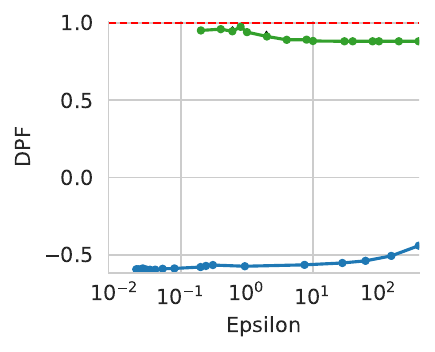}\vspace{-2mm}\label{fig:plot15_DPF}}
	\caption{DPF Across Different Models and Two Datasets with DPSGD and LDP}	
	\label{fig:plots_DPF_16}
\end{figure}

\paragraph{Deviation from Producer Fairness (DPF)}
%\iw{spell out DPF in the heading}
Fairness disparities are measured by DPF (Figure \ref{fig:plots_DPF_16}) and item exposure.
LDP mostly preserves baseline fairness, while DPSGD’s impact varies by dataset.
%In particular, the baseline DPF for Yelp (e.g., Figure \ref{fig:plot3_DPF}) is negative, indicating an inherent disparity in exposure across user groups.
On Yelp, LDP (except VAE/Yelp in Figure \ref{fig:plot15_DPF}) stays close to the baseline, indicating a neutral or mitigating effect on fairness disparity.
On 1M, LDP maintains a fairness level of DPF close to the baseline, except for the BPR/1M (Figure \ref{fig:plot5_DPF}) combination, where deviation occurs.
Under DPSGD on 1M (e.g., Figures~\ref{fig:plot5_DPF} and~\ref{fig:plot13_DPF}), DPF decreases and becomes negative under strong privacy constraints, and increases again as the privacy budget grows.
In contrast, on Yelp under DPSGD, DPF remains consistently negative across different privacy levels, with minimal variation as $\epsilon$ changes.
An exception is observed for NCF/Yelp/DPSGD, where DPF increases slightly with larger $\epsilon$, yet it still remains negative (Figure~\ref{fig:plot3_DPF}). 
These results indicate that the impact of privacy-enhancing mechanisms on fairness can differ across datasets.

\begin{figure}[t]
	\centering
	\subfloat[NCF, 1M]{\includegraphics[width=0.23\textwidth]{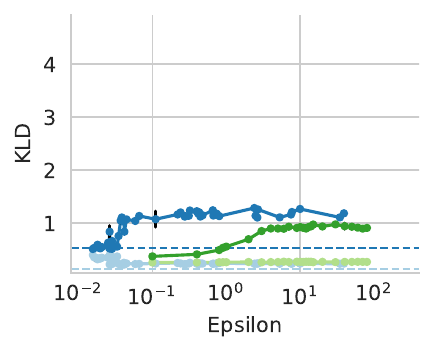}\vspace{-2mm}\label{fig:plot1_kld_per_category}} 
	\subfloat[BPR, 1M]{\includegraphics[width=0.23\textwidth]{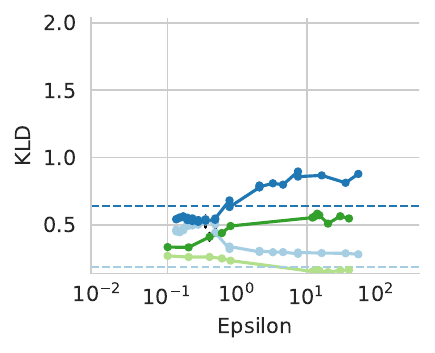}\vspace{-2mm}\label{fig:plot5_kld_per_category}} 
	\subfloat[SVD, 1M]{\includegraphics[width=0.23\textwidth]{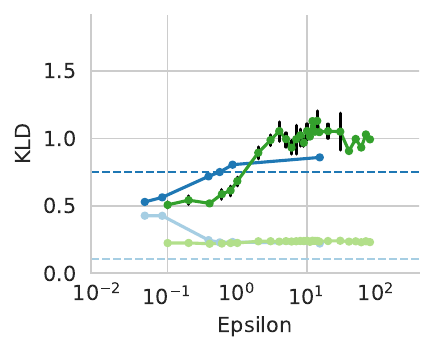}\vspace{-2mm}\label{fig:plot9_kld_per_category}} 
	\subfloat[VAE, 1M]{\includegraphics[width=0.23\textwidth]{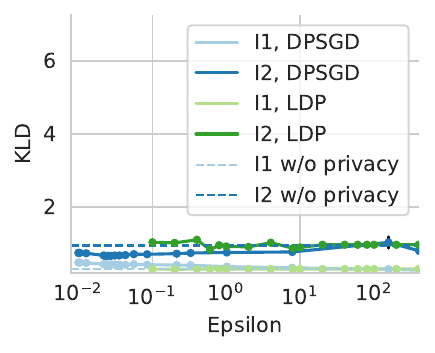}\vspace{-2mm}\label{fig:plot13_kld_per_category}} 
	
	\subfloat[NCF, Yelp]{\includegraphics[width=0.23\textwidth]{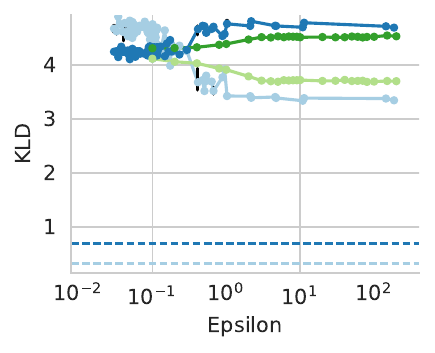}\vspace{-2mm}\label{fig:plot3_kld_per_category}} 
	\subfloat[BPR, Yelp]{\includegraphics[width=0.23\textwidth]{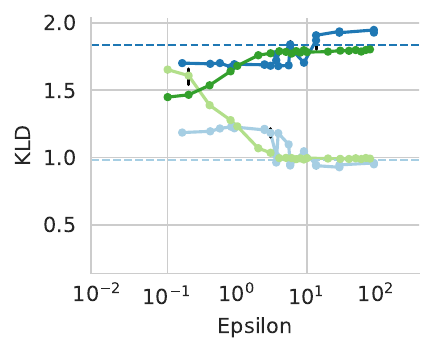}\vspace{-2mm}\label{fig:plot7_kld_per_category}} 		
	\subfloat[SVD, Yelp]{\includegraphics[width=0.23\textwidth]{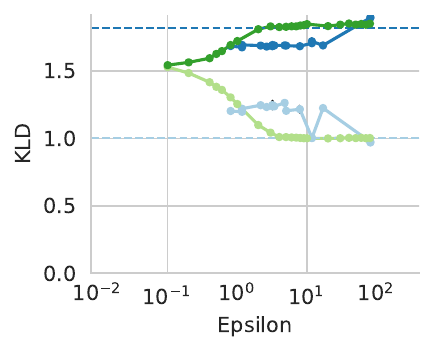}\vspace{-2mm}\label{fig:plot11_kld_per_category}}		
	\subfloat[VAE, Yelp]{\includegraphics[width=0.23\textwidth]{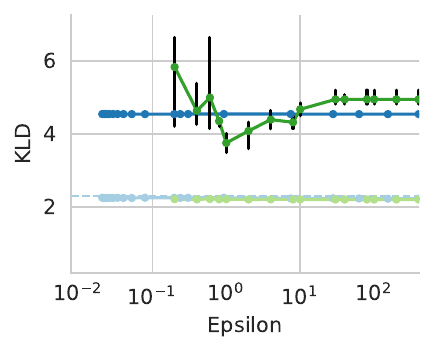}\vspace{-2mm}\label{fig:plot15_kld_per_category}} 
	
	\caption{Miscalibration (KLD) for popular items (I1, light colors) and unpopular items (I2, dark colors)}
	\label{fig:plots_kld_per_category_16}
\end{figure}

\subsubsection{Privacy-Bias Trade-off by Item Popularity}

To further understand the effect of privacy mechanisms, we analyze bias metrics separately for popular and unpopular items.

\paragraph{Miscalibration (KLD) by Item Popularity}
Miscalibration, measured by \ac{KLD} on popular and unpopular items in Figure \ref{fig:plots_kld_per_category_16}, indicates that popular items are consistently better calibrated than unpopular items, as evidenced by KLD values that are systematically closer to zero (\cref{fig:plots_kld_per_category_16}). 
This suggests a smaller divergence between user history and recommendations for popular items.
In some algorithm-dataset combinations (eg, BPR/Yelp (Figure \ref{fig:plot7_kld_per_category}) and SVD/Yelp (Figure \ref{fig:plot11_kld_per_category})), lower privacy levels (higher $\epsilon$) improve calibration for popular items, implying that differential privacy noise can degrade calibration specifically for popular items.
Conversely, unpopular items tend to be better calibrated under high privacy levels (smaller $\epsilon$), implying that stronger privacy constraints improve the recommendations for less frequent items.

\begin{figure}[t]
	\centering
	\subfloat[NCF, 1M]{\includegraphics[width=0.23\textwidth]{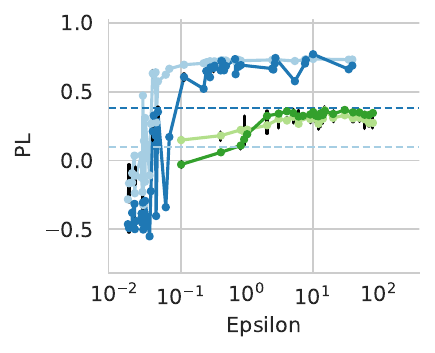}\vspace{-2mm}\label{fig:plot1_PL_per_category}} 
	\subfloat[BPR, 1M]{\includegraphics[width=0.23\textwidth]{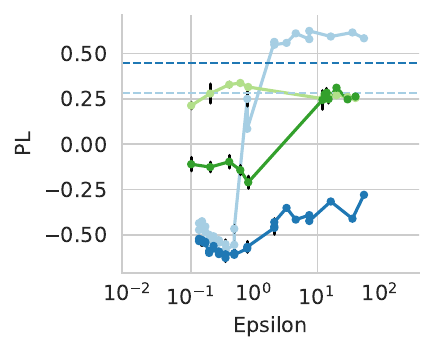}\vspace{-2mm}\label{fig:plot5_PL_per_category}} 
	\subfloat[SVD, 1M]{\includegraphics[width=0.23\textwidth]{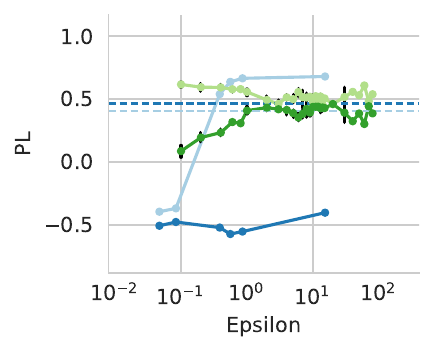}\vspace{-2mm}\label{fig:plot9_PL_per_category}} 
	\subfloat[VAE, 1M]{\includegraphics[width=0.23\textwidth]{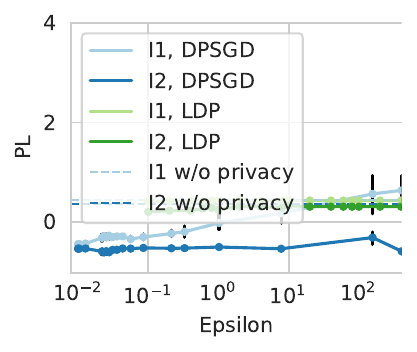}\vspace{-2mm}\label{fig:plot13_PL_per_category}} 
	
	\subfloat[NCF, Yelp]{\includegraphics[width=0.23\textwidth]{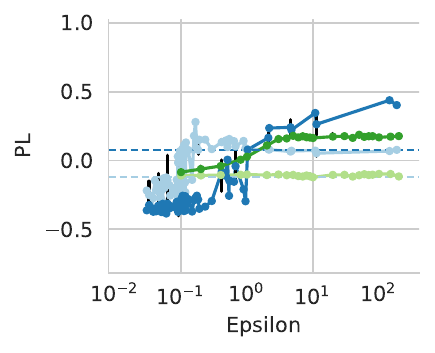}\vspace{-2mm}\label{fig:plot3_PL_per_category}} 
	\subfloat[BPR, Yelp]{\includegraphics[width=0.23\textwidth]{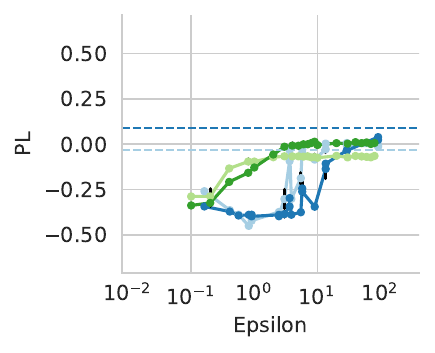}\vspace{-2mm}\label{fig:plot7_PL_per_category}} 
	\subfloat[SVD, Yelp]{\includegraphics[width=0.23\textwidth]{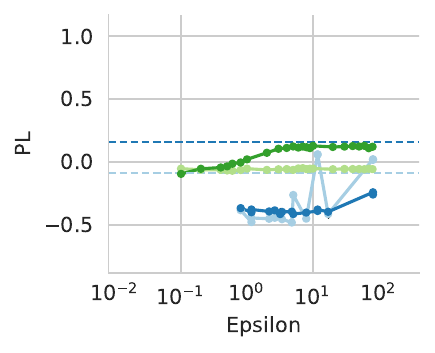}\vspace{-2mm}\label{fig:plot11_PL_per_category}} 
	\subfloat[VAE, Yelp]{\includegraphics[width=0.23\textwidth]{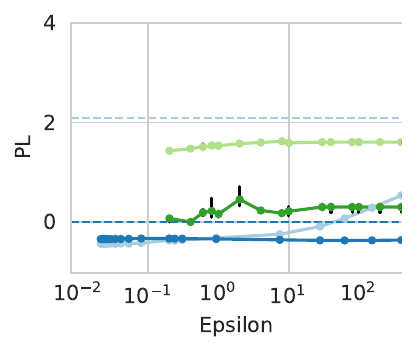}\vspace{-2mm}\label{fig:plot15_PL_per_category}} 
	\caption{Popularity lift for popular items (I1, light colors) and unpopular items (I2, dark colors)}
	\label{fig:plots_PL_per_category_16}
\end{figure}

\paragraph{Popularity Lift (PL) by Item Popularity}

Figure~\ref{fig:plots_PL_per_category_16} shows that 
%analyzes the impact of LDP and DPSGD with varying privacy levels ($\epsilon$) on item popularity.
the effect of \ac{ldp} and \ac{DPSGD} with varying privacy levels ($\epsilon$) on popularity lift (PL) depends on the popularity of items as well as on the recommender algorithm.
Across all models and datasets, as $\epsilon$ increases, the popularity lift for popular items (I1) grows, indicating that less privacy amplifies popularity bias.
Conversely, stronger privacy (smaller $\epsilon$) tends to suppress this bias, reducing the relative advantage of popular items.
%Among models, NCF/1M (Figure \ref{fig:plot1_PL_per_category}) shows the most significant sensitivity to $\epsilon$, with both DPSGD and LDP producing substantial increases in PL as privacy relaxes.
%BPR (Figures \ref{fig:plot5_PL_per_category} and \ref{fig:plot7_PL_per_category}) and SVD (Figures \ref{fig:plot9_PL_per_category} and \ref{fig:plot11_PL_per_category}) show moderate changes, though BPR is more variable on the Yelp dataset.
%VAE (Figures \ref{fig:plot13_PL_per_category} and \ref{fig:plot15_PL_per_category}) is the most robust, showing minimal changes in PL, especially on Yelp.
Overall, differential privacy mechanisms, especially under stricter settings (lower $\epsilon$), can help mitigate popularity bias by preventing models from disproportionately emphasizing popular items.

%Also, as yelp has a much larger item space, the user preferences are spread across many items, and the relative dominance of popular items is smaller than 1M, where fewer items concentrate user attention.
%On average, LDP is closer to the baseline than DPSGD, indicating that LDP has a more neutral effect on popularity bias.
% 1M dataset, DPSGD: when privacy is higher, the I1 experienced more popularity lift from negative to positive 
% 11M, ldp: THE CHANGES FOR i2 IS MORE SHARP, i1 stays almost similar with varying epsilon, but I2 PL will be increased 
% Yelp, DPSGD: Both I1 and I2 experience the same pattern, increasing epsilon for NCF, changing PL for both I1 and I2, and making it more positive for the rest. It is still negative and near zero, but goes to positive
% Yelp, LDp: By increasing privacy, it goes to the positive side for both I1 and I2, but the changes are more stable for VAE and also NCF, and more changes for BPR and the SVD

\begin{figure}[t]
	\centering
	\subfloat[NCF with DPSGD]{\includegraphics[width=0.23\textwidth]{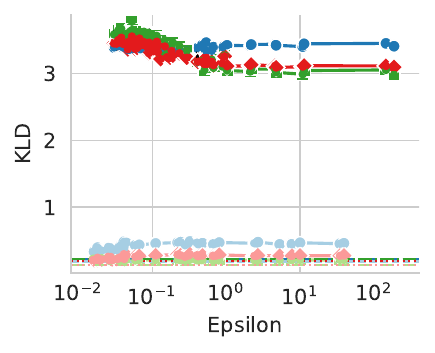}\label{fig:plot1_kld_user_based}} 
	\subfloat[BPR with DPSGD]{\includegraphics[width=0.23\textwidth]{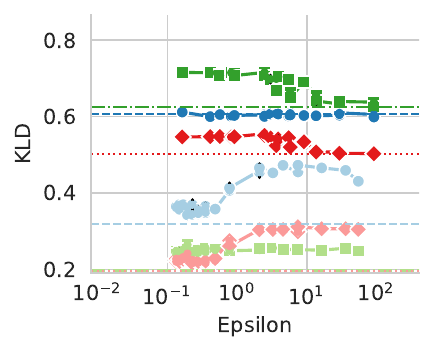}\label{fig:plot3_kld_user_based}} 
	\subfloat[SVD with DPSGD]{\includegraphics[width=0.23\textwidth]{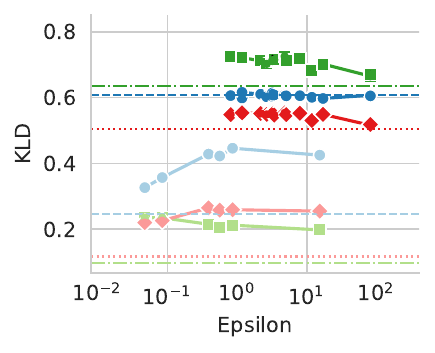}\label{fig:plot5_kld_user_based}}
	\subfloat[VAE with DPSGD]{\includegraphics[width=0.23\textwidth]{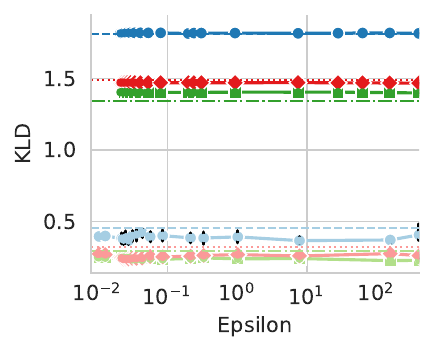}\label{fig:plot7_kld_user_based}}
	
	\subfloat[NCF with LDP]{\includegraphics[width=0.23\textwidth]{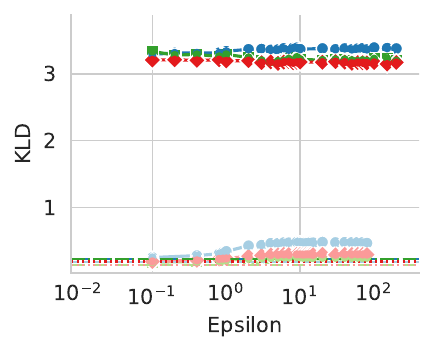}\label{fig:plot2_kld_user_based}} 
	\subfloat[BPR with LDP]{\includegraphics[width=0.23\textwidth]{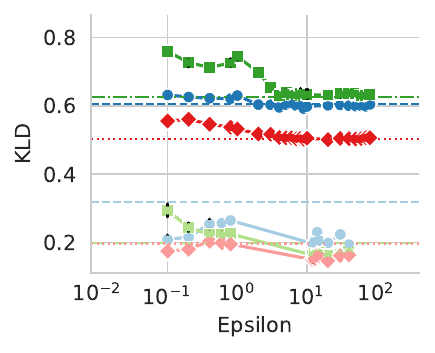}\label{fig:plot4_kld_user_based}} 
	\subfloat[SVD with LDP]{\includegraphics[width=0.23\textwidth]{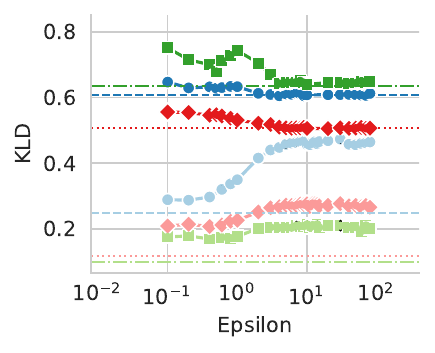}\label{fig:plot6_kld_user_based}}
	\subfloat[VAE with LDP]{\includegraphics[width=0.23\textwidth]{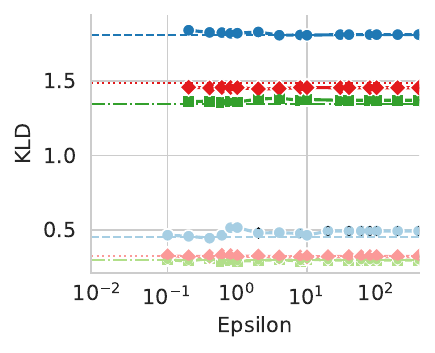}\label{fig:plot8_kld_user_based}} 
	
	\caption{Miscalibration (KLD) by user group (niche (blue circles), diverse (red diamonds), and blockbuster (green squares)) and dataset (Yelp (dark colors) and 1M (light colors))}
	\label{fig:plots_kld_user_based_8}
\end{figure}

\subsubsection{Privacy-Bias Trade-off by User Type}

In this section, we analyze the bias metrics across different user types (niche, blockbuster and diverse), to examine whether bias varies depending on the user type.

\paragraph{Miscalibration (KLD) by User Type}

Miscalibration varies across both user groups and datasets (see Figure~\ref{fig:plots_kld_user_based_8}).
On Yelp, miscalibration is highest for blockbuster users with BPR (Figures \ref{fig:plot3_kld_user_based} and \ref{fig:plot4_kld_user_based}) and SVD models (Figures \ref{fig:plot5_kld_user_based} and \ref{fig:plot6_kld_user_based}), whereas with NCF (Figures \ref{fig:plot1_kld_user_based} and \ref{fig:plot2_kld_user_based}) and VAE (Figures \ref{fig:plot7_kld_user_based} and \ref{fig:plot8_kld_user_based}) it is highest for niche users.
On 1M, miscalibration is highest for niche users across all models and both privacy mechanisms (Figure \ref{fig:plots_kld_user_based_8}).

%For diverse users in Yelp, high KLD values are observed in BPR, SVD and VAE.
%Over all usesr types, the Yelp dataset shows systematically higher miscalibration compared to 1M, which may be attributed to its higher sparsity.
Among all model-dataset combinations and user types, NCF on Yelp (Figures \ref{fig:plot1_kld_user_based} and \ref{fig:plot2_kld_user_based}), shows the highest KLD (above 3), whereas all other cases remain below 1, indicating severe miscalibration for this setting.
The effect is especially pronounced for niche users.
This is likely because NCF achieves high NDCG by heavily favoring popular items, which inherently misaligns with niche users' preferences, resulting in significant distributional mismatch and thus high KLD.
%High privacy improves calibration in most cases. 
%An exception is BPR/LDP, shown in Figure \ref{fig:plot4_kld_user_based}, where high privacy does not lead to improved calibration.
On Yelp under \ac{ldp}, BPR (Figure \ref{fig:plot4_kld_user_based}) and SVD (Figure \ref{fig:plot6_kld_user_based}) have comparable overall KLD and also show the largest KLD disparity across user groups among all models. 
%\iw{something is wrong with this analysis: bpr/ldp is only 14d, not 14c and 14d; 14c/d looks very similar to 14e/f, but we do not mention it}
%\shp{updated, it is very similar for 14c and 14f for LDP case}

\begin{figure}[t]
	\centering
	\subfloat[NCF with DPSGD]{\includegraphics[width=0.23\textwidth]{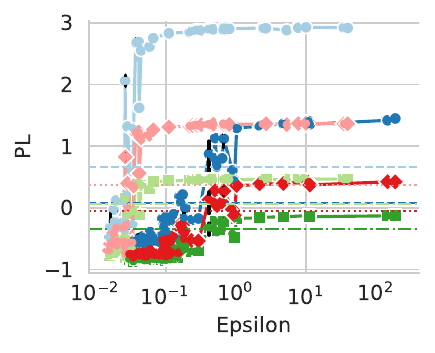}\label{fig:plot1_PL_user_based}} 
	\subfloat[BPR with DPSGD]{\includegraphics[width=0.23\textwidth]{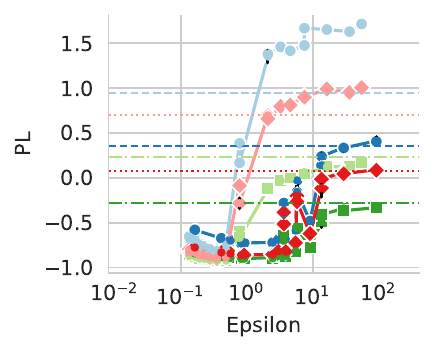}\label{fig:plot3_PL_user_based}} 
	\subfloat[SVD with DPSGD]{\includegraphics[width=0.23\textwidth]{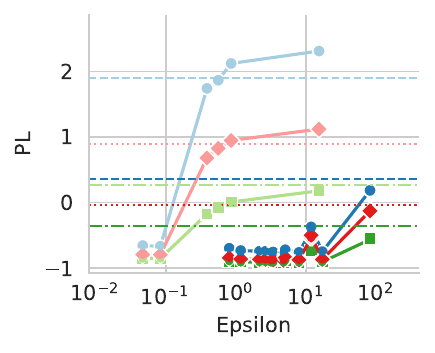}\label{fig:plot5_PL_user_based}}
	\subfloat[VAE with DPSGD]{\includegraphics[width=0.23\textwidth]{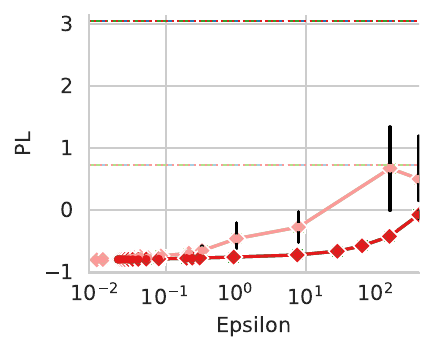}\label{fig:plot7_PL_user_based}}
	
	\subfloat[NCF with LDP]{\includegraphics[width=0.23\textwidth]{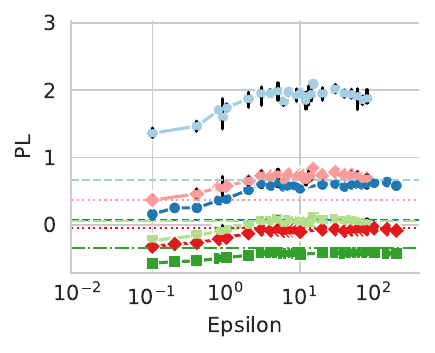}\label{fig:plot2_PL_user_based}} 
	\subfloat[BPR with LDP]{\includegraphics[width=0.23\textwidth]{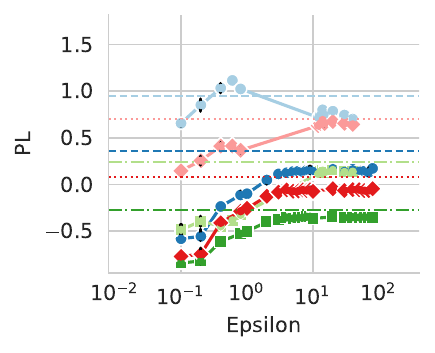}\label{fig:plot4_PL_user_based}}
	\subfloat[SVD with LDP]{\includegraphics[width=0.23\textwidth]{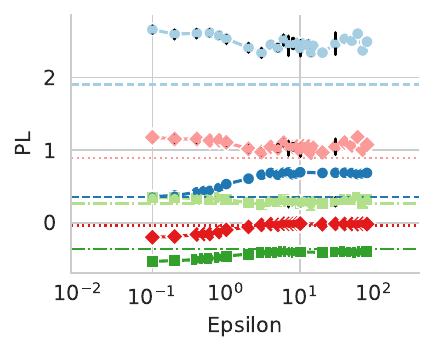}\label{fig:plot6_PL_user_based}}
	\subfloat[VAE with LDP]{\includegraphics[width=0.23\textwidth]{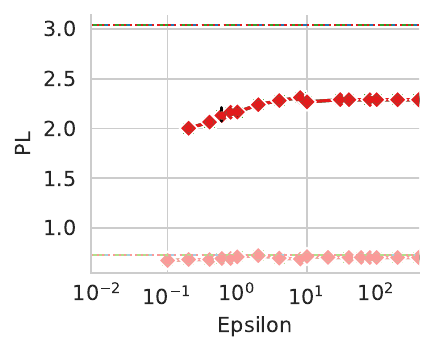}\label{fig:plot8_PL_user_based}} 
	
	\caption{Popularity lift by user group (niche (blue circles), diverse (red diamonds), and blockbuster (green squares)) and dataset (Yelp (dark colors) and 1M (light colors))}
	\label{fig:plots_PL_user_based_8}
\end{figure}

\paragraph{Popularity Lift (PL) by User Type}
Figure~\ref{fig:plots_PL_user_based_8} examines how \ac{PL} varies for different user types, showing that lower privacy levels generally lead to higher \ac{PL} across most configurations.
Niche users consistently experience the highest popularity lift, suggesting their recommendations are disproportionately skewed toward popular items.
An exception occurs for SVD/LDP (Figure \ref{fig:plot6_PL_user_based}), where privacy level does not significantly influence PL.
For niche users, the NCF/DPSGD (Figure \ref{fig:plot1_PL_user_based}) and SVD/DPSGD (Figure \ref{fig:plot5_PL_user_based}), PL jumps dramatically at higher epsilon (weaker privacy), reaching values above 2.5.
This shows that when privacy constraints are relaxed, the model increasingly pushes popular items to niche users, amplifying the popularity bias.
Blockbuster users show moderate PL in most settings, but exceptionally high in VAE/LDP (Figure \ref{fig:plot8_PL_user_based}), suggesting that even popular-item-preferring users receive disproportionately popular recommendations.
Diverse users are less affected by popularity amplification, possibly due to their more balanced historical behavior.

\begin{figure}[t]
	\centering
	\subfloat[NCF with DPSGD]{\includegraphics[width=0.23\textwidth]{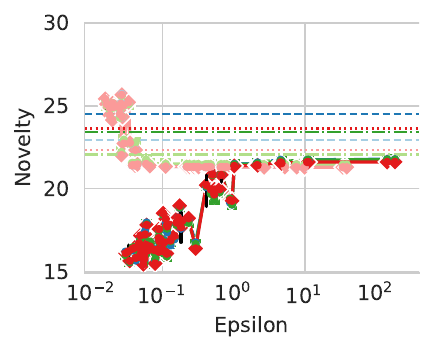}\label{fig:plot1_novelty_user_based}}
	\subfloat[BPR with DPSGD]{\includegraphics[width=0.23\textwidth]{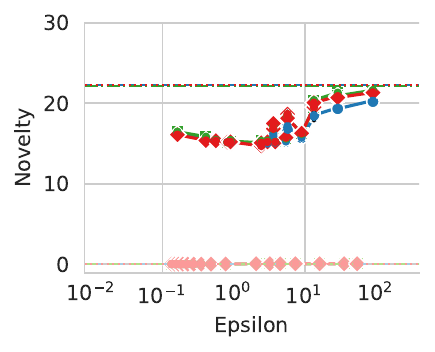}\label{fig:plot3_novelty_user_based}}
	\subfloat[SVD with DPSGD]{\includegraphics[width=0.23\textwidth]{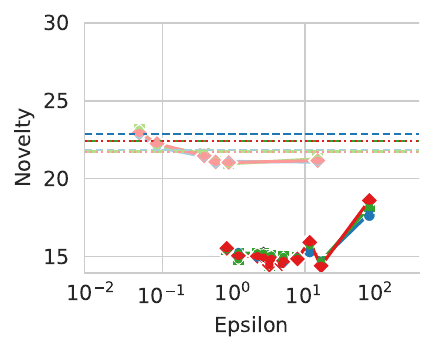}\label{fig:plot5_novelty_user_based}}
	\subfloat[VAE with DPSGD]{\includegraphics[width=0.23\textwidth]{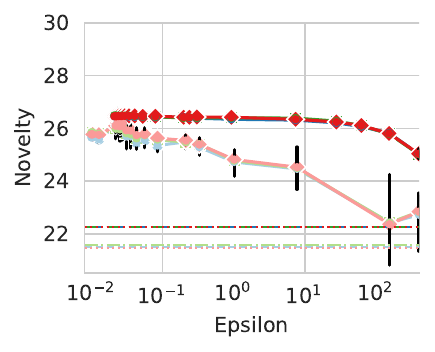}\label{fig:plot7_novelty_user_based}}
	
	\subfloat[NCF with LDP]{\includegraphics[width=0.23\textwidth]{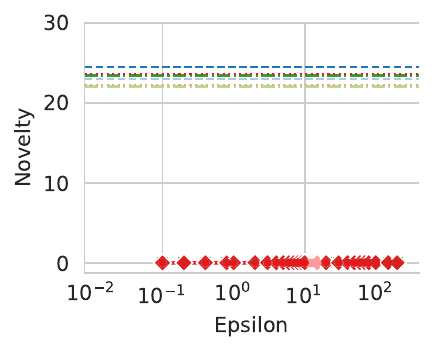}\label{fig:plot2_novelty_user_based}}
	\subfloat[BPR with LDP]{\includegraphics[width=0.23\textwidth]{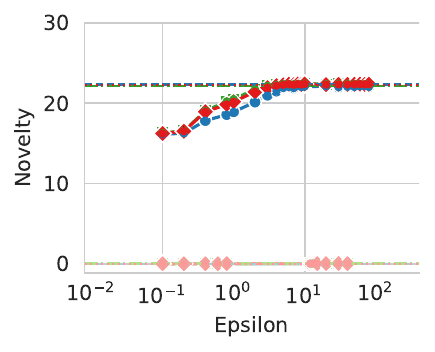}\label{fig:plot4_novelty_user_based}} 
	\subfloat[SVD with LDP]{\includegraphics[width=0.23\textwidth]{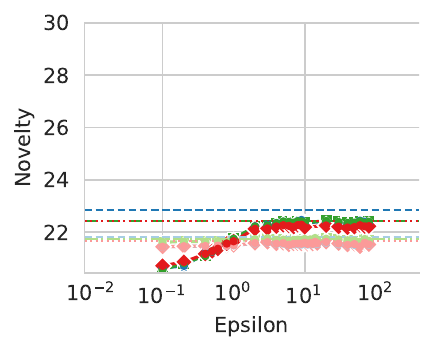}\label{fig:plot6_novelty_user_based}}
	\subfloat[VAE with LDP]{\includegraphics[width=0.23\textwidth]{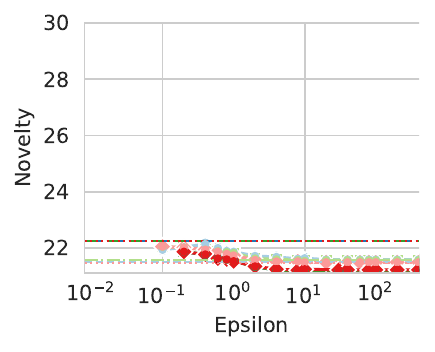}\label{fig:plot8_novelty_user_based}}
	\caption{Novelty by user group (niche (blue circles), blockbuster (red diamonds), and diverse (green squares)) and dataset (Yelp (dark colors) and 1M (light colors))}
	\label{fig:plots_novelty_user_based_8}
\end{figure}

\paragraph{Novelty by User Type}
Novelty across different user types, as shown in Figure~\ref{fig:plots_novelty_user_based_8}, shows that its dependency on $\epsilon$ differs between the 1M and Yelp datasets.
Intuitively, we expected that novelty would be higher for higher privacy levels because of higher noise levels.
This is indeed the case for 1M, however, for Yelp, contrary to expectation, novelty generally increases with lower privacy levels (higher $\epsilon$).
% (except for VAE)

%For Yelp, lower privacy increases novelty, indicating that more relaxed privacy constraints encourage recommendations of unexpected items.
%For 1M, privacy level generally has no effect on novelty, with two exceptions: SVD/DPSGD in Figure \ref{fig:plot5_novelty_user_based}, and NCF/DPSGD in Figure \ref{fig:plot1_novelty_user_based}, exhibit lower novelty at lower privacy levels.
%\iw{again, no mention of user types, but repetition of overall results from before.}

\begin{figure}[t]
	\centering
	\subfloat[NCF with DPSGD]{\includegraphics[width=0.23\textwidth]{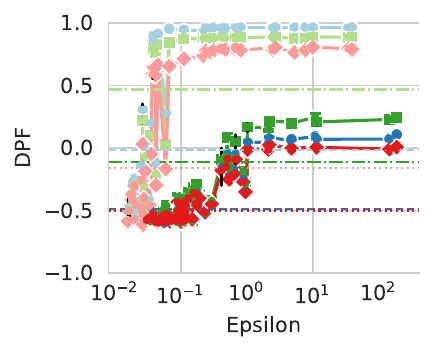}\label{fig:plot1_DPF_user_based}} 
	\subfloat[BPR with DPSGD]{\includegraphics[width=0.23\textwidth]{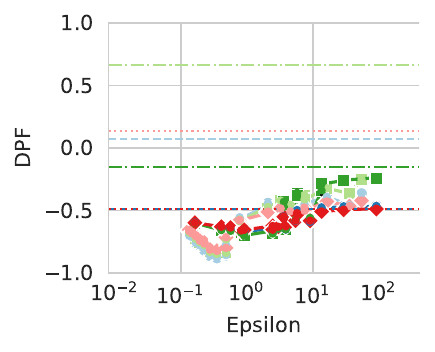}\label{fig:plot3_DPF_user_based}}
	\subfloat[SVD with DPSGD]{\includegraphics[width=0.23\textwidth]{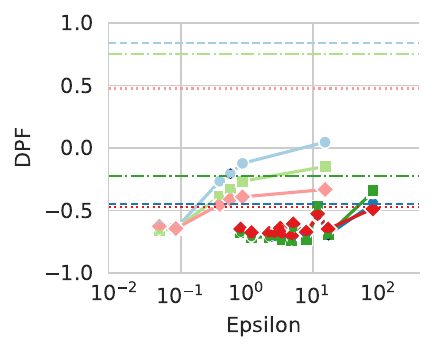}\label{fig:plot5_DPF_user_based}}
	\subfloat[VAE with DPSGD]{\includegraphics[width=0.23\textwidth]{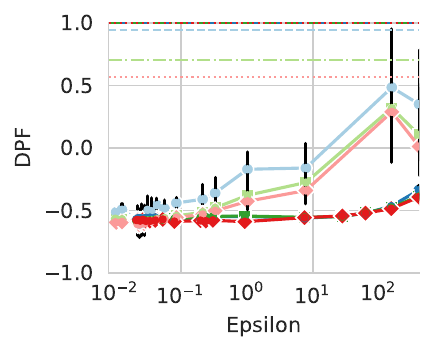}\label{fig:plot7_DPF_user_based}}
	
	\subfloat[NCF with LDP]{\includegraphics[width=0.23\textwidth]{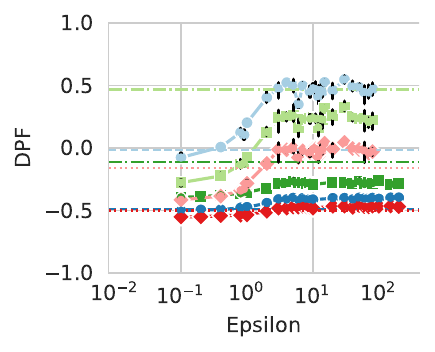}\label{fig:plot2_DPF_user_based}} 
	\subfloat[BPR with LDP]{\includegraphics[width=0.23\textwidth]{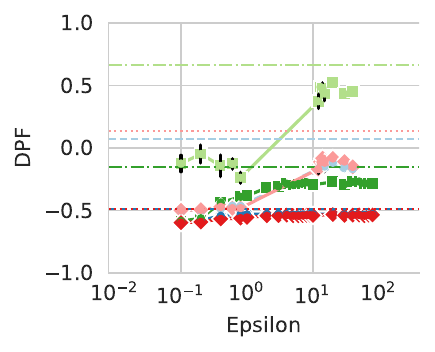}\label{fig:plot4_DPF_user_based}}
	\subfloat[SVD with LDP]{\includegraphics[width=0.23\textwidth]{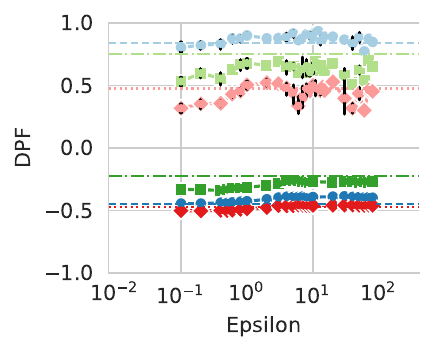}\label{fig:plot6_DPF_user_based}}
	\subfloat[VAE with LDP]{\includegraphics[width=0.23\textwidth]{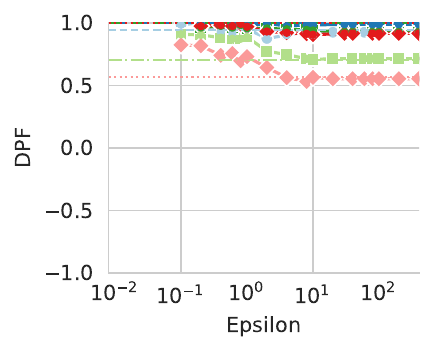}\label{fig:plot8_DPF_user_based}} 
	
	\caption{DPF by user group (niche (blue circles), diverse (red diamonds), and blockbuster (green squares)) and dataset (Yelp (dark colors) and 1M (light colors))}
	\label{fig:plots_DPF_user_based_8}
\end{figure}

\paragraph{Deviation from Producer Fairness (DPF) by User Type}
%\iw{spell out DPF in heading}
DPF across different user types (Figure \ref{fig:plots_DPF_user_based_8}) shows that lower privacy generally increases DPF, indicating that higher privacy levels reduce the exposure of popular items.
%The 1M dataset shows higher DPF than Yelp, indicating a stronger bias toward popular items in this dataset.
%An exception occurs for SVD/LDP in Figure \ref{fig:plot6_DPF_user_based}, where privacy level has no impact on DPF.
Niche users show the highest DPF in 1M, while diverse users have the highest DPF in Yelp.
This indicates that in a 1M dataset, niche users received more popular items than non-popular ones. 
%DPSGD tends to produce DPF values closer to 0, whereas LDP favors popular items more than DPSGD, indicating that LDP generally skews recommendations toward already popular content.
For all models, datasets, and privacy mechanisms, blockbuster users received more diverse (combination of popular and non-popular items) recommendations; therefore, they have the lowest DPF in different $\epsilon$ values.

\begin{figure}[t]
	\centering
	\subfloat[NCF with DPSGD]{\includegraphics[width=0.23\textwidth]{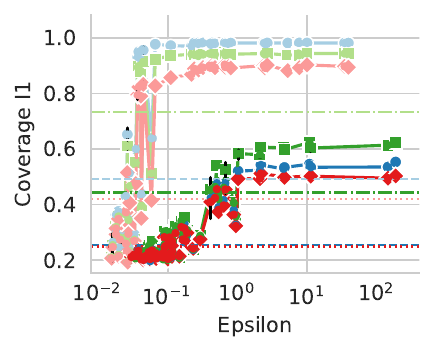}\label{fig:plot1_normalized_coverage_category_user_based}} 
	\subfloat[BPR with DPSGD]{\includegraphics[width=0.23\textwidth]{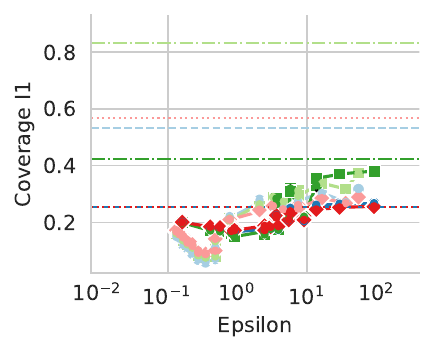}\label{fig:plot3_normalized_coverage_category_user_based}} 
	\subfloat[SVD with DPSGD]{\includegraphics[width=0.23\textwidth]{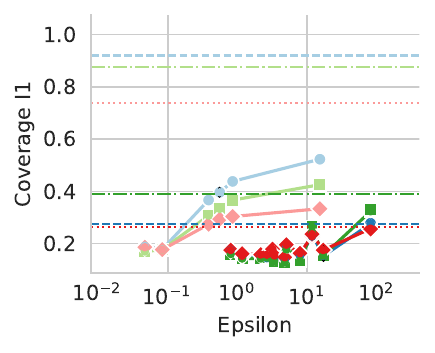}\label{fig:plot5_normalized_coverage_category_user_based}}
	\subfloat[VAE with DPSGD]{\includegraphics[width=0.23\textwidth]{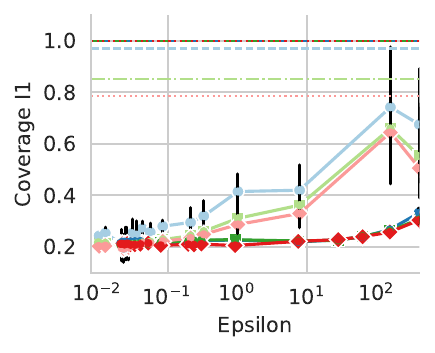}\label{fig:plot7_normalized_coverage_category_user_based}}
	
	\subfloat[NCF with LDP]{\includegraphics[width=0.23\textwidth]{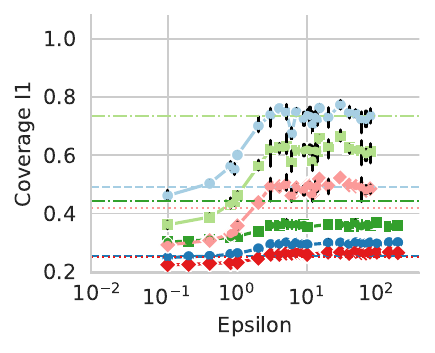}\label{fig:plot2_normalized_coverage_category_user_based}} 
	\subfloat[BPR with LDP]{\includegraphics[width=0.23\textwidth]{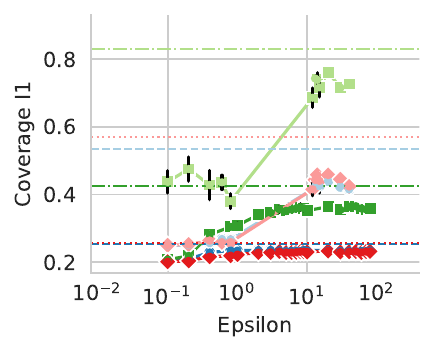}\label{fig:plot4_normalized_coverage_category_user_based}}
	\subfloat[SVD with LDP]{\includegraphics[width=0.23\textwidth]{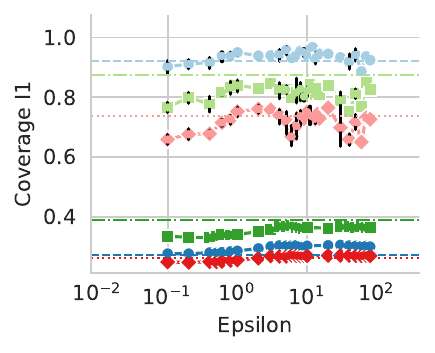}\label{fig:plot6_normalized_coverage_category_user_based}}
	\subfloat[VAE with LDP]{\includegraphics[width=0.23\textwidth]{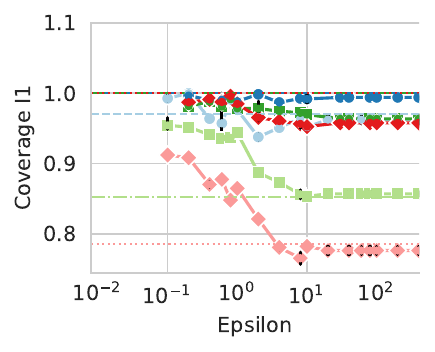}\label{fig:plot8_normalized_coverage_category_user_based}} 
	
	\caption{Coverage I1 by user group (niche (blue circles), diverse (red diamonds), and blockbuster (green squares)) and dataset (Yelp (dark colors) and 1M (light colors))}
	\label{fig:plots_normalized_coverage_category_user_based_8}
\end{figure}

\paragraph{Coverage I1}

In Figure~\ref{fig:plots_normalized_coverage_category_user_based_8}, coverage I1 measures the proportion of popular items in recommendation lists (the results for coverage I2, representing the proportion of unpopular items, are not shown because they are exactly reversed).
The results show that the user type receiving the highest coverage varies between the Yelp and 1M datasets.
Coverage is generally highest for niche users, suggesting that they receive a more diverse set of popular items. 
However, for Yelp, blockbuster users experience the highest coverage of popular items.
%\iw{I do not understand the logic in the last two sentences: when niche users have high coverage, you say they receive "diverse" recommendations, but when blockbuster users have high coverage, you say their recommendations are concentrated on popular items. This does not seem to make sense.}
%\shp{They represent the same idea: when Coverage I1 is high, it means there is greater diversity within I1, which are the popular items. (However, I have edited my text to make this clearer.)}
%Overall, the 1M dataset shows higher item coverage than Yelp, with coverage increasing as privacy constraints are relaxed.
%LDP tends to recommend more popular items than DPSGD (except NCF/LDP in Figure \ref{fig:plot2_normalized_coverage_category_user_based}).
%\iw{this sentence does not follow logically from the one before}
%The results for coverage I2 are exactly reversed.

\subsection{Utility-Bias Trade-off}
\label{sec:utility-bias}
\begin{figure}[t]
		\centering
		\subfloat{\includegraphics[width=0.23\textwidth]{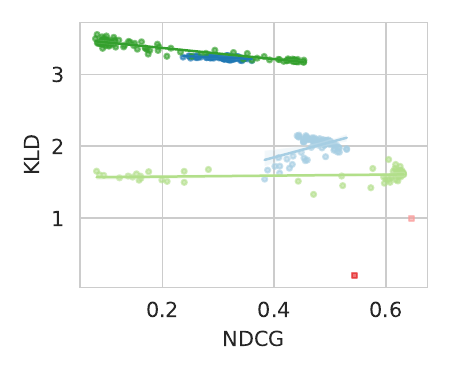}\label{fig:plot4_utility_bias}}
		\subfloat{\includegraphics[width=0.23\textwidth]{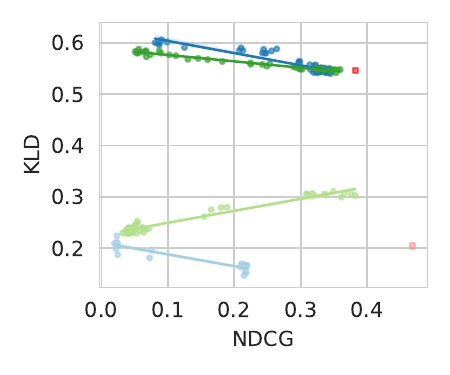}\label{fig:plot8_utility_bias}} 
		\subfloat{\includegraphics[width=0.23\textwidth]{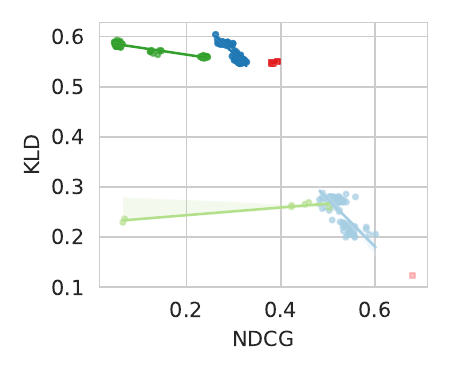}\label{fig:plot12_utility_bias}}
		\subfloat{\includegraphics[width=0.23\textwidth]{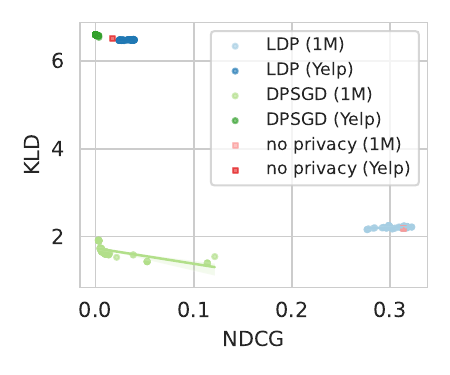}\label{fig:plot16_utility_bias}}

		\subfloat{\includegraphics[width=0.23\textwidth]{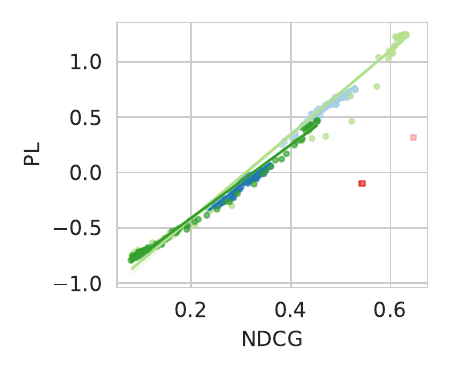}\label{fig:plot3_utility_bias}} 
		\subfloat{\includegraphics[width=0.23\textwidth]{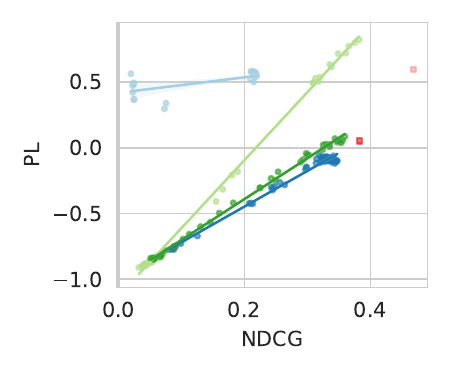}\label{fig:plot7_utility_bias}}
		\subfloat{\includegraphics[width=0.23\textwidth]{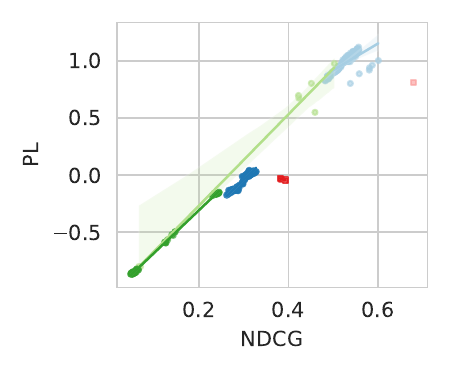}\label{fig:plot11_utility_bias}} 
		\subfloat{\includegraphics[width=0.23\textwidth]{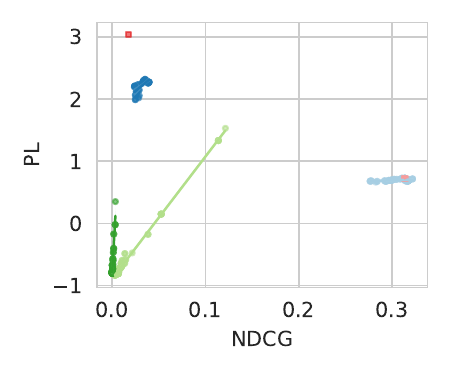}\label{fig:plot15_utility_bias}} 

		\subfloat{\includegraphics[width=0.23\textwidth]{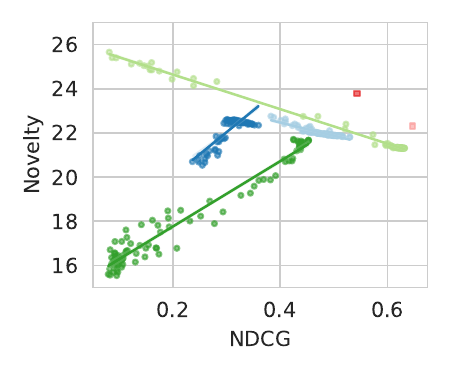}\label{fig:plot2_utility_bias}} 
		\subfloat{\includegraphics[width=0.23\textwidth]{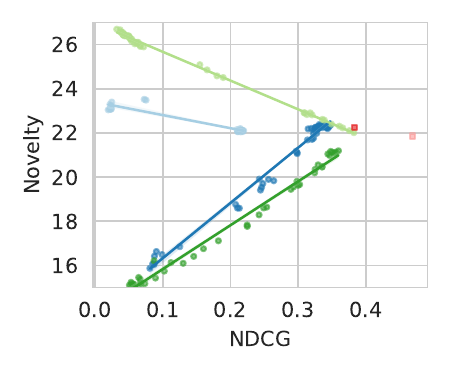}\label{fig:plot6_utility_bias}}
		\subfloat{\includegraphics[width=0.23\textwidth]{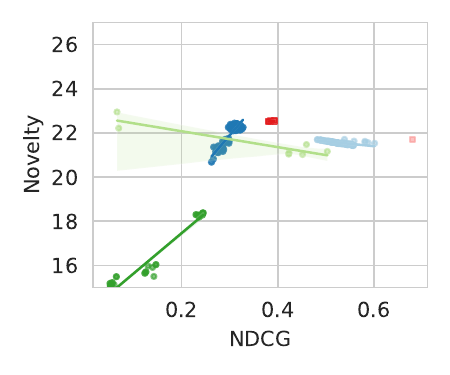}\label{fig:plot10_utility_bias}} 
		\subfloat{\includegraphics[width=0.23\textwidth]{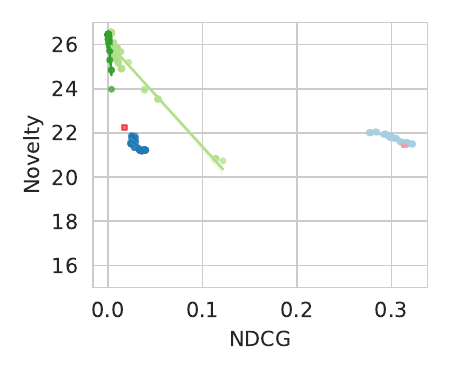}\label{fig:plot14_utility_bias}} 

		\subfloat{\includegraphics[width=0.23\textwidth]{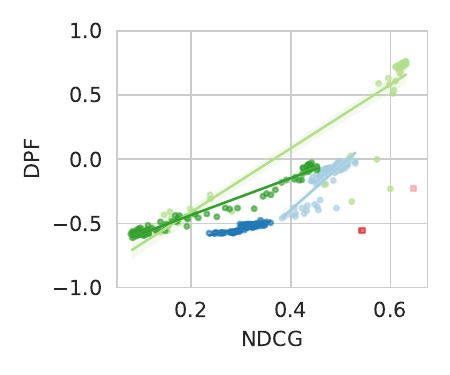}\label{fig:plot1_utility_bias}} 
		\subfloat{\includegraphics[width=0.23\textwidth]{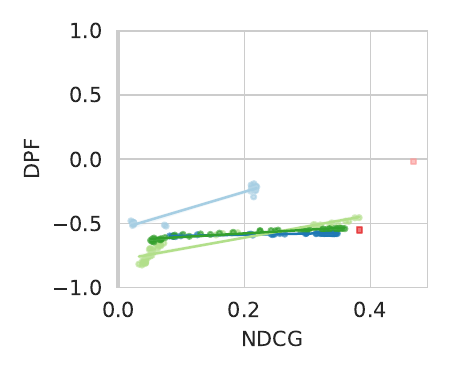}\label{fig:plot5_utility_bias}}
		\subfloat{\includegraphics[width=0.23\textwidth]{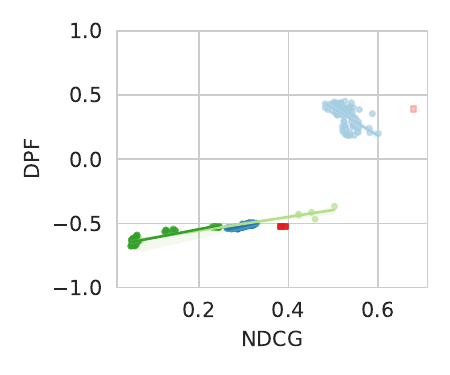}\label{fig:plot9_utility_bias}} 
		\subfloat{\includegraphics[width=0.23\textwidth]{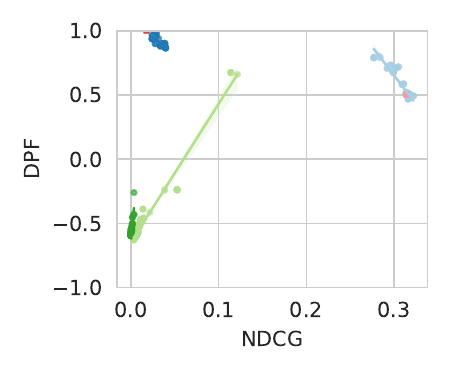}\label{fig:plot13_utility_bias}}

		\makebox[0.23\textwidth][c]{\textbf{NCF}}
		\makebox[0.23\textwidth][c]{\textbf{BPR}}
		\makebox[0.23\textwidth][c]{\textbf{SVD}}
		\makebox[0.23\textwidth][c]{\textbf{VAE}}
	
		\caption{Utility–bias trade-offs across models evaluated under DPSGD and LDP: The Y-axis shows different bias metrics, the X-axis shows utility (NDCG@10), each row represents a different bias metric, and separate columns correspond to different models.
		For LDP and DPSGD, each point corresponds to a specific $\epsilon$.
		The solid line is obtained by fitting a linear regression to the points, showing the trend between utility (NDCG@10) and the corresponding bias metric.}
		\label{fig:plots_utility}
		%\iw{could we make the red circles squares instead, so they are distinguishable for colorblind people? In addition, the legend should say "no privacy" instead of "Normal"}
		%\iw{please standardize y axis limits at least for novelty and dpf, in line with other plots.}
		%\iw{dark blue circles for PL/NCF are not much visible. try to make circle sizes same as in previous plots to reduce overlaps}
	\end{figure}

We now discuss the relationships between utility and bias metrics. 
Figure \ref{fig:plots_utility} shows how bias metrics depend on utility levels (measured by ranking quality NDCG), regardless of the privacy level, i.e., Figure \ref{fig:plots_utility} includes results for \textit{all} $\epsilon$ levels presented in the previous sections.

%Figure \ref{fig:plots_utility} shows the trade-offs between utility (measured by NDCG) and different bias metrics.
%Higher NDCG indicates strong overall performance in predicting user interests.
%If the NDCG for a user or item is low in a non-private setting, this pattern generally persists after applying any Differential Privacy (DP)-based privacy-enhancing approach. 
%Across both datasets (1M and Yelp), at high $\epsilon$ (lower privacy constraints), DPSGD consistently outperforms LDP in ranking effectiveness (measured by NDCG), especially for the NCF model.

\paragraph{Miscalibration (KLD)}
Both LDP and DPSGD show increasing KLD when NDCG rises, meaning that better utility comes at the cost of deviating further from the true user-item interaction distribution.
DPSGD tends to produce lower KLD than LDP at similar utility levels, showing slightly better calibration.
%LDP offers stable performance across privacy levels ($\epsilon$) but with lower NDCG and a pronounced popularity bias, skewing recommendations toward popular items. 

\paragraph{Popularity Lift (PL)}
DPSGD tends to reduce popularity lift (PL of Figure \ref{fig:plots_utility}) compared to LDP, particularly at lower NDCG values.
For both LDP and DPSGD, PL tends to increase toward positive values as NDCG improves, indicating a stronger tendency to recommend popular items when utility increases. 
Under lower NDCG, DPSGD reduces popularity bias (negative popularity lift) by promoting unpopular items, whereas LDP preserves or exacerbates the baseline popularity bias (positive popularity lift).

\paragraph{Novelty}
LDP shows higher novelty when NDCG is higher, as observed for NCF and BPR models.
This indicates that LDP can push the recommender to recommend less frequently seen items. 
DPSGD, however, often sacrifices novelty when NDCG is low.

For 1M, novelty decreases with higher NDCG, whereas for Yelp, it increases. 
The 1M dataset is relatively dense, with a high overlap across users in the training data (i.e., many users rate the same popular items).
Consequently, when the model achieves higher NDCG, it tends to predict mainstream preferences and recommend popular items, resulting in lower novelty.
In contrast, the Yelp dataset is much sparser, with highly heterogeneous user interests.
As a result, a user's top recommendations may correspond to less globally popular but personally relevant businesses.
Models that effectively learn personalized embeddings can therefore identify less popular yet more relevant items.
Hence, in Yelp, higher NDCG indicates more personalized predictions that deviate from global popularity patterns, leading to higher novelty.

\paragraph{Deviation from Producer Fairness (DPF)}
Across all models, DPF generally increases with NDCG, indicating that higher-utility (more accurate) models tend to favor popular items, whereas lower-utility models under privacy constraints may give relatively greater exposure to long-tail/unpopular items. Under DPSGD, DPF becomes negative at low NDCG, suggesting that the injected gradient noise reduces the model’s ability to distinguish between frequent and infrequent items.
This results in a more uniform exposure distribution, sometimes over-representing unpopular items relative to the head/popular items.
However, this apparent fairness improvement is an indirect consequence of degraded model accuracy.
In contrast, DPF under LDP remains largely positive, particularly for NCF and BPR, as perturbing individual user ratings preserves existing popularity trends, maintaining a bias toward head items.
Overall, DPSGD alters training dynamics more globally, flattening popularity biases while simultaneously lowering recommendation utility.
%\iw{please add actual analysis of what we can observe, and why}

\paragraph{Summary}
%Yelp dataset shows greater performance degradation than MovieLens 1M under privacy constraints, likely due to its higher sparsity, which amplifies the disruptive effect of DP noise.
These trade-offs suggest DPSGD is preferable for high-utility, fairness-focused systems, while LDP suits applications prioritizing consistent performance.

\subsection{Discussion}
\label{sec:results-discussion}
Our results show nuanced relationships between different DP approaches, their impact on bias metrics, and the trade-offs between privacy, utility, and bias.
We now discuss what recommendations can be derived from our experiments for scenarios that require (1) high utility, (2) low bias, and (3) high privacy.
%Based on our experiments, we recommend the following approaches.
%\iw{the paragraphs in this section do not flow well. It would be better to start each paragraph with a very brief sentence that summarizes our findings for each point.}

\paragraph{High utility requirement}

%Additionally, for bias-sensitive applications, \ac{DPSGD} should be preferred, as it tends to mitigate exposure bias (DPF) better than LDP.
%However, \ac{DPSGD} can reduce coverage and disproportionately affect niche users.
%\iw{analysis for this paragraph needs to be more in-depth. The sentence about bias-sensitive applications does not belong in a paragraph about high utility requirement.}
When maintaining high recommendation accuracy is the primary objective, \ac{DPSGD} generally outperforms \ac{ldp} at moderate or relaxed privacy budgets ($\epsilon \geq 1$).
Under these settings, DPSGD achieves higher NDCG across models—particularly for NCF—while remaining competitive with the non-private baseline even under $\epsilon < 1$.
This advantage arises because DPSGD perturbs gradients during centralized optimization, preserving global patterns across users that are critical for ranking accuracy.
In contrast, LDP injects noise locally into each user’s data before aggregation, fragmenting user signals and reducing the model’s ability to generalize collectively.

However, this advantage reverses under strict privacy (very low $\epsilon$), where gradient noise dominates learning dynamics.
In such regimes, DPSGD flattens popularity biases but significantly lowers ranking quality.
Achieving high utility with DPSGD therefore requires careful tuning of hyperparameters—such as gradient clipping thresholds, learning rates, and batch sizes—to control the trade-off between noise magnitude and convergence.
In our experiments, moderate privacy levels ($10^{0}<\epsilon<10^{1}$) showed the best trade-offs: NCF/DPSGD achieved about 0.6 NDCG on 1M, while BPR/DPSGD reached around 0.4 on both 1M and Yelp.

Although DPSGD is the preferred option for utility-oriented scenarios, its improvements primarily benefit popular items and blockbuster users, leaving long-tail items comparatively underrepresented. Therefore, when maximizing utility under differential privacy, post-processing calibration or re-ranking may be required to maintain recommendation diversity without compromising accuracy.

%\iw{I think we should add the following two paragraph headings to make the discussion more coherent. add insight to each paragraph, possibly from the text above (marked as "does not fit anymore") or below. In the conclusion, we say ``we found parameter ranges with acceptable trade-offs for some models'' -- these parameter ranges and acceptable trade-offs need to be explained and discussed here.}

\paragraph{Low bias requirement}
%\iw{this is one long paragraph, which tries to make many points. this makes it unreadable and appear rambly. please create smaller paragraphs and think about how to structure your paragraphs, and which argument each paragraph should make - also making sure that the content fits the heading of "low bias requirement".}

For evaluating popularity bias, We recommended using either PL or DPF, as they follow similar patterns.
If the goal is to measure general popularity bias, PL is sufficient, while DPF is preferable when assessing exposure fairness.
For fairness evaluation, it is important to consider the coverage of I1 and I2 (rather than looking only at overall item coverage), as this directly reflects exposure imbalance between head/popular and tail/unpopular items.

Bias should also be analyzed with respect to user type, since its impact is not uniform across user populations.
Niche users typically experience stronger bias toward popular content, especially in denser datasets such as 1M, whereas blockbuster users receive more balanced exposure.
Addressing this disproportionate bias requires mechanisms that explicitly mitigate the skew toward popular items.

Existing bias mitigation strategies can be broadly categorized into pre-processing, in-processing, and post-processing approaches.
In-processing methods, such as Inverse Propensity Scoring \cite{schnabel2016recommendationsa} and the Personal Popularity-Aware Counterfactual (PPAC) framework \cite{ning2024debiasing}, incorporate bias correction into the model’s optimization objective or architecture.
However, their impact on DP remains unclear and requires further investigation.
Post-processing techniques, on the other hand, modify the recommendations after model training. For example, the Calibrated Popularity \cite{abdollahpouri2021usercentered} greedily minimizes the discrepancy between the distribution of item types in a user's profile and their recommended list, thereby addressing popularity bias from the users’ perspective.
These post-processing approaches benefit from the post-processing property \cite{dwork2014algorithmicc} of the DP mechanism, which ensures that any random, data-independent transformations of differentially private outputs does not compromise privacy guarantees.
%\iw{what does this mean? why is it important?For example, }
%Also, the NDCG should be evaluated across different user groups, not just overall performance. 
%A high NDCG for mainstream users does not necessarily mean strong performance for niche users.

Pre-processing approaches also play a role in reducing bias by modifying the dataset before it is fed into the recommender system.
For highly sparse datasets (e.g., Yelp), calibration techniques are particularly important to reduce miscalibration, especially for models like NCF trained under privacy constraints. 
Privacy constrains can exacerbate ranking misalignment, particularly for unpopular items.
%Calibration approaches \cite{steck2018calibrated} can improve this miscalibration.
Calibration ensures that a user's past interests (for example, their preferred genre distribution) are accurately reflected in their recommended list, maintaining the same proportional representation as in their original preferences \cite{steck2018calibrated}.
Item feature calibration, such as genre-based calibration, is a post-processing approach and inherently preserves DP.
However, implementing calibration requires candidate generation, which can pose privacy risks if not done carefully.
Instead of computing scores for all items, recommender systems often rank a smaller candidate set (e.g., the user’s interacted items plus k randomly sampled items).
This approach, however, can leak private information if the candidate selection process does not guarantee differential privacy.
Therefore, DP-aware candidate generation, such as differentially private sampling, should be employed to ensure privacy, or alternatively, fully random selection may be necessary.

\paragraph{High privacy requirement}
%\iw{this is one long paragraph, which tries to make many points. this makes it unreadable and appear rambly. please create smaller paragraphs and think about how to structure your paragraphs, and which argument each paragraph should make - also making sure that the content fits the heading of "high privacy requirement"}

Under strict privacy budgets (low $\epsilon$), both LDP and DPSGD cause a clear degradation in ranking quality (NDCG) across models and datasets, with DPSGD showing the strongest sensitivity to privacy strength.
For NCF and SVD on 1M, lower $\epsilon$ values reduce novelty, implying that privacy noise suppresses exploration of less popular items and reinforces mainstream recommendations.
%Conversely, on Yelp, novelty tends to increase with higher NDCG—suggesting that accurate predictions correspond to more personalized, locally relevant items. 
This contrast arises because MovieLens-1M is dense, with high user–item overlap and strong popularity bias, whereas Yelp is sparse and heterogeneous. 
Thus, in dense datasets, accuracy is driven by popularity alignment (reducing novelty), while in sparse datasets, accuracy stems from personalization (increasing novelty).
These opposite trends reflect the role of dataset density: in dense settings, accuracy is driven by popularity alignment, whereas in sparse datasets, accuracy depends on personalized exploration.
In practice, we observed reasonable trade-offs for moderate privacy budgets, for instance, NCF on 1M achieves approximately 0.6 NDCG at $\epsilon!\approx!10^{-1}$, comparable to higher $\epsilon$ values, while NCF/Yelp exceeds 0.4 NDCG at $\epsilon!\approx!10^{0}$; BPR attains about 0.4 NDCG for both datasets within $10^{0}!<!\epsilon!<!10^{1}$.

%These findings indicate that high privacy constraints interact differently with dataset density, and careful parameter tuning is required to maintain both utility and diversity.
High privacy also affects fairness and bias differently across datasets.
On Yelp, LDP often mitigates fairness disparities (DPF trends toward zero), while on 1M, DPSGD can exacerbate disparities (negative DPF), reflecting a shift toward relatively greater exposure of unpopular items but reduced overall utility.
This suggests that dataset characteristics, such as item popularity distribution, influence how privacy mechanisms affect fairness. 
Popularity lift (PL) further shows that strict privacy (small $\epsilon$) suppresses the amplification of popular items, meaning that both DPSGD and LDP can partially offset inherent popularity bias at the cost of accuracy.

Miscalibration, measured by KLD, remains mostly stable across privacy levels, with one exception:NCF on Yelp shows significantly higher KLD under both DPSGD and LDP.
This pattern points to model-specific vulnerabilities in sparse datasets to the privacy-induced noise.

User-group analysis reveals that high privacy intensifies heterogeneity in performance across users.
Blockbuster users (those preferring popular items) consistently receive the highest NDCG scores across models and datasets, indicating that privacy mechanisms disproportionately preserve accuracy for users whose behavior aligns with global popularity.
In contrast, niche users, who prefer unpopular items, consistently have the lowest NDCG scores, highlighting a utility gap.
This disparity implies that strong privacy constraints hinder the model’s ability to recommend relevant long-tail content effectively. 
Moreover, improvements in overall NDCG as privacy relaxes are primarily driven by popular items, while unpopular items remain largely unaffected.
Similarly, popular items are better calibrated (lower KLD) than unpopular ones, although strict privacy can occasionally narrow this calibration gap by reducing over-confidence in head items.

Overall, high privacy constraints amplify the privacy–utility–fairness tension: while stricter privacy reduces popularity bias and exposure imbalance, it also limits ranking quality, especially for niche users and sparse data regimes.
These findings emphasize that achieving equitable and effective recommendations under strong privacy requires model-specific parameter tuning and, potentially, post-processing adjustments for fairness and calibration.

\subsection{Sustainability}
We estimate energy consumption and carbon emissions for the computations underlying this paper because the global environmental impact of computational activities is an increasingly important concern \cite{lannelongue2021green, patterson2021carbon}.
Our results were obtained using an NVIDIA GeForce RTX 4090 GPU with a rated power consumption of 450 W \cite{nvidiaRTX4090}.
Training our models across four recommender algorithms, ten different noise levels, a non-private baseline, and multiple iterations took approximately 64,000 GPU hours in total (see Appendix \ref{sus_cal} for details).
This corresponds to an estimated 28,800 kWh of energy usage\footnote{Our estimates exclude energy used by the rest of the system (CPU, cooling, storage, etc.) and do not include time spent on hyperparameter tuning and code debugging, which would further increase energy consumption and emissions.}.
Using a global average electricity emissions factor of 460 g CO$_2$/kWh for 2025, this translates to approximately 13.25 metric tons of CO$_2$ emissions from the training process, nearly equivalent to a full commercial flight's emissions of 13.8 metric tons \cite{icao2020freighter2025, IEA2025}. 

To put that number into perspective, we compare it with the energy consumption of a short-haul flight. 
An 808km flight from Zurich (ZRH) to London (LHR), typically operated by an Airbus A320, consumes about 52,439 kWh of jet fuel energy, of which around 18,354 kWh are converted to practical work. 
The GPU training task therefore used approximately 55\% of the flight's total fuel energy, and 157\% of its useful energy.

%For these calculations, we assumed a single RTX 4090, one GPU. 
%Use of multiple GPUs would scale the energy consumption, though the total runtime might decrease with parallelization.

\section{Conclusion}
In this paper, we presented a comprehensive evaluation %\iw{"initial results" are not what we want to present in a journal paper!} 
of the impact of differentially private training on the performance and bias of recommender systems.
Our evaluation considered multiple metrics to measure utility, bias, and privacy, and assessed the effects of both LDP and DPSGD.
We experimented with four recommender systems -- SVD and BPR as classic approaches, and NCF and VAE as deep learning-based methods -- on the MovieLens and Yelp datasets.
Our findings highlight how different DP techniques influence various aspects of these models, providing insights into the trade-offs between privacy, accuracy, and bias in recommender systems.
%differentially private systems may not lose much accuracy when 
%In conclusion, the study demonstrates that increasing the noise multiplier leads to higher Epsilon values, affecting the privacy-accuracy trade-off across various recommendation models and datasets.
%The Yelp dataset, in particular, shows higher sensitivity to noise compared to the 1M dataset.
%Among the models, SVD exhibits the largest performance gaps, indicating a higher sensitivity to privacy mechanisms, while the BPR model requires higher Epsilon values to maintain satisfactory performance.
%Notably, the DRS model incurs an acceptable performance cost in implementing privacy, particularly over the 1M dataset.

We identified ranges of the privacy budget in which acceptable trade-offs between privacy, accuracy, and bias can be achieved. 
These findings highlight that practitioners should not rely on a single metric or fixed privacy level; instead, they must systematically experiment with different parameters, such as the choice of $\epsilon$, to determine which configuration best aligns with their priorities. 
Depending on whether utility, bias mitigation, or calibration is more critical, different parameter settings may produce more desirable outcomes. 
Consequently, evaluating multiple metrics simultaneously is essential, and the intended objective must be explicitly defined to select the most appropriate privacy configuration.

The results also show several areas that deserve further study, including the effects of the model characteristics on privacy and bias, and the design of more robust models that can better balance privacy and performance.
Future work could explore the development of calibration-based post-processing methods that preserve privacy guarantees, as well as extensions to more robust privacy-preserving recommenders.
This includes evaluating user-level differential privacy instead of interaction-level protection~\cite{fang2022differentially}, and integrating embedding-aware noise mechanisms~\cite{ning2022eana} to better preserve representational structure under privacy constraints.

%specifically assessing its effects on metrics like NDCG and various bias measurements.
%Therefore, we think that by exploring the space of various model and DPSGD parameters, it is possible to achieve satisfactory performance at certain points with a reasonable privacy budget
%Additionally, privacy mechanisms significantly impact model calibration, especially in the Yelp dataset, with the BPR model showing the greatest shifts in behavior.
%These findings emphasize the need for more robust models that can better balance privacy and performance in recommendation systems.
%Additionally, we will explore deeper into the trade-offs between calibration, privacy, performance, and bias to better understand their interdependencies.
%An in-depth comparative analysis with related works such as DP-VAE \cite{fang2022differentially}, and EANA \cite{ning2022eana}  will also be conducted to evaluate the strengths and limitations of these methods in enhancing privacy-preserving recommendation systems.

%\todo[inline]{future work:
%- calibration can be applied as a post-processing step that is privacy-neutral; evaluate the effect on ndcg and bias metrics
%- further facets of trade-off: impact of calibration on privacy, performance, bias
%- add in-depth comparison with related work such as DP-VAE, DP-CMF, EANA
%}

%\iw{check references: there are duplicates; check locations of conferences that currently list New York, this is likely incorrect. Please fix entries in Zotero, not in the bib file!}
\bibliographystyle{ACM-Reference-Format}
\bibliography{2023_Shiva_Fairness_and_Performance_of_privacy_preserving_recommender_systems}

\appendix
\section*{Appendix}
\section{Sustainability Calculations}
\label{sus_cal}

In this section, we report the average energy consumption required to produce the results presented in this paper, focusing on the GPU training tasks.
Our approach estimates the total energy consumed by the NVIDIA GeForce RTX 4090 graphics card during model training, expressed in kilowatt-hours (kWh). 
To provide context and make these numbers more tangible, we compare this energy consumption to that of a typical short-haul flight on an Airbus A320, calculating both the energy content of the fuel burned and the resulting carbon dioxide ($CO_{2}$) emissions in metric tons.
Finally, we estimate the carbon emissions attributable to the GPU training and compare these to the flight emissions, highlighting the environmental impact of computational workloads.

%\iw{introductory sentence: why sustainability reporting matters, which metrics we are estimating and reporting, and what our approach to estimation is}
%\iw{we don't really care about Table 1 -- we care about all training that was needed to generate the results in the entire paper. What is the estimate for that?}
%This corresponds to an energy usage of 450 joules per second.
%When running continuously at this power, the RTX 4090 consumes: \[
%0.45\,\text{kW} \times 1\,\text{hour} = 0.45\,\text{kWh/hour}
%\]
For our training task, completing one epoch across the four approaches, 10 different noise/epsilon levels, and one non-private case requires approximately 160 hours.
With an average of 100 epochs per noise/epsilon level, the total training time per noise/epsilon level (in a single replication) is:
%\iw{what are "runs"? epochs? or replications of model training?}
\[
160\ \text{hours/run} \times 100\ \text{runs} = 16{,}000\ \text{hours}
\]
The results are based on different iterations per setting(as explained in Section \ref{sec:expsettings}), here we consider in average 4 iterations which leading to a total training time of: 
\[
\frac{16{,}000\ \text{hours}}{\text{iteration}} \times 4\ \text{iterations} = 64{,}000\ \text{hours}
\]
%\iw{why is epsilon in this equation?}
Therefore, the total energy consumed by the RTX 4090 for making this table is:
\(
64{,}000\ \text{hours} \times 0.45\ \text{kW} = 28{,}800\ \text{kWh}
\).
%\iw{check your math. the previous line is not correct.}An 
The flight distance between Zurich (ZRH) and London (LHR) is approximately 808 km, typically covered by a short-haul aircraft like the Airbus A320 in 1.75 hours.
Airbus A320 consumes approximately 2,500 kg of Jet A-1 fuel per hour during flight \cite{aviexFuelUse}.
For a 1.75-hour flight, the total fuel consumption is: \(2{,}500\,\text{kg/hour} \times 1.75\,\text{hours} = 4{,}375\,\text{kg}\).
Jet A-1 fuel has an energy content of 43.15 MJ/kg (43,150,000 J/kg).
The total energy content of the fuel is: \(4{,}375\,\text{kg} \times 43{,}150{,}000\,\text{J/kg} = 188{,}781{,}250{,}000\,\text{J}\).
Converting to kilowatt-hours (1 kWh = 3,600,000 J): \(\frac{188{,}781{,}250{,}000\,\text{J}}{3{,}600{,}000} = 52{,}439\,\text{kWh}\).
Jet engines have an efficiency of approximately 30–40\% in converting fuel energy to practical work (e.g., thrust).
Assuming 35\% efficiency, the useful energy output is: \(52{,}439\,\text{kWh} \times 0.35 = 18{,}354\,\text{kWh}\).
The RTX 4090's training task consumes 28,800 kWh, which is approximately 55\% of the flight's total fuel energy (52,439 kWh), and is approximately 157\% of the flight's useful energy output (18,354 kWh).

For Jet A-1 fuel, the emission factor is approximately 3.16 (\cite{icao2020freighter2025}).
%\iw{reference is missing}
This emission factor is the number of tonnes of CO2 produced by burning a tonne of aviation fuel.
Based on the consumed fuel and the emission factor, the total emission is: $4{,}375\,\text{kg fuel} \times 3.16\,\frac{\text{kg CO}_2}{\text{kg fuel}} = 13{,}825\,\text{kg CO}_2$.
This is approximately 13.8 metric tons of CO2 for one flight.
The RTX 4090 training task's carbon emissions are calculated using a global average emission factor for electricity of 460 g CO$_2$/kWh for 2025.
This value is estimated by applying a 3\% reduction in emissions intensity, as reported for 2024 \citep{IEA2025}, to the 2018 baseline of 475 g CO$_2$/kWh \citep{IEA2019}.
For 28,800 kWh, the emissions are: 
\begin{equation}
	28{,}800 \ \text{kWh} \times 0.46 \ \frac{\text{kg CO}_2}{\text{kWh}} = 13{,}248 \ \text{kg CO}_2 \approx 13.25 \ \text{metric tons CO}_2.
	\end{equation}
This emission level is more than a full flight's emissions.

\end{document}